\documentclass{article}
\usepackage[OT1]{fontenc}
\providecommand{\OVDPath}{}
\PassOptionsToPackage{table}{xcolor}
\usepackage{iclr2027_conference,times}

\usepackage{microtype}
\usepackage{hyperref}
\usepackage{url}
\usepackage{booktabs}

\definecolor{darkblue}{rgb}{0, 0, 0.5}
\hypersetup{colorlinks=true, citecolor=darkblue, linkcolor=darkblue, urlcolor=darkblue}

\usepackage{graphicx}
\usepackage{subcaption}
\usepackage{amssymb}
\usepackage{amsmath}
\usepackage{mathtools}
\usepackage{amsthm}
\usepackage{multirow}
\usepackage{xcolor}
\usepackage{arydshln}
\usepackage{tabularx}
\usepackage{adjustbox}
\usepackage{pifont}
\usepackage{tcolorbox}
\tcbuselibrary{breakable}
\definecolor{calloutback}{gray}{0.97}
\definecolor{calloutframe}{gray}{0.68}
\definecolor{callouttitle}{gray}{0.88}
\usepackage{tikz}
\usepackage{pgfplots}
\pgfplotsset{compat=1.18}
\usepackage{float}
\usepackage{wrapfig}
\usepackage{needspace}
\makeatletter
\@ifundefined{WFclear}{\def\WFclear{\par}}{}%
\makeatother
\usepackage{placeins}
\usepackage{algorithm}
\usepackage{algorithmic}
\usepackage[capitalize,noabbrev]{cleveref}

\theoremstyle{plain}
\newtheorem{theorem}{Theorem}[section]
\newtheorem{proposition}[theorem]{Proposition}

\theoremstyle{definition}

\newtheorem{assumption}[theorem]{Assumption}
\theoremstyle{remark}

\title{\textsc{OVD}: On-policy Verbal Distillation}

\iclrfinalcopy
\author{%
\begin{minipage}{\textwidth}
\centering\normalfont\large
Jing Xiong\textsuperscript{1},
Hui Shen\textsuperscript{1},
Shansan Gong\textsuperscript{1},
Yuxin Cheng\textsuperscript{1},
Jianghan Shen\textsuperscript{2},\\[2pt]
Chaofan Tao\textsuperscript{3},
Haochen Tan\textsuperscript{3},
Haoli Bai\textsuperscript{3},
Lifeng Shang\textsuperscript{3},
Ngai Wong\textsuperscript{1}\\[9pt]
{\normalsize
\adjustbox{max width=\textwidth}{\small
\textsuperscript{1}\,The University of Hong Kong\enspace
\textsuperscript{2}\,Nanjing University\enspace
\textsuperscript{3}\,Huawei Technologies}\endgraf}
\end{minipage}%
}

\makeatletter
\renewcommand{\@maketitle}{%
  \vbox{\hsize\textwidth
    \centering
    {\LARGE\bfseries\@title\par}
    \vskip 12pt
    {\normalfont\@author\par}
    \vskip 8pt
  }%
}
\makeatother

\hypersetup{pdfauthor={Jing Xiong, Hui Shen, Shansan Gong, Yuxin Cheng, Jianghan Shen, Chaofan Tao, Haochen Tan, Haoli Bai, Lifeng Shang, Ngai Wong},pdftitle={OVD: On-policy Verbal Distillation}}

\begin{document}

\raggedbottom

\maketitle
\fancyhead{}
\fancyhead[L]{\small\textsc{Technical Report}}
\renewcommand{\headrulewidth}{0.3pt}
\suppressfloats[t]

\begin{abstract}
Knowledge distillation transfers reasoning capabilities from large teachers to efficient students. However, token-level on-policy distillation (OPD) constrains student exploration and requires teacher token probabilities, precluding distillation from black-box teachers that provide only text outputs. We introduce \emph{On-policy Verbal Distillation} (\textsc{OVD}), a framework that uses verbal scores from black-box teachers to rank student-generated sub-trajectories, retaining high-scoring ones and replacing low-scoring ones with teacher-generated continuations. We analyze when ranking induced by verbal scores can guide distribution approximation: under a density-ratio calibration condition on acceptance probabilities and bounded teacher-replacement error, we bound the approximation error between the resulting mixed trajectory distribution and a teacher-preferred target. On Web Q\&A, \textsc{OVD} achieves 41.09\% average EM with teacher feedback at inference, exceeding the strongest evaluated baseline by 5.89 percentage points. On AMC23, \textsc{OVD}-FR improves accuracy over RLVR by 10.0 percentage points (52.5\% to 62.5\%) after 600 training steps on 128 problems. Further experiments suggest that retaining student-generated prefixes helps preserve exploration and mitigate trajectory-level entropy collapse. \textsc{OVD} also improves training efficiency: resampling selected suffixes rather than entire responses reduces mean per-step training time by 10.2\% in the 128-problem setting. Project page: \url{https://menik1126.github.io/ovd-project-page/}.

\end{abstract}

\vspace{-3mm}
\section{Introduction}
\label{sec:introduction}
\vspace{-2mm}
Knowledge distillation transfers capabilities from large teachers to smaller
policies. On-policy distillation (OPD) learns from student rollouts but typically
matches next-token distributions~\citep{hinton2015distilling,agarwal2024onpolicy}.
For long-horizon reasoning, particularly in agentic settings, these dense
probabilities are costly to retain and unavailable from black-box teachers,
while token-level matching may penalize deviations from the teacher at every
step, potentially suppressing trajectory-level exploration. These limitations
motivate a trajectory-level alternative that uses verbal feedback to shape the
distribution of on-policy student rollouts toward trajectories preferred by
the teacher.

We introduce \emph{On-policy Verbal Distillation} (\textsc{OVD}). A teacher
scores each student sub-trajectory; high-scoring segments are retained,
low-scoring ones are replaced with teacher continuations, and the student then
resumes generation. This procedure preserves student exploration while
steering the induced mixed-trajectory distribution toward a teacher-preferred
target; the resulting trajectories are then used for outcome-based policy
optimization.

\vspace{-1mm}
\subsection{Limitations of Token-Level OPD}
\vspace{-1mm}
\label{subsec:token_opd_limitations}
\begin{figure*}[t]
\centering
\captionsetup[subfigure]{skip=1pt}
\vspace{0.8em}
\begin{subfigure}[t]{0.247\textwidth}
\centering
\begin{minipage}[t][3.3cm][t]{\linewidth}
\centering
\definecolor{memLogit}{HTML}{D55E00}
\definecolor{memCache}{HTML}{0072B2}

\begin{tikzpicture}
\begin{axis}[
  scale only axis,
  width=2.45cm,
  height=2.45cm,
  xmode=log,
  log basis x=2,
  ymode=log,
  log basis y=10,
  xmin=450,
  xmax=38000,
  ymin=0.02,
  ymax=40,
  xtick={512,2048,8192,32768},
  xticklabels={0.5K,2K,8K,32K},
  ytick={0.03,0.1,1,10,30},
  yticklabels={0.03,0.1,1,10,30},
  xlabel={Context length $L$},
  ylabel={Memory (GB)},
  tick label style={font=\tiny},
  label style={font=\scriptsize},
  xlabel shift=-2pt,
  ylabel shift=-2pt,
  axis line style={draw=black!55,line width=0.4pt},
  tick style={draw=black!55,line width=0.4pt},
  grid=major,
  major grid style={draw=black!12,line width=0.35pt},
  minor tick num=0,
  legend style={
    at={(0.025,0.975)},
    anchor=north west,
    draw=black!15,
    fill=white,
    fill opacity=0.92,
    text opacity=1,
    rounded corners=1pt,
    inner sep=1.2pt,
    font=\tiny,
    row sep=-1pt
  },
  legend image post style={scale=0.7},
  clip=false
]

\addplot[
  color=memLogit,
  line width=1.0pt,
  mark=*,
  mark size=1.5pt
] coordinates {
  (512,0.4375) (1024,0.875) (2048,1.750) (4096,3.500)
  (8192,7.000) (16384,14.000) (32768,28.000)
};
\addlegendentry{Dense logits}

\addplot[
  color=memCache,
  line width=1.0pt,
  mark=square*,
  mark size=1.4pt
] coordinates {
  (512,0.0275) (1024,0.055) (2048,0.110) (4096,0.220)
  (8192,0.440) (16384,0.880) (32768,1.760)
};
\addlegendentry{KV cache}

\end{axis}
\end{tikzpicture}
\end{minipage}
\caption{Per-trajectory memory.}
\label{fig:memory_breakdown}
\end{subfigure}%
\hspace{0.004\textwidth}%
\begin{subfigure}[t]{0.247\textwidth}
\centering
\begin{minipage}[t][3.3cm][t]{\linewidth}
\centering
\begin{tikzpicture}
\begin{axis}[
  scale only axis,
  width=2.45cm,
  height=2.45cm,
  xlabel={Context length $L$},
  ylabel={Memory (GB)},
  xmin=0, xmax=32768,
  ymin=0, ymax=1000,
  xtick={0,8192,16384,32768},
  xticklabels={0,8K,16K,32K},
  scaled x ticks=false,
  grid=major,
  legend style={
    font=\tiny,
    at={(0.34,1.02)},
    anchor=south,
    legend columns=2,
    inner sep=1.2pt,
    column sep=0.4em,
    overlay
  },
  tick label style={font=\tiny},
  label style={font=\scriptsize},
  xlabel shift=-2pt,
  ylabel shift=-2pt
]
\addplot[black,very thick] coordinates {(0,0) (4096,119.54) (8192,239.08) (16384,478.15) (24576,717.23) (32768,956.30)};
\addlegendentry{Full-$V$}
\addplot[violet,very thick,dotted] coordinates {(0,0) (4096,6.44) (8192,12.88) (16384,25.77) (24576,38.65) (32768,51.54)};
\addlegendentry{Top-$k$}
\end{axis}
\end{tikzpicture}
\end{minipage}
\caption{Aggregate ($N{=}32$).}
\label{fig:memory_scaling}
\end{subfigure}%
\hspace{0.004\textwidth}%
\begin{subfigure}[t]{0.247\textwidth}
\centering
\begin{minipage}[t][3.3cm][t]{\linewidth}
\centering
\begin{tikzpicture}
\begin{axis}[
    scale only axis,
    width=2.45cm,
    height=2.45cm,
    xlabel={Training step},
    ylabel={PopQA EM},
    xmin=20, xmax=205,
    enlarge x limits=false,
    ymin=0.08, ymax=0.62,
    grid=major,
    grid style={thin,gray!30},
    legend style={
      font=\tiny,
      at={(1.20,1.02)},
      anchor=south,
      legend columns=4,
      inner xsep=0.6pt,
      inner ysep=1.0pt,
      overlay,
      /tikz/every even column/.append style={column sep=0pt},
      legend image code/.code={\draw[##1] (0cm,0cm)--(0.22cm,0cm);}
    },
    x tick label style={font=\tiny},
    y tick label style={font=\tiny},
    label style={font=\scriptsize},
    xlabel shift=-2pt,
    ylabel shift=-2pt,
    xtick={25,75,125,175},
]
\addplot[color=teal!80!black, mark=*, thick, mark size=1.4pt] coordinates {
    (25, 0.196) (50, 0.338) (75, 0.474) (100, 0.538) (125, 0.524) (150, 0.552) (175, 0.544) (200, 0.575)
};
\addlegendentry{Top-5 val.}

\addplot[color=orange!85!black, mark=square*, thick, mark size=1.4pt] coordinates {
    (25, 0.198) (50, 0.334) (75, 0.478) (100, 0.528) (125, 0.524) (150, 0.540) (175, 0.462) (200, 0.414)
};
\addlegendentry{Top-1 val.}

\addplot[color=teal!80!black, dashed, thick, line width=0.85pt] coordinates {(20, 0.112) (205, 0.112)};
\addlegendentry{Top-5 test}

\addplot[color=orange!85!black, dashed, thick, line width=0.85pt] coordinates {(20, 0.134) (205, 0.134)};
\addlegendentry{Top-1 test}
\end{axis}
\end{tikzpicture}
\end{minipage}
\caption{PopQA.}
\label{fig:token_distill_popqa_sub}
\end{subfigure}%
\hspace{0.004\textwidth}%
\begin{subfigure}[t]{0.247\textwidth}
\centering
\begin{minipage}[t][3.3cm][t]{\linewidth}
\centering
\begin{tikzpicture}
\begin{axis}[
    scale only axis,
    width=2.45cm,
    height=2.45cm,
    xlabel={Training step},
    ylabel={HotpotQA EM},
    xmin=20, xmax=205,
    enlarge x limits=false,
    ymin=0.08, ymax=0.48,
    grid=major,
    grid style={thin,gray!30},
    x tick label style={font=\tiny},
    y tick label style={font=\tiny},
    label style={font=\scriptsize},
    xlabel shift=-2pt,
    ylabel shift=-2pt,
    xtick={25,75,125,175},
]
\addplot[color=teal!80!black, mark=*, thick, mark size=1.4pt] coordinates {
    (25, 0.096) (50, 0.210) (75, 0.252) (100, 0.352) (125, 0.306) (150, 0.406) (175, 0.368) (200, 0.398)
};

\addplot[color=orange!85!black, mark=square*, thick, mark size=1.4pt] coordinates {
    (25, 0.126) (50, 0.192) (75, 0.270) (100, 0.340) (125, 0.310) (150, 0.354) (175, 0.378) (200, 0.332)
};

\addplot[color=teal!80!black, dashed, thick, line width=0.85pt] coordinates {(20, 0.120) (205, 0.120)};

\addplot[color=orange!85!black, dashed, thick, line width=0.85pt] coordinates {(20, 0.132) (205, 0.132)};
\end{axis}
\end{tikzpicture}
\end{minipage}
\caption{HotpotQA.}
\label{fig:token_distill_hotpotqa_sub}
\end{subfigure}
\caption{Two limitations of token-level distillation. \textbf{(a,b)} On Qwen2.5-7B, full-vocabulary FP32+BF16 logits scale linearly with context length: at $L{=}8192$, they require 7.0\,GB per trajectory versus 0.44\,GB for the BF16 KV cache, and approximately 240\,GB in aggregate for $N{=}32$ versus 13\,GB for top-$k$ ($k{=}8192$). \textbf{(c,d)} On PopQA and HotpotQA, simulator validation EM generally increases over training under top-1/top-5 token-level distillation, whereas held-out live-search EM is much lower; top-5 underperforms top-1 on both datasets.}
\label{fig:token_distillation_limits}
\vspace{-0.35\baselineskip}
\end{figure*}

\paragraph{Standard OPD.}\label{para:problem_formulation}
Let $\pi_S$ and $\pi_\mathcal{E}$ denote the student and teacher policies, respectively. OPD~\citep{agarwal2024onpolicy} trains on trajectories sampled from $\pi_S$ while matching the next-token distribution of $\pi_S$ to the \emph{full-vocabulary} distribution of $\pi_\mathcal{E}$ (dense logits over all $V$ tokens at each decoding step). For a problem $q$, its objective is the reverse KL over next tokens $x_\ell$ sampled from $\pi_S$,
\begin{equation}
\label{eq:opd_reverse_kl}
\mathrm{KL}\bigl(\pi_S\Vert\pi_{\mathcal{E}}\bigr)
= \mathbb{E}\Bigl[
  \log \pi_S(x_\ell\mid q,x_{<\ell}) - \log \pi_{\mathcal{E}}(x_\ell\mid q,x_{<\ell})
\Bigr].
\end{equation}
\noindent For policy optimization, the sampled negative log-ratio $\log \pi_{\mathcal E}(x_\ell\mid q,x_{<\ell}) - \log \pi_S(x_\ell\mid q,x_{<\ell})$ serves as the token-level advantage. Its expectation under $\pi_S$ equals the negative reverse KL, so maximizing the corresponding importance-weighted policy objective minimizes Eq.~\eqref{eq:opd_reverse_kl}~\citep{lu2025onpolicydistillation}. We refer to this full-vocabulary objective as \emph{standard OPD}.
\begin{tcolorbox}[
  colback=callouttitle,
  colframe=calloutframe,
  boxrule=0.6pt,
  arc=1pt,
  boxsep=3pt,
  left=4pt,right=4pt,top=3pt,bottom=3pt,
  before skip=5pt,
  after skip=5pt,
]
\textbf{Findings 1}: \textit{Full-vocabulary probability matching creates a memory bottleneck}
\end{tcolorbox}
\vspace{-0.3\baselineskip}

\paragraph{Memory Bottleneck.}\label{para:memory_bottleneck}
The full-vocabulary objective above requires dense student and teacher distributions over all $V$ tokens at every decoding step of each student-generated trajectory. Figure~\ref{fig:token_distillation_limits}(a,b) shows that dense logit storage grows linearly with context length on Qwen2.5-7B~\citep{qwen2025qwen25technicalreport} and becomes substantially larger when aggregated over $N{=}32$ rollouts. Closed-source teachers also typically do not expose such distributions. With a batch of $B$ problems, $N$ trajectories per problem, and rollout length $L$, a standard full-vocabulary logit-storage budget is
\begin{equation}
\mathcal{M} = B \cdot N \cdot L \cdot V \cdot (d_{32} + d_{16}) = 1.5 \cdot B \cdot N \cdot L \cdot V \cdot d_{32}.
\end{equation}
\noindent Here $d_{32}{=}4$ and $d_{16}{=}2$ bytes; many stacks materialize logits in both precisions~\citep{apple2024cce}, using BF16 for most training computation and FP32 for numerically sensitive operations such as softmax and KL. Consequently, full-vocabulary logit storage scales as $\mathcal{O}(B\cdot N\cdot L\cdot V)$. At long context lengths and large rollout counts, this scaling makes logit storage a dominant memory bottleneck, motivating sparse alternatives.

\vspace{-2mm}
\paragraph{Top-$k$ OPD.}
One extreme retains only the teacher log-probability assigned to each student-sampled token, exactly matching the sampled-token reverse-KL estimator of~\citet{lu2025onpolicydistillation}. A broader sparse alternative retains the teacher's $k$ highest-probability tokens per decoding step. For this top-$k$ OPD baseline, we restrict both the student and teacher distributions to this support and renormalize them before matching. This lowers logit-storage costs to $\mathcal{O}(B\cdot N\cdot L\cdot k)$, an approximately $V/k$-fold reduction, but top-$k$ OPD still relies on token-level probability matching and requires white-box teacher access to obtain truncated logits. How the choice of $k$ affects policy exploration remains an open question.

\begin{tcolorbox}[
  colback=callouttitle,
  colframe=calloutframe,
  boxrule=0.6pt,
  arc=1pt,
  boxsep=3pt,
  left=4pt,right=4pt,top=3pt,bottom=3pt,
  before skip=5pt,
  after skip=5pt,
]
\textbf{Findings 2}: \textit{Token-level probability matching may restrict exploration}
\end{tcolorbox}

Token-level matching to the teacher's next-token distribution~\citep{hinton2015distilling,gu2024minillm} may reduce behavioral diversity~\citep{agarwal2024onpolicy}. In the agentic setting, Figures~\ref{fig:token_distill_popqa_sub} and~\ref{fig:token_distill_hotpotqa_sub} show that simulator validation EM generally increases over training under both top-1 and top-5 token-level distillation, whereas held-out live-search EM is much lower, with top-5 underperforming top-1 on both datasets. This pattern suggests that denser token-level matching may narrow policy exploration.
\WFclear

\vspace{-3mm}
\subsection{Distribution Approximation via Ranking}
\vspace{-2mm}
\label{subsec:verbal_distribution_approximation}

\noindent These limitations motivate our central question: \textit{under black-box teacher access, can ranking on-policy student trajectories with discrete verbal scores approximate a teacher-preferred distribution while still allowing the student to explore its trajectory space?}

\noindent We use a \emph{verbal scoring interface} in which the teacher provides quality assessments as discrete score tokens rather than full-vocabulary logits. Given a problem $q$ and a student-generated sub-trajectory $y$, the teacher induces a distribution over a verbal vocabulary of size $v=10$. By default, we greedily decode the highest-probability score token and use it as $S(y,q)\in\{0,1,\ldots,9\}$, with higher values indicating stronger preference for the sub-trajectory. When logits for these score tokens are available, applying a softmax over the ten logits yields the score distribution, from which scores can instead be sampled. The default greedy procedure remains compatible with black-box access because it does not require the logits to be exposed.

\begin{tcolorbox}[
  colback=callouttitle,
  colframe=calloutframe,
  boxrule=0.6pt,
  arc=1pt,
  boxsep=3pt,
  left=4pt,right=4pt,top=3pt,bottom=3pt,
  before skip=5pt,
  after skip=5pt,
]
\textbf{Findings 3}: \textit{Ranking induced by verbal scores can select high-quality sub-trajectories}
\end{tcolorbox}

\begin{wrapfigure}{r}{0.44\textwidth}
  \vspace{-1.0\baselineskip}
  \centering
  \begin{tikzpicture}
    \begin{axis}[
      width=\linewidth,
      height=4.0cm,
      xmin=-0.3,
      xmax=9.3,
      ymin=10,
      ymax=82,
      xlabel={Verbal-score threshold $\tau_{\mathrm{te}}$},
      ylabel={Final-answer EM (\%)},
      xtick={0,1,2,3,4,5,6,7,8,9},
      ytick={20,30,40,50,60,70,80},
      grid=major,
      tick label style={font=\tiny},
      label style={font=\tiny},
      legend style={
        font=\fontsize{4.2}{4.7}\selectfont,
        at={(0.03,0.97)},
        anchor=north west,
        draw=none,
        fill=white,
        fill opacity=0.88,
        text opacity=1,
      },
      legend image post style={line width=0.6pt,mark size=1.1pt},
      every axis plot/.append style={line width=0.7pt},
      axis on top,
      clip=false,
    ]
      \addplot[
        draw=none,
        fill=blue!45,
        fill opacity=0.20,
        forget plot,
      ] coordinates {
        (0,21.90)
        (1,23.34)
        (2,23.48)
        (3,23.64)
        (4,40.66)
        (5,44.76)
        (6,61.63)
        (7,69.49)
        (8,70.66)
        (9,78.14)
        (9,50.00)
        (8,49.77)
        (7,48.42)
        (6,41.90)
        (5,28.41)
        (4,25.56)
        (3,14.37)
        (2,14.29)
        (1,14.20)
        (0,13.31)
      } \closedcycle;

      \addplot[
        draw=none,
        fill=orange!55,
        fill opacity=0.16,
        forget plot,
      ] coordinates {
        (0,21.90)
        (1,22.77)
        (2,22.76)
        (3,22.93)
        (4,35.05)
        (5,38.28)
        (6,51.62)
        (7,57.12)
        (8,58.31)
        (9,67.70)
        (9,41.73)
        (8,37.43)
        (7,36.68)
        (6,32.10)
        (5,23.25)
        (4,21.01)
        (3,13.80)
        (2,13.66)
        (1,13.74)
        (0,13.31)
      } \closedcycle;

      \addplot[
        gray!75!black,
        densely dotted,
        line width=0.65pt,
        forget plot,
      ] coordinates {
        (0,17.36)
        (9,17.36)
      };

      \addplot[
        blue!70!black,
        solid,
        mark=*,
        mark size=1.3pt,
        mark options={fill=blue!70!black},
      ] coordinates {
        (0,17.36)
        (1,18.48)
        (2,18.59)
        (3,18.69)
        (4,33.20)
        (5,36.81)
        (6,52.50)
        (7,60.00)
        (8,61.22)
        (9,65.94)
      };
      \addlegendentry{Verbal-score selection}

      \addplot[
        orange!85!black,
        dashed,
        mark=*,
        mark size=1.3pt,
        mark options={fill=orange!85!black},
      ] coordinates {
        (0,17.36)
        (1,18.00)
        (2,17.98)
        (3,18.10)
        (4,28.14)
        (5,31.02)
        (6,42.67)
        (7,48.07)
        (8,49.06)
        (9,56.84)
      };
      \addlegendentry{Per-prompt random (same count)}

      \addlegendimage{gray!75!black,densely dotted,line width=0.65pt}
      \addlegendentry{Pooled random (same total count)}
    \end{axis}
  \end{tikzpicture}
  \captionsetup{font=normalsize,skip=1pt}
  \caption{Initial-prefix selection and continuation EM on a fixed 128-problem subset of the DeepScaleR-Preview-Dataset, with 32 fixed student-generated prefixes per problem (4,096 total), reused across thresholds. The teacher scores only prefixes; those scoring at least $\tau_{\mathrm{te}}$ receive student-generated continuations for EM evaluation. At $\tau_{\mathrm{te}}{=}8$, 441 prefixes remain: EM is 61.22\% for verbal-score selection, 49.06\% for per-prompt random, and 17.36\% for pooled random. Shading denotes 95\% confidence intervals.}
  \label{fig:vrs_precision_cost_frontier}
  \vspace{-1.0\baselineskip}
\end{wrapfigure}

\noindent Figure~\ref{fig:vrs_precision_cost_frontier} asks whether the teacher can select promising initial prefixes. In the verbal-score condition, the teacher scores each initial prefix; given a test-time threshold $\tau_{\mathrm{te}}$, the student continues from prefixes whose scores are at least $\tau_{\mathrm{te}}$. In the \emph{per-prompt random} condition, prefixes are chosen randomly within each prompt, preserving the number selected by verbal-score selection. In the \emph{pooled random} condition, all prefixes are pooled together. We randomly select the same total number of prefixes as in the verbal-score condition, without requiring the selected prefixes to be distributed across prompts in the same way. If verbal scores successfully identify promising initial prefixes, continuing from the selected prefixes should yield higher final-answer EM than either random baseline. Consistent with this prediction, verbal-score selection outperforms both baselines; in particular, its gain over per-prompt random provides evidence that the scores distinguish prefix quality within a prompt.

\noindent \textsc{OVD} uses this signal for \emph{verbal rejection sampling},
retaining strong student sub-trajectories and replacing weak ones with teacher
continuations. Thresholding verbal scores converts the teacher's ranking into
acceptance probabilities $a_\tau$, which reweight student proposals. We analyze
when this reweighting recovers a teacher-preferred distribution $p_T$: under a
valid density-ratio condition
$a_\tau(y\mid q)\propto p_T(y\mid q)/\pi_S(y\mid q)$, the accepted student
component matches $p_T$. Ranking accuracy alone does not establish this
condition. Under approximate calibration and bounded teacher-replacement error
at each prefix, we further quantify how these errors accumulate across
reasoning steps and bound the resulting deviation from $p_T$.
Section~\ref{sec:vrs_distribution_approximation}
formalizes this distributional framework.~\textbf{Contributions.}
\WFclear
\vspace{-0.35\baselineskip}
\begingroup
\setlength{\leftmargini}{1.0em}
\begin{itemize}\setlength{\itemsep}{0pt}\setlength{\parsep}{0pt}\setlength{\topsep}{1pt}
    \item We introduce \textsc{OVD}, a black-box on-policy distillation framework
    that steers student rollouts with verbal scores and selective teacher
    continuations, without requiring teacher logits.
    \item We identify sufficient conditions under which verbal-score filtering
    and teacher replacement approximate a teacher-preferred trajectory
    distribution, and quantify the approximation error.
    \item We evaluate \textsc{OVD} across eight Web Q\&A benchmarks and tool-free
    mathematics, examining the effectiveness of ranking induced by verbal
    scores and selective teacher resampling.
\end{itemize}
\endgroup

\FloatBarrier

\vspace{-0.35\baselineskip}

\kern-0.5ex
\section{On-policy Verbal Distillation}
\label{sec:method}
\vspace{-0.5ex}
In this section, we present \textsc{OVD}, which explores how verbal rejection sampling over on-policy student rollouts can approximate a teacher-preferred trajectory distribution. It then trains the student on the resulting mixed trajectories.

\vspace{-3mm}
\paragraph{Problem Setup.}
Consider a teacher model $\pi_\mathcal{E}$ and a student policy $\pi_S$. Given a problem $q$, the student policy generates multi-step reasoning trajectories, while the teacher model provides verbal feedback scores. We write a trajectory as $y = (s_1, \ldots, s_T)$, where $t\in\{1,\ldots,T\}$ indexes the reasoning steps, each $s_t$ is a reasoning step (e.g., a sentence or a logical derivation), $y_{<t} = (s_1,\ldots,s_{t-1})$ is the trajectory prefix, and $T$ is the number of steps. We reserve $x_\ell$ for token-level quantities, where $\ell$ indexes token positions; in Eq.~\eqref{eq:opd_reverse_kl}, standard OPD matches $\pi_S(x_\ell\mid q,x_{<\ell})$ to $\pi_\mathcal{E}(x_\ell\mid q,x_{<\ell})$ over the vocabulary. By contrast, $s_t$ indexes \emph{coarse} segments of the rollout, not single tokens.
We use $\tau$ for a generic score threshold and write $\tau_{\mathrm{tr}}$ and $\tau_{\mathrm{te}}$ for the training- and test-time thresholds, respectively.

\begin{figure}[t!]
\centering
\includegraphics[width=\textwidth]{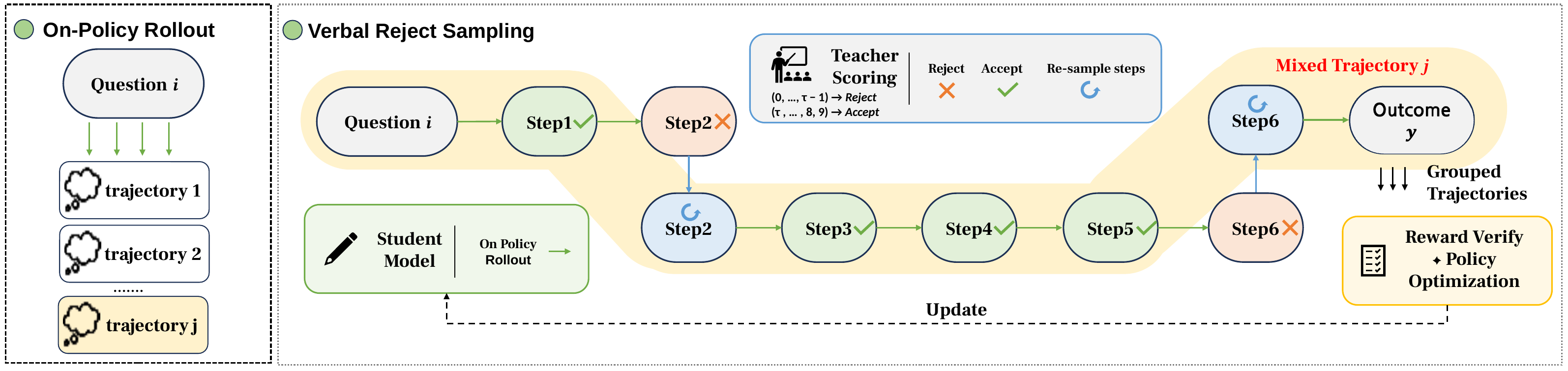}
\caption{Student on-policy rollouts (left); teacher-guided rejection sampling and simulated search interaction for Web Q\&A (right).}
\label{fig:base}
\vspace{-5mm}

\end{figure}

\subsection{On-policy Rollouts}
\label{sec:verbal_scoring}

\paragraph{Rollout Generation.}
The student explores reasoning paths by sampling candidate sub-trajectories from its current policy. After a teacher replacement, the student resumes generation conditioned on the updated prefix.
At each iteration, we sample a batch of $B$ problems $\{q_i\}_{i=1}^B$ and generate $N$ on-policy rollouts for each problem, yielding $B \times N$ trajectories in total. Here $i$ indexes problems in the batch and $j$ indexes the $N$ rollouts for each $q_i$:
$y_i^{(j)} \sim \pi_S(\cdot \mid q_i)$ for $j = 1, \ldots, N$.
Each trajectory $y_i^{(j)} = \bigl(s_{1}^{(i,j)}, \ldots, s_{T_{ij}}^{(i,j)}\bigr)$ has variable-length reasoning steps.
Step boundaries are task-specific. In Web Q\&A, the \texttt{<search>}, \texttt{<information>}, and \texttt{<think>} markers provide explicit boundaries (Appendix~\ref{sec:algorithm_details}). Mathematical rollouts have no such markers. We therefore compute the student's next-token entropy along each rollout and use the maximum-entropy position, which often marks a branch in the reasoning path, as the split point; the suffix beginning at that position is treated as a sub-trajectory for verbal rejection sampling.
On-policy rollouts let the teacher judge sub-trajectories generated by the current student policy, avoiding the distribution mismatch that arises when training on pre-generated teacher outputs.

\vspace{-2mm}
\paragraph{Verbal Rejection Sampling.}

We use the verbal scoring interface in Section~\ref{subsec:verbal_distribution_approximation} to score student sub-trajectories. If a sub-trajectory scores below the training threshold $\tau_{\mathrm{tr}}$, the teacher resamples and replaces that sub-trajectory. After the teacher's replacement is inserted, the student generates the remaining trajectory from its end. This produces \emph{mixed} trajectories that interleave student-authored prefixes with sub-trajectories resampled by the teacher. Because the teacher replaces only low-scoring sub-trajectories, the training trajectories remain largely grounded in the student's on-policy exploration. The mixed trajectories are used to train the student with outcome-based rewards.

\subsection{Distribution Approximation via Verbal Rejection Sampling}
\label{sec:vrs_distribution_approximation}

\paragraph{Constructing the Mixed Training Distribution.}
\emph{Classical rejection sampling}~\citep{devroye1986nonuniform} transforms a proposal distribution into a target distribution by accepting proposals according to a suitably bounded target-to-proposal density ratio. Related methods adapt this distribution-shaping principle to different learning objectives. Density-ratio-based methods correct off-policy distribution mismatch using explicit policy ratios~\citep{chung2018rejection,schlegel2019importance}. Reward-model-based statistical rejection sampling selects complete responses for preference optimization~\citep{liu2024rspo}, while reward-based reasoning methods filter high-reward samples~\citep{yuan2025minimalist}. \emph{Verbal rejection sampling} uses teacher-assigned verbal scores to reweight the student's on-policy trajectory distribution toward a target $p_T(y\mid q)$ that reflects the teacher's preferences over trajectories. For a fixed teacher and threshold $\tau$, we define the acceptance probability as
\begin{equation}
a_\tau(y\mid q)
=
\Pr\!\left[S(y,q)\geq\tau\mid y,q\right],
\label{eq:verbal_acceptance_probability}
\end{equation}
where $S(y,q)$ is the teacher-assigned verbal score and $\Pr$ denotes probability over randomness in teacher scoring. With deterministic scoring, this reduces to $a_\tau(y\mid q)=\mathbf{1}[S(y,q)\geq\tau]$. Conditioning student proposals on acceptance induces the verbally filtered trajectory distribution
\begin{equation}
p_{\mathrm a,\tau}(y\mid q)
=
\frac{\pi_S(y\mid q)\,a_\tau(y\mid q)}
{Z_\tau(q)},
\qquad
Z_\tau(q)
=
\mathbb E_{y'\sim\pi_S(\cdot\mid q)}
\!\left[a_\tau(y'\mid q)\right],
\label{eq:vrs_distribution}
\end{equation}
where $Z_\tau(q)>0$ is the acceptance rate and normalizing constant. Equation~\eqref{eq:vrs_distribution} characterizes the accepted student component of the training distribution. It follows directly from conditioning on acceptance and holds for any acceptance rule with a nonzero acceptance rate. It captures the same distribution-reweighting principle as classical rejection sampling. Classical rejection sampling chooses acceptance probabilities from a target-to-proposal density ratio, whereas our method obtains them from thresholded verbal scores.

To include teacher resampling, fix a problem $q$ and consider a candidate reasoning step $s_t$ at trajectory prefix $y_{<t}$. We suppress the fixed $q$ in the local conditionals. At this level, $a_\tau(s_t\mid y_{<t})$ is the probability of accepting the candidate, and $Z_\tau(y_{<t})=\mathbb E_{s_t\sim\pi_S(\cdot\mid y_{<t})}[a_\tau(s_t\mid y_{<t})]$ is its average over student proposals. Let $r_{\mathcal E,\tau}(s_t\mid y_{<t})$ denote the normalized distribution of teacher replacements conditional on rejection, averaging over rejected candidates and feedback when the replacement depends on them. The reasoning step inserted at this prefix follows
\begin{equation}
p_{\mathrm m,\tau}(s_t\mid y_{<t})=\underbrace{\pi_S(s_t\mid y_{<t})\,a_\tau(s_t\mid y_{<t})}_{\text{retained student exploration}}+\underbrace{\bigl(1-Z_\tau(y_{<t})\bigr)\,r_{\mathcal E,\tau}(s_t\mid y_{<t})}_{\text{teacher guidance}}.
\label{eq:vrs_mixed_subtrajectory}
\end{equation}
The student proposal $\pi_S(s_t\mid y_{<t})$ drives exploration by generating candidate reasoning segments. The first term retains accepted student proposals and has total mass $Z_\tau(y_{<t})$, while the second introduces teacher guidance through resampled replacements and has mass $1-Z_\tau(y_{<t})$. Rejected student proposals also arise from exploration, but are replaced in the training trajectory. After each replacement, the student resumes exploration from the updated prefix. Repeatedly applying the retain-and-replace procedure in Eq.~\eqref{eq:vrs_mixed_subtrajectory} induces a distribution over complete mixed trajectories, denoted by $p_{\mathrm m,\tau}(y\mid q)$, with $\tau=\tau_{\mathrm{tr}}$ during training. Each policy token inherits the source of its sub-trajectory. We record this source using binary indicators $m_\ell^S$ and $m_\ell^T$, which identify tokens in retained student segments and teacher-resampled segments, respectively. These indicators are used to decompose the policy optimization objective in Section~\ref{sec:optimization_objective}.

\vspace{-2mm}
\paragraph{Density-Ratio Condition.}
To analyze when verbal rejection sampling follows the acceptance rule of classical rejection sampling, we consider the following density-ratio condition on the probability that a trajectory's verbal score meets the threshold:
\begin{equation}
a_\tau(y\mid q) \propto \frac{p_T(y\mid q)}{\pi_S(y\mid q)}
\label{eq:verbal_density_ratio_calibration}
\end{equation}
\begin{tcolorbox}[
  colback=callouttitle,
  colframe=calloutframe,
  boxrule=0.6pt,
  arc=1pt,
  boxsep=3pt,
  left=4pt,right=4pt,top=3pt,bottom=3pt,
  before skip=5pt,
  after skip=5pt,
]
\textbf{Takeaway.} \textit{Thresholding verbal scores at $\tau$ converts their ranking into acceptance. The mixture approaches the target when acceptance tracks the target-to-student density ratio and the replacement distribution is close to the target.}
\end{tcolorbox}
\vspace{-0.3\baselineskip}
Under the exact proportionality condition in Eq.~\eqref{eq:verbal_density_ratio_calibration}, the accepted student distribution matches $p_T$. For the mixed training distribution to also match $p_T$, teacher replacements must follow the corresponding target distribution at each prefix where replacement occurs. In practice, approximation error relative to this ideal case arises from two sources: verbal-score acceptance probabilities may deviate from the density ratio, and teacher replacements may deviate from the target.

\Needspace{6cm}
\begin{wrapfigure}{r}{0.50\textwidth}
\vspace{-5mm}
\centering
\begin{minipage}{\linewidth}
\centering
\begin{tikzpicture}
\begin{axis}[width=\linewidth,height=3.6cm,scale only axis=false,
ylabel={Teacher acceptance rate},ymin=0,ymax=1,
axis lines=left,ymajorgrids=true,grid style={gray!20},
tick label style={font=\scriptsize},label style={font=\scriptsize},
legend style={font=\scriptsize,draw=none,at={(0.02,0.98)},anchor=north west},
xlabel={Centered $\log(\pi_{\mathcal E}/\pi_S)$}]
\addplot+[color=blue!75!black,mark=*,mark size=1.5pt,line width=0.8pt,error bars/.cd,y dir=both,y explicit] table[x=x,y=y,y error plus=plus,y error minus=minus,row sep=\\]{
x y plus minus\\
-1016.389267 0.589063 0.044176 0.045874\\
-316.084502 0.637500 0.042113 0.041269\\
-100.698856 0.785937 0.034699 0.033251\\
18.452301 0.843750 0.032008 0.033410\\
88.514558 0.870313 0.028521 0.032012\\
140.438597 0.878125 0.031815 0.033972\\
185.693899 0.873437 0.031113 0.032803\\
237.850273 0.846875 0.034900 0.036030\\
305.836314 0.857812 0.028698 0.030889\\
456.386682 0.857812 0.034933 0.035077\\
};
\addlegendentry{$\tau=3$}
\addplot+[color=orange!85!black,mark=square*,mark size=1.5pt,line width=0.8pt,error bars/.cd,y dir=both,y explicit] table[x=x,y=y,y error plus=plus,y error minus=minus,row sep=\\]{
x y plus minus\\
-1016.389267 0.579688 0.044156 0.045637\\
-316.084502 0.634375 0.041662 0.042137\\
-100.698856 0.779687 0.035076 0.034404\\
18.452301 0.834375 0.032104 0.035044\\
88.514558 0.867188 0.028045 0.031128\\
140.438597 0.875000 0.032321 0.035574\\
185.693899 0.870313 0.031181 0.032611\\
237.850273 0.846875 0.034900 0.036030\\
305.836314 0.857812 0.028698 0.030889\\
456.386682 0.856250 0.034469 0.034824\\
};
\addlegendentry{$\tau=5$}
\addplot+[color=teal!80!black,mark=triangle*,mark size=1.5pt,line width=0.8pt,error bars/.cd,y dir=both,y explicit] table[x=x,y=y,y error plus=plus,y error minus=minus,row sep=\\]{
x y plus minus\\
-1016.389267 0.135937 0.027140 0.027541\\
-316.084502 0.242188 0.033420 0.032312\\
-100.698856 0.337500 0.044597 0.043102\\
18.452301 0.442188 0.066160 0.066361\\
88.514558 0.506250 0.059045 0.058590\\
140.438597 0.562500 0.060406 0.064073\\
185.693899 0.523438 0.056030 0.055600\\
237.850273 0.548438 0.056472 0.054933\\
305.836314 0.545312 0.051059 0.054025\\
456.386682 0.545312 0.055683 0.055345\\
};
\addlegendentry{$\tau=7$}
\end{axis}
\end{tikzpicture}
\end{minipage}
\captionsetup{font=small,skip=2pt}
\caption{Verbal acceptance versus teacher-to-student log density ratio on 6,400 Math500 answers. Points show within-question-centered bin estimates; bars indicate 95\% question-bootstrap intervals.}
\label{fig:verbal_acceptance_likelihood}
\vspace{-2mm}
\end{wrapfigure}

The approximation question thus concerns both the accepted student component and the teacher replacements. The threshold controls their relative contributions: at prefix $y_{<t}$, teacher replacement is invoked with probability $1-Z_\tau(y_{<t})$. Appendix~\ref{sec:theory} formalizes an \emph{Approximate Density-Ratio Condition} and gives error bounds for the accepted and mixed distributions in Proposition~\ref{prop:approx_density_ratio}. Figure~\ref{fig:verbal_acceptance_likelihood} shows that verbal acceptance generally increases with the teacher-to-student log density ratio before leveling off, supporting the distribution-reweighting rationale of our analysis.

\WFclear
\vspace{-3mm}
\subsection{Optimization Objective}
\vspace{-2mm}
\label{sec:optimization_objective}
During training, the retain-and-replace procedure produces mixed trajectories, whose outcome rewards determine group-relative advantages. These advantages are applied to both student-generated and teacher-resampled tokens: verbal rejection sampling shapes the training data, while outcome-based advantages determine the direction of policy optimization. The complete \textsc{OVD} training pipeline is illustrated in Figure~\ref{fig:base}.
\paragraph{Reward Design.}
For trajectories, we compute the \emph{outcome-based reward} $R(y)$ as:
\begin{equation}
R(y)=\mathrm{F}_1(y,y^*)\;\text{(Web Q\&A)},
\qquad R(y)=\delta(y,y^*)\;\text{(math)}.
\end{equation}
where \(y^*\) denotes the ground-truth reference answer, $\mathrm{F}_1(y, y^*) = \frac{2|y \cap y^*|}{|y| + |y^*|}$ measures word overlap, and $\delta(y, y^*) \in \{0, 1\}$ denotes exact matching under mathematical equivalence (numerical tolerance, algebraic normalization). This design provides soft partial credit for open-ended answers and strict binary feedback for correctness.

\textbf{Policy Gradient Objective.}
For each problem $q_i$, we sample a group of complete trajectories $y_i^{(j)}\sim p_{\mathrm m,\tau_{\mathrm{tr}}}(\cdot\mid q_i)$, $j=1,\ldots,N$, using the retain-and-replace procedure in Section~\ref{sec:vrs_distribution_approximation}. For each training batch, this sampling distribution is constructed using the student policy before the update, $\pi_S^{\mathrm{old}}$; the sampled trajectories and their token-source indicators are held fixed during optimization. Inspired by GRPO~\citep{shao2024deepseekmath}, we compute group-relative advantages and student-policy probability ratios on mixed trajectories:

\begingroup
\setlength{\abovedisplayskip}{4pt}
\setlength{\belowdisplayskip}{4pt}
\setlength{\abovedisplayshortskip}{2pt}
\setlength{\belowdisplayshortskip}{4pt}
\noindent
\begin{minipage}[t]{0.43\linewidth}
\begin{equation}
A(y_i^{(j)})=\frac{R(y_i^{(j)})-\mu_i}{\sigma_i+\epsilon},
\end{equation}
\end{minipage}\hfill
\begin{minipage}[t]{0.55\linewidth}
\begin{equation}
\rho_\ell
=\frac{\pi_S(x_\ell\mid q,x_{<\ell})}
{\pi_S^{\mathrm{old}}(x_\ell\mid q,x_{<\ell})}.
\label{eq:mixed_token_ratio}
\end{equation}
\end{minipage}
\par\endgroup

\noindent Here, $\mu_i$ and $\sigma_i$ are the mean and standard deviation of outcome rewards within the mixed-trajectory group for $q_i$. We recompute old-student log probabilities on the complete mixed trajectories and hold them fixed during optimization. Both expressions apply to student-generated and teacher-resampled tokens.
For \emph{Web Q\&A}, tool-returned \emph{information} spans remain masked out of the policy-gradient and KL terms. The update applies to search-query and reasoning tokens, including teacher-resampled replacements, rather than environment observations. At every unmasked position, the source indicators satisfy $m_\ell^S+m_\ell^T=1$.

\begingroup
\setlength{\abovedisplayskip}{4pt}
\setlength{\belowdisplayskip}{4pt}
\setlength{\abovedisplayshortskip}{2pt}
\setlength{\belowdisplayshortskip}{4pt}
Trajectories whose sub-trajectories are all accepted remain entirely student-generated. They contribute to policy optimization without matching the teacher's token-level probability distribution, preserving room for student exploration. Teacher-resampled segments provide guidance under the same outcome-based advantages. To distinguish these contributions, define the clipped token objective as
\begin{equation}
g_\ell(y)=\min\!\left(
\rho_\ell A(y),
\operatorname{clip}(\rho_\ell,1-\epsilon_c,1+\epsilon_c)A(y)
\right),
\end{equation}
where we use PPO-style clipping~\citep{schulman2017proximal} with $\epsilon_c=0.2$. We optimize over these samples from $p_{\mathrm m,\tau_{\mathrm{tr}}}$, writing $\mathbb E_{\mathrm m}$ for the expectation over training problems, sampled trajectory groups, their recorded token sources, and unmasked token positions. The objective decomposes by token source as
\begin{equation}
\mathcal L_{\mathrm{RL}}=\underbrace{-\mathbb E_{\mathrm m}\!\left[m_\ell^S g_\ell(y)\right]}_{\text{learning from student exploration}}+\underbrace{-\mathbb E_{\mathrm m}\!\left[m_\ell^T g_\ell(y)\right]}_{\text{learning from teacher guidance}}.
\label{eq:grpo_source_decomposition}
\end{equation}
Both terms use the same sampling distribution and normalization, so their sum equals $-\mathbb E_{\mathrm m}[g_\ell(y)]$. For entirely student-generated trajectories, the teacher-guidance term is zero. Positive advantages reinforce tokens from either source, while negative advantages suppress them. Thus, the student learns from its own exploration and from teacher guidance through a shared outcome-based objective.
\par\endgroup

\begin{table*}[!t]
\centering
\caption{Web Q\&A results with Qwen-2.5-3B-Base as the student and Qwen2.5-7B as the teacher. \textsc{OVD} rows give $(\tau_{\mathrm{tr}},\tau_{\mathrm{te}})$. Best results are in bold. $\dagger$: score-vocabulary sampling at test time; $\ddagger$: process reward model (SmartSearch and StepSearch). $^\ast$: OPD baselines using either full-vocabulary reverse KL in Eq.~\eqref{eq:opd_reverse_kl} or its top-$k$ approximation described in Section~\ref{subsec:token_opd_limitations}.}
\label{tab:qa_em_results}
\footnotesize
\setlength{\tabcolsep}{2pt}
\begin{adjustbox}{max width=\textwidth}
\begin{tabular}{lccccccccc}
\toprule
& \multicolumn{3}{c}{\textbf{Single-Hop QA}} & \multicolumn{5}{c}{\textbf{Multi-Hop QA}} & \\
\cmidrule(lr){2-4}\cmidrule(lr){5-9}
\textbf{Method} & \textbf{NQ} & \textbf{TriviaQA} & \textbf{PopQA} & \textbf{HotpotQA} & \textbf{2Wiki} & \textbf{Musique} & \textbf{Bamboogle} & \textbf{GAIA} & \textbf{Avg.} \\
\midrule
\multicolumn{10}{c}{\cellcolor[HTML]{E5E5FC}\textbf{\textit{Qwen-2.5-3B-Base}}} \\
\midrule
RAG & 31.60 & 58.00 & 15.20 & 24.20 & 23.20 & 8.20 & 15.20 & 5.40 & 22.63 \\
RA-Agent & 15.20 & 28.40 & 6.60 & 12.60 & 16.60 & 2.60 & 13.60 & 3.00 & 12.33 \\
Search-o1 & 16.60 & 31.00 & 8.20 & 14.80 & 22.40 & 5.20 & 22.40 & 2.40 & 15.38 \\
R1 & 14.20 & 34.80 & 20.80 & 19.60 & 28.40 & 6.40 & 5.56 & 1.80 & 16.45 \\
Search-R1 & 40.60 & 60.00 & 44.20 & 29.20 & 32.00 & 11.20 & 12.50 & 7.81 & 29.69 \\
ZeroSearch & 43.00 & 61.60 & 41.40 & 33.80 & 34.60 & 13.00 & 13.89 & 7.27 & 31.07 \\
OPD$^\ast$ (reverse KL) & 14.40 & 25.40 & 13.40 & 13.20 & 21.60 & 2.20 & 8.80 & 2.90 & 14.14 \\
OPD$^\ast$ (top-$k$ logits) & 9.80 & 20.60 & 11.20 & 12.00 & 17.00 & 3.20 & 7.20 & 2.50 & 11.51 \\
\midrule
SmartSearch$^\ddagger$ & 39.80 & 64.40 & 41.20 & 38.40 & \textbf{45.40} & 15.20 & \textbf{28.80} & 9.09 & 35.20 \\
StepSearch$^\ddagger$ & 16.20 & 28.00 & 20.40 & 18.20 & 16.20 & 3.60 & 17.60 & 7.27 & 18.60 \\
\midrule
\textsc{OVD} ($\tau_{\mathrm{tr}}{=}5$, $\tau_{\mathrm{te}}{=}0$) & \textbf{49.20} & 67.20 & 63.20 & 39.20 & 35.20 & 21.80 & 15.20 & 9.09 & 37.51 \\
\textsc{OVD} ($\tau_{\mathrm{tr}}{=}5$, $\tau_{\mathrm{te}}{=}5$) & 47.00 & 67.40 & 66.40 & 38.00 & 37.60 & 23.20 & 24.80 & 7.88 & 39.04 \\
\textsc{OVD}$^\dagger$ ($\tau_{\mathrm{tr}}{=}5$, $\tau_{\mathrm{te}}{=}5$) & 48.80 & \textbf{68.00} & \textbf{66.60} & \textbf{39.40} & 41.40 & \textbf{24.20} & \textbf{28.80} & \textbf{11.50} & \textbf{41.09} \\
\bottomrule
\end{tabular}
\end{adjustbox}
\end{table*}

\vspace{-0.35\baselineskip}

\vspace{-2mm}
\section{Experiments}
\label{sec:experiments}

\vspace{-3mm}
\subsection{Evaluation Tasks}
\label{subsec:evaluation_tasks}
\vspace{-2mm}
\noindent We evaluate \textsc{OVD} in two complementary settings: Web Q\&A tasks in an agentic setting, where trajectories decompose into multiple search and chain-of-thought (CoT) steps, and mathematical reasoning tasks that provide a tool-free setting for isolating the effects of tool use on distillation.
\par\penalty0\vspace{4pt}

\noindent\textbf{Web Q\&A Tasks.}
We evaluate on eight question-answering benchmarks spanning both single-hop and multi-hop reasoning: NQ~\citep{kwiatkowski2019natural}, TriviaQA~\citep{joshi2017triviaqa}, PopQA~\citep{mallen2023popqa}, HotpotQA~\citep{yang2018hotpotqa}, 2Wiki~\citep{ho2020constructing}, Musique~\citep{trivedi2022musique}, Bamboogle~\citep{press2023measuring}, and GAIA. GAIA is the most challenging benchmark in our evaluation suite, testing general AI assistants on real-world questions that require combinations of reasoning, multimodal understanding, web browsing, and tool use~\citep{mialon2023gaia}.

\par\medskip
\vspace{-3mm}
\noindent\textbf{Mathematical Reasoning Tasks.}
We evaluate mathematical reasoning on the nine benchmarks in Table~\ref{tab:fixed_nine_math}; dataset descriptions and sources are provided in Appendix~\ref{sec:math_dataset_descriptions}.
The DeepScaleR-Preview-Dataset contains approximately 40{,}000 mathematics problems~\citep{deepscaler2025}. We use one fixed problem or a fixed set of 128 distinct problems from this dataset for the two training regimes, respectively. The 128-problem set also underlies the prefix-selection analysis in Figure~\ref{fig:vrs_precision_cost_frontier}.
For these tasks, we use \path{deepseek-ai/DeepSeek-R1-Distill-Qwen-7B}~\citep{deepseek2025r1} as the teacher and Qwen2.5-Math-1.5B as the student. The teacher assigns each reasoning sub-trajectory a single quality score, thereby ranking candidates from incorrect or irrelevant to optimal steps toward the solution; Appendix~\ref{sec:hyperparameters} gives the remaining implementation details.

\vspace{-2mm}
\subsection{Agentic Search Setting}
\vspace{-2mm}
\label{subsec:main_results_qa}

\begingroup
\setlength{\intextsep}{4pt}
\setlength{\columnsep}{6pt}
\newsavebox{\ovdRejectionTableBox}
\sbox{\ovdRejectionTableBox}{%
\small
\setlength{\tabcolsep}{1pt}%
\begin{tabular}{@{}lccc@{}}
\toprule
\rowcolor[HTML]{E5E5FC}
Sampling Target & Single-Hop & Multi-Hop & Overall \\
\midrule
Query Rejection & \textbf{59.87} & 24.10 & 37.51 \\
Query+CoT Rejection & 54.27 & 26.53 & 36.93 \\
CoT-only Rejection & 55.93 & \textbf{27.02} & \textbf{37.87} \\
\bottomrule
\end{tabular}%
}
\leavevmode
\begin{wraptable}[8]{r}{\wd\ovdRejectionTableBox}
\raggedright
\vspace{-5mm}
\begin{minipage}[t]{\wd\ovdRejectionTableBox}
\captionsetup{font=small,skip=2pt,width=\linewidth}
\captionof{table}{Effect of the part of the response rewritten during training on Web Q\&A. Results are macro-average exact match (\%) with real search and $(\tau_{\mathrm{tr}},\tau_{\mathrm{te}})=(5,0)$. No teacher is used at test time.}
\label{tab:rejection_granularity}
\noindent\usebox{\ovdRejectionTableBox}
\end{minipage}
\par
\end{wraptable}

\noindent\textit{(i) Comparison with token-level OPD.} Table~\ref{tab:qa_em_results} compares \textsc{OVD} with full-vocabulary reverse-KL and top-$k$ logit OPD. Without teacher intervention at test time, \textsc{OVD} achieves 37.51\% average EM, compared with 14.14\% and 11.51\%, respectively, and outperforms both baselines on all eight benchmarks. These results support the effectiveness of verbal guidance over token-level probability matching in this agentic setting.
\textit{(ii) Overall performance.} Table~\ref{tab:qa_em_results} reports Web Q\&A exact match (EM) on eight test sets for Qwen-2.5-3B-Base. \textsc{OVD} outperforms Search-o1, Search-R1, ZeroSearch, SmartSearch, and StepSearch in average EM. \textit{(iii) Test-time score sampling.} The $\dagger$ variant samples the \emph{test-time} score from the teacher's ten-token score distribution at temperature 0.7 instead of selecting the highest-probability score token. Under matched $(\tau_{\mathrm{tr}},\tau_{\mathrm{te}})$, \emph{test-time} sampling improves average EM from 39.04\% to 41.09\%. The teacher's most likely score need not reflect a confident judgment. Test-time sampling accounts for this ambiguity when deciding which student continuations to retain.
\par
\endgroup
\begin{table*}[!t]
\centering
\caption{Math accuracy after training on one fixed problem or 128 distinct problems. FR and HES use full-response and high-entropy-suffix resampling. Avg. is the macro average over the nine benchmarks; bold marks each method's highest Avg. Benchmark selection and random-suffix results are given in Appendix~\ref{sec:additional_math_results}.}
\vspace{-2mm}
\label{tab:fixed_nine_math}
\scriptsize
\setlength{\tabcolsep}{2.2pt}
\resizebox{\textwidth}{!}{%
\begin{tabular}{lccccccccccc}
\toprule
\textbf{Method} & \textbf{Steps} & \textbf{AIME24} & \textbf{AMC23} & \textbf{CARP} & \textbf{Middle} & \textbf{College} & \textbf{Gaokao-II} & \textbf{MATH} & \textbf{MAWPS} & \textbf{SAT-Math} & \textbf{Avg.} \\
\midrule
Initial Student & 0 & 13.3 & 35.0 & 45.3 & 30.7 & 5.2 & 35.7 & 34.3 & 22.5 & 78.1 & 33.36 \\
\midrule
\multicolumn{12}{l}{\textit{Training on One Fixed Problem}} \\
RLVR & 300 & 6.7 & 50.0 & 51.3 & 42.6 & 33.0 & 50.0 & 66.0 & 91.3 & 81.3 & 52.46 \\
RLVR & 500 & 13.3 & 47.5 & 51.7 & 40.6 & 34.7 & 42.9 & 66.3 & 94.1 & 81.3 & 52.49 \\
RLVR & 600 & 10.0 & 50.0 & 51.7 & 39.6 & 34.5 & 50.0 & 66.9 & 94.1 & 81.3 & 53.13 \\
RLVR & 800 & 10.0 & 52.5 & 51.0 & 39.6 & 36.3 & 57.1 & 67.3 & 95.2 & 78.1 & \textbf{54.14} \\
\midrule
\textsc{OVD}-FR & 300 & 6.7 & 50.0 & 50.3 & 42.6 & 30.6 & 57.1 & 65.8 & 87.4 & 81.3 & 52.42 \\
\textsc{OVD}-FR & 500 & 10.0 & 52.5 & 50.5 & 45.5 & 34.6 & 50.0 & 66.4 & 93.8 & 84.4 & 54.19 \\
\textsc{OVD}-FR & 600 & 10.0 & 55.0 & 51.0 & 42.6 & 36.5 & 50.0 & 67.6 & 94.9 & 90.6 & \textbf{55.36} \\
\textsc{OVD}-FR & 800 & 13.3 & 60.0 & 51.8 & 40.6 & 37.4 & 42.9 & 68.2 & 95.2 & 84.4 & 54.87 \\
\midrule
\textsc{OVD}-HES & 300 & 6.7 & 50.0 & 51.2 & 38.6 & 34.7 & 50.0 & 66.5 & 93.4 & 81.3 & 52.47 \\
\textsc{OVD}-HES & 500 & 10.0 & 50.0 & 50.6 & 42.6 & 36.7 & 42.9 & 67.3 & 94.7 & 84.4 & 53.24 \\
\textsc{OVD}-HES & 600 & 10.0 & 50.0 & 51.7 & 40.6 & 37.6 & 42.9 & 67.6 & 95.3 & 84.4 & 53.33 \\
\textsc{OVD}-HES & 800 & 13.3 & 50.0 & 51.5 & 43.6 & 38.1 & 57.1 & 69.1 & 96.0 & 81.3 & \textbf{55.56} \\
\midrule
\multicolumn{12}{l}{\textit{Training on 128 Distinct Problems}} \\
RLVR & 300 & 20.0 & 52.5 & 52.6 & 50.5 & 38.1 & 50.0 & 69.2 & 96.1 & 71.9 & \textbf{55.65} \\
RLVR & 500 & 20.0 & 55.0 & 53.6 & 47.5 & 39.2 & 28.6 & 70.1 & 96.4 & 75.0 & 53.93 \\
RLVR & 600 & 16.7 & 52.5 & 54.2 & 52.5 & 39.3 & 35.7 & 70.6 & 96.8 & 78.1 & 55.15 \\
RLVR & 800 & 16.7 & 47.5 & 54.9 & 48.5 & 39.3 & 42.9 & 71.1 & 96.8 & 78.1 & 55.08 \\
\midrule
\textsc{OVD}-FR & 300 & 16.7 & 55.0 & 53.0 & 50.5 & 37.1 & 42.9 & 68.8 & 95.3 & 71.9 & 54.57 \\
\textsc{OVD}-FR & 500 & 13.3 & 52.5 & 54.1 & 49.5 & 39.0 & 50.0 & 69.8 & 96.1 & 71.9 & \textbf{55.13} \\
\textsc{OVD}-FR & 600 & 16.7 & 62.5 & 54.6 & 44.6 & 39.4 & 50.0 & 70.0 & 96.3 & 59.4 & 54.82 \\
\textsc{OVD}-FR & 800 & 16.7 & 52.5 & 54.4 & 41.6 & 39.0 & 42.9 & 70.8 & 96.6 & 53.1 & 51.95 \\
\midrule
\textsc{OVD}-HES & 300 & 23.3 & 50.0 & 53.7 & 47.5 & 38.4 & 64.3 & 69.1 & 95.3 & 75.0 & \textbf{57.40} \\
\textsc{OVD}-HES & 500 & 13.3 & 55.0 & 54.3 & 49.5 & 39.0 & 35.7 & 70.0 & 96.5 & 75.0 & 54.25 \\
\textsc{OVD}-HES & 600 & 6.7 & 57.5 & 53.9 & 46.5 & 38.7 & 35.7 & 70.6 & 96.5 & 75.0 & 53.45 \\
\textsc{OVD}-HES & 800 & 13.3 & 55.0 & 53.7 & 51.5 & 39.4 & 28.6 & 70.5 & 97.0 & 75.0 & 53.77 \\
\bottomrule
\end{tabular}%
}
\end{table*}

\begin{tcolorbox}[
  colback=callouttitle,
  colframe=calloutframe,
  boxrule=0.6pt,
  arc=1pt,
  boxsep=3pt,
  left=4pt,right=4pt,top=3pt,bottom=3pt,
  before skip=0pt,
  after skip=5pt,
]
\textbf{Findings 4}: \textit{Score sampling improves test-time performance by preserving teacher uncertainty}
\end{tcolorbox}

\subsection{Where Should Rejection Sampling Be Applied?}
\vspace{-2mm}
\label{sec:rejection_scope}

The preferred rejection target depends on the task. In this agentic setting, trajectories comprise search-query and CoT sub-trajectories, and Table~\ref{tab:rejection_granularity} compares applying rejection sampling to queries, CoT, or both during training. At fixed thresholds, query-only rejection performs best on single-hop questions, whereas CoT-only rejection achieves the highest multi-hop and overall EM. Rewriting both queries and reasoning steps offers no advantage over these targeted alternatives. For our search-focused Web Q\&A experiments, we use query-only rejection as the default; CoT distillation is examined separately in the mathematical-reasoning experiments.

\par\noindent\begin{minipage}{\linewidth}
\subsection{Can We Approximate the Teacher Distribution at Test Time?}
\label{sec:ablation}
\leavevmode
\begingroup
\setlength{\intextsep}{2pt}
\begin{wrapfigure}[13]{r}{0.54\textwidth}
    \centering
    \includegraphics[width=\linewidth]{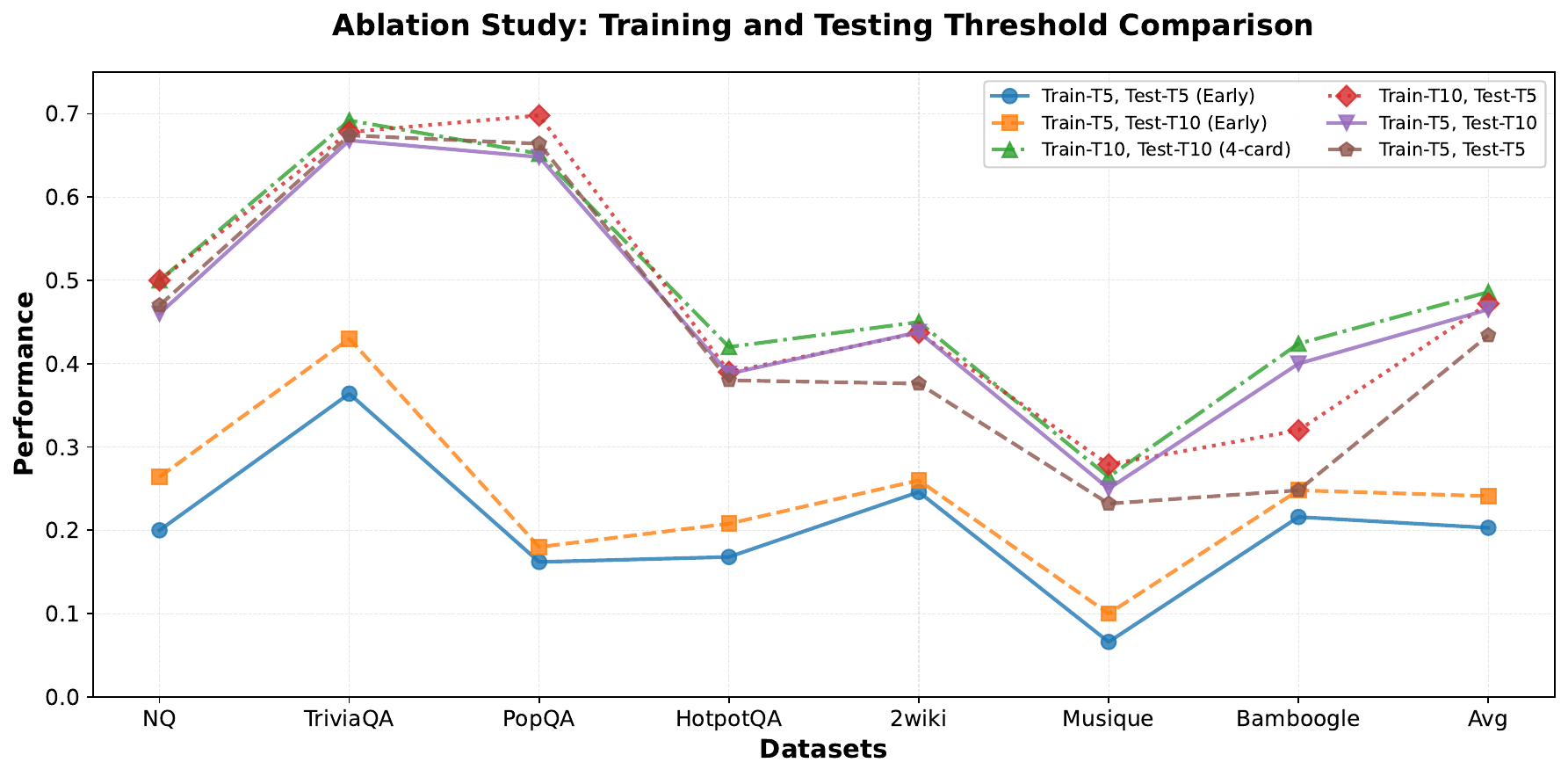}
    \captionsetup{font=small,skip=2pt,justification=justified,singlelinecheck=false}
    \caption{Web Q\&A performance under different training- and test-time rejection thresholds. Results are reported for each dataset and as a macro-average (Avg.).}
    \label{fig:ablation_study}
\end{wrapfigure}

At test time, setting $\tau=\tau_{\mathrm{te}}$ in Eq.~\eqref{eq:vrs_mixed_subtrajectory} determines the acceptance rate $Z_{\tau_{\mathrm{te}}}$: increasing $\tau_{\mathrm{te}}$ retains fewer student sub-trajectories and invokes the teacher more often. Thus, $\tau_{\mathrm{te}}$ controls the strength of test-time selection, trading off final-answer quality against teacher intervention. We compare teacher models with different output distributions: \textit{(i)} a prompt-based teacher agent, configured through prompting without task-specific fine-tuning (dashed curves in Fig.~\ref{fig:ablation_study}), with $\tau_{\mathrm{tr}}=5$ and $\tau_{\mathrm{te}}\in\{0,5,10\}$; and \textit{(ii)} an SFT-based teacher agent, fine-tuned on simulated-search trajectories (solid curves in Fig.~\ref{fig:ablation_study}), with $\tau_{\mathrm{tr}}\in\{5,10\}$ and $\tau_{\mathrm{te}}\in\{5,10\}$.
\par
\endgroup
\end{minipage}\par

At test time, a moderate threshold of $\tau_{\mathrm{te}}=5$ outperforms both $\tau_{\mathrm{te}}=0$ (no teacher intervention) and $\tau_{\mathrm{te}}=10$ (replacing every scored sub-trajectory). These results suggest that selective teacher guidance during inference is more effective than either accepting all student continuations or replacing them all. \textit{(iii)} Fully replacing the student policy trajectories with those from the teacher during training (i.e., using $\tau_{\mathrm{tr}}=10$) does not necessarily yield better gains, as it shifts the mixed training distribution almost entirely toward teacher-generated trajectories, replacing even potentially correct student trajectories and weakening useful exploration.

\WFclear

\begin{figure}[!t]
\centering
\definecolor{ovdplotblue}{HTML}{1F77B4}
\definecolor{ovdplotorange}{HTML}{FF7F0E}
\definecolor{ovdplotgreen}{HTML}{2CA02C}
\pgfplotsset{
  ovd paired plot/.style={
    width=0.94\linewidth,
    height=3.8cm,
    axis lines=box,
    axis line style={draw=black!75,line width=.45pt},
    grid=major,
    major grid style={dashed,line width=.3pt,draw=gray!30},
    tick label style={font=\tiny\sffamily},
    label style={font=\scriptsize\sffamily\bfseries},
    legend style={
      font=\tiny\sffamily,
      at={(0.5,1.03)},
      anchor=south,
      legend columns=3,
      draw=gray!55,
      fill=white,
      fill opacity=.92,
      text opacity=1,
      rounded corners=1pt,
      inner sep=1.5pt,
      /tikz/every even column/.append style={column sep=0.4em},
    },
    clip=false,
  },
  ovd hes curve/.style={ovdplotblue,line width=1pt,mark=*,mark size=1.8pt},
  ovd fr curve/.style={ovdplotorange,line width=1pt,dashed,mark=square*,mark size=1.7pt},
  ovd rs curve/.style={ovdplotgreen,line width=1pt,dashdotted,mark=triangle*,mark size=1.8pt},
}
\captionsetup{font=small,skip=2pt,justification=justified,singlelinecheck=false}
\begin{minipage}[t]{0.485\textwidth}
\vspace{0pt}
\centering
\begin{tikzpicture}
\begin{axis}[
    ovd paired plot,
    xmin=260,
    xmax=900,
    ymin=74.5,
    ymax=79.1,
    xtick={300,500,600,800},
    ytick={75,76,77,78,79},
    xlabel={Training steps},
    ylabel={Avg. P@16 (\%)},
]
\addplot[ovd hes curve] coordinates {
    (300,77.84) (500,77.22) (600,77.96) (800,78.50)
};
\addlegendentry{\textsc{OVD}-HES}
\addplot[ovd fr curve] coordinates {
    (300,77.72) (500,77.20) (600,78.40) (800,75.26)
};
\addlegendentry{\textsc{OVD}-FR}
\addplot[ovd rs curve] coordinates {
    (300,78.06) (500,77.20) (600,76.18) (800,76.14)
};
\addlegendentry{\textsc{OVD}-RS}
\draw[<->,semithick,gray!75!black]
    (axis cs:835,75.26) -- node[midway,right,font=\tiny\sffamily,fill=white,inner sep=0.5pt]{+3.24}
    (axis cs:835,78.50);
\end{axis}
\end{tikzpicture}
\caption{Average P@16 (the percentage of problems with at least one correct answer among 16 sampled responses) on ASDiv~\citep{miao2020diverse}, MAWPS~\citep{koncel2016mawps}, TABMWP, Olympiad, and AIME24. At 800 steps, \textsc{OVD}-HES reaches 78.50\%, versus 76.14\% (\textsc{OVD}-RS) and 75.26\% (\textsc{OVD}-FR).}
\label{fig:hes_sampling_collapse}
\end{minipage}%
\hfill
\vspace{-6mm}
\begin{minipage}[t]{0.485\textwidth}
\vspace{0pt}
\centering
\begin{tikzpicture}
\begin{axis}[
 ovd paired plot,
 xlabel={Number of training problems},
 ylabel={Mean time per step (s)},
 xmin=-5,xmax=134,ymin=130,ymax=430,
 xtick={1,16,32,64,128},ytick={150,200,250,300,350,400},
]
\addplot[ovd hes curve] coordinates
 {(1,177.9272) (16,313.2114) (32,342.5640) (64,352.0541) (128,359.6136)};
\addlegendentry{\textsc{OVD}-HES}
\addplot[ovd fr curve] coordinates
 {(1,169.3922) (16,368.4183) (32,388.6755) (64,402.3124) (128,400.3816)};
\addlegendentry{\textsc{OVD}-FR}
\addplot[ovd rs curve] coordinates
 {(1,167.2586) (16,300.0448) (32,315.2008) (64,332.1538) (128,328.5483)};
\addlegendentry{\textsc{OVD}-RS}
\end{axis}
\end{tikzpicture}
\caption{Mean time per step, averaged over 10 updates on four AMD GPUs (batch size 128; eight rollouts per input). Training problems are repeated to keep batch size fixed; shared resources may add timing noise.}
\label{fig:datasize_training_time}
\end{minipage}
\par
\end{figure}

\vspace{-3mm}
\subsection{Mathematical Reasoning Setting}
\label{subsec:scaling_rl}
\vspace{-0.35\baselineskip}

To study verbal rejection sampling without the confounding effects of tool interaction, we evaluate it on CoT sub-trajectories in tool-free mathematical reasoning.

\vspace{-4mm}
\paragraph{Trajectory-Space Approximation.}
Unlike Web Q\&A, mathematical rollouts have no explicit markers for step boundaries. We use high-entropy token positions as a heuristic for potential branch points in a reasoning path. Accordingly, \textsc{OVD}-HES splits each rollout at the student's maximum-entropy token, retains the preceding prefix, and asks the teacher to resample the suffix if its score is below $\tau_{\mathrm{tr}}$. By contrast, \textsc{OVD}-FR resamples the entire low-scoring trajectory. We evaluate two training regimes across checkpoints from 300 to 800 steps: one fixed problem used throughout optimization and 128 distinct problems. The best HES and RLVR macro averages are 55.56\% and 54.14\% with one problem (both at step 800), and 57.40\% and 55.65\% with 128 problems (both at step 300). Compared with full-response replacement, \textsc{OVD}-HES achieves higher best-checkpoint accuracy in both training settings, suggesting that retaining student-generated prefixes can support effective learning while high-entropy token positions guide teacher suffix replacement. We next compare per-step and cumulative training time to assess the computational cost of this approach.

\vspace{-3mm}
\paragraph{Training Efficiency.}
Figure~\ref{fig:datasize_training_time} reports per-step training costs. As problem diversity grows, suffix resampling becomes faster than \textsc{OVD}-FR. With 128 problems, \textsc{OVD}-HES reduces mean per-step training time by 10.2\% relative to \textsc{OVD}-FR. Over 800 training steps, \textsc{OVD}-HES and \textsc{OVD}-RS reduce cumulative training time by 7.4\% and 30.5\%, respectively, relative to \textsc{OVD}-FR.

\vspace{-3mm}
\paragraph{Entropy Collapse.}
\textit{Does policy optimization concentrate probability on a narrow set of reasoning paths, reducing exploration and solution coverage?}~\citep{yue2025does} We use P@16 as a behavioral indicator of this loss of exploration, not as a direct measure of token entropy. At 800 steps in Figure~\ref{fig:hes_sampling_collapse}, \textsc{OVD}-HES exceeds \textsc{OVD}-FR and \textsc{OVD}-RS by 3.24 and 2.36 points, respectively. This advantage suggests that retaining student-generated prefixes, rather than replacing entire trajectories, mitigates trajectory-level entropy collapse as training progresses.

\begin{tcolorbox}[
  colback=callouttitle,
  colframe=calloutframe,
  boxrule=0.6pt,
  arc=1pt,
  boxsep=3pt,
  left=4pt,right=4pt,top=3pt,bottom=3pt,
  before skip=5pt,
  after skip=5pt,
]
\textbf{Findings 5}: \textit{\textsc{OVD}-HES mitigates trajectory-level entropy collapse by preserving student exploration in the prefix space while repairing weak continuations.}
\end{tcolorbox}


\vspace{-0.35\baselineskip}

\section{Conclusion}
\vspace{-2mm}
\label{sec:conclusion}

\noindent We introduced \emph{On-policy Verbal Distillation} (\textsc{OVD}), a black-box-compatible framework using verbal scores for selective teacher replacement of student sub-trajectories and GRPO for mixed-trajectory training, without dense teacher logits. Under score-calibration and replacement-quality assumptions, we bound accepted and mixed distributions' deviation from a teacher-preferred target. Experiments on agentic Web Q\&A and tool-free mathematics show gains over evaluated baselines, while high-entropy-suffix resampling preserves late-stage solution coverage.

\clearpage

\subsection*{AI Use Statement}
Generative AI tools were used to assist with manuscript drafting and revision,
clarification of theoretical assumptions and claims, interpretation and
presentation of experimental results, checking summary calculations, and
LaTeX formatting and template migration. The authors reviewed and revised the
AI-assisted material and take full responsibility for the final content of this
work, including its text, claims, results, code, and other artifacts.

\begingroup
\sloppy
\bibliographystyle{\OVDPath iclr2027_conference}
\bibliography{\OVDPath ovd_colm}

\begin{thebibliography}{63}
\providecommand{\natexlab}[1]{#1}
\providecommand{\url}[1]{\texttt{#1}}
\expandafter\ifx\csname urlstyle\endcsname\relax
  \providecommand{\doi}[1]{doi: #1}\else
  \providecommand{\doi}{doi: \begingroup \urlstyle{rm}\Url}\fi

\bibitem[Agarwal et~al.(2024)Agarwal, Vieillard, Zhou, Stanczyk, Ramos~Garea,
  Geist, and Bachem]{agarwal2024onpolicy}
Rishabh Agarwal, Nino Vieillard, Yongchao Zhou, Piotr Stanczyk, Sabela
  Ramos~Garea, Matthieu Geist, and Olivier Bachem.
\newblock On-policy distillation of language models: Learning from
  self-generated mistakes.
\newblock In \emph{International Conference on Learning Representations},
  volume 2024, pp.\  21246--21263, 2024.

\bibitem[Ahmadian et~al.(2024)Ahmadian, Cremer, Gall{\'e}, Fadaee, Kreutzer,
  Pietquin, {\"U}st{\"u}n, and Hooker]{ahmadian2024back}
Arash Ahmadian, Chris Cremer, Matthias Gall{\'e}, Marzieh Fadaee, Julia
  Kreutzer, Olivier Pietquin, Ahmet {\"U}st{\"u}n, and Sara Hooker.
\newblock Back to basics: Revisiting reinforce-style optimization for learning
  from human feedback in llms.
\newblock pp.\  12248--12267, 2024.

\bibitem[Aksitov et~al.(2023)Aksitov, Miryoosefi, Li, Li, Babayan, Kopparapu,
  Fisher, Guo, Prakash, Srinivasan, et~al.]{gulcehre2023rest}
Renat Aksitov, Sobhan Miryoosefi, Zonglin Li, Daliang Li, Sheila Babayan, Kavya
  Kopparapu, Zachary Fisher, Ruiqi Guo, Sushant Prakash, Pranesh Srinivasan,
  et~al.
\newblock Rest meets react: Self-improvement for multi-step reasoning llm
  agent.
\newblock \emph{arXiv preprint arXiv:2312.10003}, 2023.

\bibitem[Azar et~al.(2024)Azar, Guo, Piot, Munos, Rowland, Valko, and
  Calandriello]{azar2024ipo}
Mohammad~Gheshlaghi Azar, Zhaohan~Daniel Guo, Bilal Piot, Remi Munos, Mark
  Rowland, Michal Valko, and Daniele Calandriello.
\newblock A general theoretical paradigm to understand learning from human
  preferences.
\newblock In \emph{International conference on artificial intelligence and
  statistics}, pp.\  4447--4455. PMLR, 2024.

\bibitem[Christiano et~al.(2017)Christiano, Leike, Brown, Martic, Legg, and
  Amodei]{christiano2017deep}
Paul~F Christiano, Jan Leike, Tom Brown, Miljan Martic, Shane Legg, and Dario
  Amodei.
\newblock Deep reinforcement learning from human preferences.
\newblock \emph{Advances in neural information processing systems}, 30, 2017.

\bibitem[Chung et~al.(2018)Chung, Ghiassian, Nath, and
  White]{chung2018rejection}
Wesley Chung, Sina Ghiassian, Somjit Nath, and Martha White.
\newblock Rejection sampling for off-policy learning.
\newblock In \emph{NeurIPS 2018 Workshop on Continual Learning}, 2018.

\bibitem[Devroye(1986)]{devroye1986nonuniform}
Luc Devroye.
\newblock \emph{Non-Uniform Random Variate Generation}.
\newblock Springer-Verlag, New York, 1986.

\bibitem[Fang et~al.(2025)Fang, Cai, Li, Wu, Li, Yin, Wang, Wang, Su, Zhang,
  et~al.]{fang2025envscaling}
Runnan Fang, Shihao Cai, Baixuan Li, Jialong Wu, Guangyu Li, Wenbiao Yin, Xinyu
  Wang, Xiaobin Wang, Liangcai Su, Zhen Zhang, et~al.
\newblock Towards general agentic intelligence via environment scaling.
\newblock \emph{arXiv preprint arXiv:2509.13311}, 2025.

\bibitem[Fang et~al.(2026)Fang, Cai, Li, Wu, Li, Yin, Wang, Wang, Su, Zhang,
  et~al.]{cui2025process}
Runnan Fang, Shihao Cai, Baixuan Li, Jialong Wu, Guangyu Li, Wenbiao Yin, Xinyu
  Wang, Xiaobin Wang, Liangcai Su, Zhen Zhang, et~al.
\newblock Towards general agentic intelligence via environment scaling.
\newblock pp.\  17610--17621, 2026.

\bibitem[Fu et~al.(2023)Fu, Peng, Ou, Sabharwal, and Khot]{fu2023specializing}
Yao Fu, Hao Peng, Litu Ou, Ashish Sabharwal, and Tushar Khot.
\newblock Specializing smaller language models towards multi-step reasoning.
\newblock In \emph{International Conference on Machine Learning}, pp.\
  10421--10430. PMLR, 2023.

\bibitem[Gu et~al.(2024)Gu, Dong, Wei, and Huang]{gu2024minillm}
Yuxian Gu, Li~Dong, Furu Wei, and Minlie Huang.
\newblock Minillm: Knowledge distillation of large language models.
\newblock 2024:\penalty0 32694--32717, 2024.

\bibitem[Guo et~al.(2025)Guo, Yang, Zhang, Song, Wang, Zhu, Xu, Zhang, Ma, Bi,
  et~al.]{deepseek2025r1}
Daya Guo, Dejian Yang, Haowei Zhang, Junxiao Song, Peiyi Wang, Qihao Zhu,
  Runxin Xu, Ruoyu Zhang, Shirong Ma, Xiao Bi, et~al.
\newblock Deepseek-r1: Incentivizing reasoning capability in llms via
  reinforcement learning.
\newblock \emph{arXiv preprint arXiv:2501.12948}, 2025.

\bibitem[He et~al.(2024)He, Luo, Bai, Hu, Thai, Shen, Hu, Han, Huang, Zhang,
  et~al.]{he2024olympiadbench}
Chaoqun He, Renjie Luo, Yuzhuo Bai, Shengding Hu, Zhen Thai, Junhao Shen, Jinyi
  Hu, Xu~Han, Yujie Huang, Yuxiang Zhang, et~al.
\newblock Olympiadbench: A challenging benchmark for promoting agi with
  olympiad-level bilingual multimodal scientific problems.
\newblock In \emph{Proceedings of the 62nd Annual Meeting of the Association
  for Computational Linguistics (Volume 1: Long Papers)}, pp.\  3828--3850,
  2024.

\bibitem[Hinton et~al.(2015)Hinton, Vinyals, and Dean]{hinton2015distilling}
Geoffrey Hinton, Oriol Vinyals, and Jeff Dean.
\newblock Distilling the knowledge in a neural network.
\newblock \emph{arXiv preprint arXiv:1503.02531}, 2015.

\bibitem[Ho et~al.(2023)Ho, Schmid, and Yun]{ho2023large}
Namgyu Ho, Laura Schmid, and Se-Young Yun.
\newblock Large language models are reasoning teachers.
\newblock In \emph{Proceedings of the 61st annual meeting of the association
  for computational linguistics (volume 1: long papers)}, pp.\  14852--14882,
  2023.

\bibitem[Ho et~al.(2020)Ho, Nguyen, Sugawara, and Aizawa]{ho2020constructing}
Xanh Ho, Anh-Khoa~Duong Nguyen, Saku Sugawara, and Akiko Aizawa.
\newblock Constructing a multi-hop qa dataset for comprehensive evaluation of
  reasoning steps.
\newblock \emph{arXiv preprint arXiv:2011.01060}, 2020.

\bibitem[Hong et~al.(2024)Hong, Lee, and Thorne]{hong2024orpo}
Jiwoo Hong, Noah Lee, and James Thorne.
\newblock Orpo: Monolithic preference optimization without reference model.
\newblock \emph{arXiv preprint arXiv:2403.07691}, 2024.

\bibitem[Hsieh et~al.(2023)Hsieh, Li, Yeh, Nakhost, Fujii, Ratner, Krishna,
  Lee, and Pfister]{hsieh2023distilling}
Cheng-Yu Hsieh, Chun-Liang Li, Chih-Kuan Yeh, Hootan Nakhost, Yasuhisa Fujii,
  Alex Ratner, Ranjay Krishna, Chen-Yu Lee, and Tomas Pfister.
\newblock Distilling step-by-step! outperforming larger language models with
  less training data and smaller model sizes.
\newblock In \emph{Findings of the Association for Computational Linguistics:
  ACL 2023}, pp.\  8003--8017, 2023.

\bibitem[Jin et~al.(2025)Jin, Zeng, Yue, Yoon, Arik, Wang, Zamani, and
  Han]{jin2025searchr1}
Bowen Jin, Hansi Zeng, Zhenrui Yue, Jinsung Yoon, Sercan Arik, Dong Wang, Hamed
  Zamani, and Jiawei Han.
\newblock Search-r1: Training llms to reason and leverage search engines with
  reinforcement learning.
\newblock \emph{arXiv preprint arXiv:2503.09516}, 2025.

\bibitem[Joshi et~al.(2017)Joshi, Choi, Weld, and
  Zettlemoyer]{joshi2017triviaqa}
Mandar Joshi, Eunsol Choi, Daniel~S Weld, and Luke Zettlemoyer.
\newblock Triviaqa: A large scale distantly supervised challenge dataset for
  reading comprehension.
\newblock \emph{arXiv preprint arXiv:1705.03551}, 2017.

\bibitem[Jung et~al.(2025)Jung, Yoon, Kim, and Lee]{song2024todi}
Seongryong Jung, Suwan Yoon, DongGeon Kim, and Hwanhee Lee.
\newblock Todi: Token-wise distillation via fine-grained divergence control.
\newblock \emph{arXiv preprint arXiv:2505.16297}, 2025.

\bibitem[Kim \& Rush(2016)Kim and Rush]{kim2016sequence}
Yoon Kim and Alexander~M Rush.
\newblock Sequence-level knowledge distillation.
\newblock In \emph{Proceedings of the 2016 conference on empirical methods in
  natural language processing}, pp.\  1317--1327, 2016.

\bibitem[Koncel-Kedziorski et~al.(2016)Koncel-Kedziorski, Roy, Amini, Kushman,
  and Hajishirzi]{koncel2016mawps}
Rik Koncel-Kedziorski, Subhro Roy, Aida Amini, Nate Kushman, and Hannaneh
  Hajishirzi.
\newblock Mawps: A math word problem repository.
\newblock In \emph{Proceedings of the 2016 conference of the north american
  chapter of the association for computational linguistics: human language
  technologies}, pp.\  1152--1157, 2016.

\bibitem[Kwiatkowski et~al.(2019)Kwiatkowski, Palomaki, Redfield, Collins,
  Parikh, Alberti, Epstein, Polosukhin, Devlin, Lee,
  et~al.]{kwiatkowski2019natural}
Tom Kwiatkowski, Jennimaria Palomaki, Olivia Redfield, Michael Collins, Ankur
  Parikh, Chris Alberti, Danielle Epstein, Illia Polosukhin, Jacob Devlin,
  Kenton Lee, et~al.
\newblock Natural questions: a benchmark for question answering research.
\newblock \emph{Transactions of the Association for Computational Linguistics},
  7:\penalty0 452--466, 2019.

\bibitem[Lewis et~al.(2020)Lewis, Perez, Piktus, Petroni, Karpukhin, Goyal,
  K{\"u}ttler, Lewis, Yih, Rockt{\"a}schel, et~al.]{lewis2020retrieval}
Patrick Lewis, Ethan Perez, Aleksandra Piktus, Fabio Petroni, Vladimir
  Karpukhin, Naman Goyal, Heinrich K{\"u}ttler, Mike Lewis, Wen-tau Yih, Tim
  Rockt{\"a}schel, et~al.
\newblock Retrieval-augmented generation for knowledge-intensive nlp tasks.
\newblock \emph{Advances in neural information processing systems},
  33:\penalty0 9459--9474, 2020.

\bibitem[Li et~al.(2025)Li, Dong, Jin, Zhang, Zhou, Zhu, Zhang, and
  Dou]{li2025search}
Xiaoxi Li, Guanting Dong, Jiajie Jin, Yuyao Zhang, Yujia Zhou, Yutao Zhu,
  Peitian Zhang, and Zhicheng Dou.
\newblock Search-o1: Agentic search-enhanced large reasoning models.
\newblock \emph{arXiv preprint arXiv:2501.05366}, 2025.

\bibitem[Lin et~al.(2020)Lin, Wohlwend, Chen, and Lei]{lin2020autoregressive}
Alexander Lin, Jeremy Wohlwend, Howard Chen, and Tao Lei.
\newblock Autoregressive knowledge distillation through imitation learning.
\newblock \emph{arXiv preprint arXiv:2009.07253}, 2020.

\bibitem[Liu et~al.(2023)Liu, Zhao, Joshi, Khalman, Saleh, Liu, and
  Liu]{liu2024rspo}
Tianqi Liu, Yao Zhao, Rishabh Joshi, Misha Khalman, Mohammad Saleh, Peter~J
  Liu, and Jialu Liu.
\newblock Statistical rejection sampling improves preference optimization.
\newblock \emph{arXiv preprint arXiv:2309.06657}, 2023.

\bibitem[Lu \& {Thinking Machines Lab}(2025)Lu and {Thinking Machines
  Lab}]{lu2025onpolicydistillation}
Kevin Lu and {Thinking Machines Lab}.
\newblock On-policy distillation.
\newblock \url{https://thinkingmachines.ai/blog/on-policy-distillation}, 2025.

\bibitem[Lu et~al.(2022)Lu, Qiu, Chang, Wu, Zhu, Rajpurohit, Clark, and
  Kalyan]{lu2022dynamic}
Pan Lu, Liang Qiu, Kai-Wei Chang, Ying~Nian Wu, Song-Chun Zhu, Tanmay
  Rajpurohit, Peter Clark, and Ashwin Kalyan.
\newblock Dynamic prompt learning via policy gradient for semi-structured
  mathematical reasoning.
\newblock \emph{arXiv preprint arXiv:2209.14610}, 2022.

\bibitem[Luo et~al.(2025)Luo, Tan, Wong, Shi, Tang, Roongta, Cai, Luo, Zhang,
  Li, Popa, and Stoica]{deepscaler2025}
Michael Luo, Sijun Tan, Justin Wong, Xiaoxiang Shi, William~Y. Tang, Manan
  Roongta, Colin Cai, Jeffrey Luo, Tianjun Zhang, Li~Erran Li, Raluca~Ada Popa,
  and Ion Stoica.
\newblock {DeepScaleR}: Surpassing {O1-Preview} with a {1.5B} model by scaling
  {RL}.
\newblock
  \url{https://pretty-radio-b75.notion.site/19681902c1468005bed8ca303013a4e2},
  2025.
\newblock Notion Blog.

\bibitem[Magister et~al.(2023)Magister, Mallinson, Adamek, Malmi, and
  Severyn]{magister2023teaching}
Lucie~Charlotte Magister, Jonathan Mallinson, Jakub Adamek, Eric Malmi, and
  Aliaksei Severyn.
\newblock Teaching small language models to reason.
\newblock In \emph{Proceedings of the 61st annual meeting of the association
  for computational linguistics (volume 2: short papers)}, pp.\  1773--1781,
  2023.

\bibitem[Mallen et~al.(2023)Mallen, Asai, Zhong, Das, Khashabi, and
  Hajishirzi]{mallen2023popqa}
Alex Mallen, Akari Asai, Victor Zhong, Rajarshi Das, Daniel Khashabi, and
  Hannaneh Hajishirzi.
\newblock When not to trust language models: Investigating effectiveness of
  parametric and non-parametric memories.
\newblock In \emph{Proceedings of the 61st Annual Meeting of the Association
  for Computational Linguistics (Volume 1: Long Papers)}, pp.\  9802--9822,
  2023.

\bibitem[Meng et~al.(2024)Meng, Xia, and Chen]{meng2024simpo}
Yu~Meng, Mengzhou Xia, and Danqi Chen.
\newblock Simpo: Simple preference optimization with a reference-free reward.
\newblock \emph{Advances in Neural Information Processing Systems},
  37:\penalty0 124198--124235, 2024.

\bibitem[Mialon et~al.(2024)Mialon, Fourrier, Wolf, LeCun, and
  Scialom]{mialon2023gaia}
Gr{\'e}goire Mialon, Cl{\'e}mentine Fourrier, Thomas Wolf, Yann LeCun, and
  Thomas Scialom.
\newblock Gaia: a benchmark for general ai assistants.
\newblock In \emph{International Conference on Learning Representations},
  volume 2024, pp.\  9025--9049, 2024.

\bibitem[Miao et~al.(2020)Miao, Liang, and Su]{miao2020diverse}
Shen-Yun Miao, Chao-Chun Liang, and Keh-Yih Su.
\newblock A diverse corpus for evaluating and developing english math word
  problem solvers.
\newblock In \emph{Proceedings of the 58th annual meeting of the Association
  for Computational Linguistics}, pp.\  975--984, 2020.

\bibitem[Niu et~al.(2025)Niu, Sun, Zhao, Chen, Bai, Zhang, Yuan, and
  Wang]{niu2025cotd}
Lujie Niu, Haochen Sun, Fangkun Zhao, Sheng Chen, Zimeng Bai, Jiawei Zhang,
  Caixia Yuan, and Xiaojie Wang.
\newblock Cotd-po: Chain-of-thought distillation with preference optimization.
\newblock In \emph{Findings of the Association for Computational Linguistics:
  EMNLP 2025}, pp.\  19975--19986, 2025.

\bibitem[Ouyang et~al.(2022)Ouyang, Wu, Jiang, Almeida, Wainwright, Mishkin,
  Zhang, Agarwal, Slama, Ray, et~al.]{ouyang2022training}
Long Ouyang, Jeffrey Wu, Xu~Jiang, Diogo Almeida, Carroll Wainwright, Pamela
  Mishkin, Chong Zhang, Sandhini Agarwal, Katarina Slama, Alex Ray, et~al.
\newblock Training language models to follow instructions with human feedback.
\newblock \emph{Advances in neural information processing systems},
  35:\penalty0 27730--27744, 2022.

\bibitem[Press et~al.(2023)Press, Zhang, Min, Schmidt, Smith, and
  Lewis]{press2023measuring}
Ofir Press, Muru Zhang, Sewon Min, Ludwig Schmidt, Noah~A Smith, and Mike
  Lewis.
\newblock Measuring and narrowing the compositionality gap in language models.
\newblock In \emph{Findings of the Association for Computational Linguistics:
  EMNLP 2023}, pp.\  5687--5711, 2023.

\bibitem[Rafailov et~al.(2023)Rafailov, Sharma, Mitchell, Manning, Ermon, and
  Finn]{rafailov2024direct}
Rafael Rafailov, Archit Sharma, Eric Mitchell, Christopher~D Manning, Stefano
  Ermon, and Chelsea Finn.
\newblock Direct preference optimization: Your language model is secretly a
  reward model.
\newblock \emph{Advances in neural information processing systems},
  36:\penalty0 53728--53741, 2023.

\bibitem[Schlegel et~al.(2019)Schlegel, Chung, Graves, Qian, and
  White]{schlegel2019importance}
Matthew Schlegel, Wesley Chung, Daniel Graves, Jian Qian, and Martha White.
\newblock Importance resampling for off-policy prediction.
\newblock \emph{Advances in Neural Information Processing Systems}, 32, 2019.

\bibitem[Schulman et~al.(2017)Schulman, Wolski, Dhariwal, Radford, and
  Klimov]{schulman2017proximal}
John Schulman, Filip Wolski, Prafulla Dhariwal, Alec Radford, and Oleg Klimov.
\newblock Proximal policy optimization algorithms.
\newblock \emph{arXiv preprint arXiv:1707.06347}, 2017.

\bibitem[Shao et~al.(2024)Shao, Wang, Zhu, Xu, Song, Bi, Zhang, Zhang, Li, Wu,
  et~al.]{shao2024deepseekmath}
Zhihong Shao, Peiyi Wang, Qihao Zhu, Runxin Xu, Junxiao Song, Xiao Bi, Haowei
  Zhang, Mingchuan Zhang, YK~Li, Yang Wu, et~al.
\newblock Deepseekmath: Pushing the limits of mathematical reasoning in open
  language models.
\newblock \emph{arXiv preprint arXiv:2402.03300}, 2024.

\bibitem[Singh et~al.(2023)Singh, Co-Reyes, Agarwal, Anand, Patil, Garcia, Liu,
  Harrison, Lee, Xu, et~al.]{singh2024humanlevel}
Avi Singh, John~D Co-Reyes, Rishabh Agarwal, Ankesh Anand, Piyush Patil, Xavier
  Garcia, Peter~J Liu, James Harrison, Jaehoon Lee, Kelvin Xu, et~al.
\newblock Beyond human data: Scaling self-training for problem-solving with
  language models.
\newblock \emph{arXiv preprint arXiv:2312.06585}, 2023.

\bibitem[Spigler(2024)]{wang2024proximal}
Giacomo Spigler.
\newblock Proximal policy distillation.
\newblock \emph{arXiv preprint arXiv:2407.15134}, 2024.

\bibitem[Stiennon et~al.(2020)Stiennon, Ouyang, Wu, Ziegler, Lowe, Voss,
  Radford, Amodei, and Christiano]{stiennon2020learning}
Nisan Stiennon, Long Ouyang, Jeffrey Wu, Daniel Ziegler, Ryan Lowe, Chelsea
  Voss, Alec Radford, Dario Amodei, and Paul~F Christiano.
\newblock Learning to summarize with human feedback.
\newblock \emph{Advances in neural information processing systems},
  33:\penalty0 3008--3021, 2020.

\bibitem[Sun et~al.(2025)Sun, Qiao, Guo, Fan, Hou, Jiang, Xie, Zhang, Huang,
  and Zhou]{sun2025zerosearch}
Hao Sun, Zile Qiao, Jiayan Guo, Xuanbo Fan, Yingyan Hou, Yong Jiang, Pengjun
  Xie, Yan Zhang, Fei Huang, and Jingren Zhou.
\newblock Zerosearch: Incentivize the search capability of llms without
  searching.
\newblock \emph{arXiv preprint arXiv:2505.04588}, 2025.

\bibitem[Trivedi et~al.(2022)Trivedi, Balasubramanian, Khot, and
  Sabharwal]{trivedi2022musique}
Harsh Trivedi, Niranjan Balasubramanian, Tushar Khot, and Ashish Sabharwal.
\newblock Musique: Multihop questions via single-hop question composition.
\newblock \emph{Transactions of the Association for Computational Linguistics},
  10:\penalty0 539--554, 2022.

\bibitem[Wei et~al.(2022)Wei, Wang, Schuurmans, Bosma, Xia, Chi, Le, Zhou,
  et~al.]{wei2022chain}
Jason Wei, Xuezhi Wang, Dale Schuurmans, Maarten Bosma, Fei Xia, Ed~Chi, Quoc~V
  Le, Denny Zhou, et~al.
\newblock Chain-of-thought prompting elicits reasoning in large language
  models.
\newblock \emph{Advances in neural information processing systems},
  35:\penalty0 24824--24837, 2022.

\bibitem[Wen et~al.(2026)Wen, Dong, and Dou]{wen2026smartsearch}
Tongyu Wen, Guanting Dong, and Zhicheng Dou.
\newblock Smartsearch: Process reward-guided query refinement for search
  agents, 2026.

\bibitem[Wen et~al.(2023)Wen, Li, Du, and Mou]{wen2023fdistill}
Yuqiao Wen, Zichao Li, Wenyu Du, and Lili Mou.
\newblock F-divergence minimization for sequence-level knowledge distillation.
\newblock \emph{arXiv preprint arXiv:2307.15190}, 2023.

\bibitem[West et~al.(2022)West, Bhagavatula, Hessel, Hwang, Jiang, Le~Bras, Lu,
  Welleck, and Choi]{west2022symbolic}
Peter West, Chandra Bhagavatula, Jack Hessel, Jena Hwang, Liwei Jiang, Ronan
  Le~Bras, Ximing Lu, Sean Welleck, and Yejin Choi.
\newblock Symbolic knowledge distillation: from general language models to
  commonsense models.
\newblock In \emph{Proceedings of the 2022 conference of the North American
  chapter of the association for computational linguistics: Human language
  technologies}, pp.\  4602--4625, 2022.

\bibitem[Wijmans et~al.(2024)Wijmans, Huval, Hertzberg, Koltun, and
  Kr{\"a}henb{\"u}hl]{apple2024cce}
Erik Wijmans, Brody Huval, Alexander Hertzberg, Vladlen Koltun, and Philipp
  Kr{\"a}henb{\"u}hl.
\newblock Cut your losses in large-vocabulary language models.
\newblock \emph{arXiv preprint arXiv:2411.09009}, 2024.

\bibitem[Wu et~al.(2023)Wu, Chen, Quan, Wang, and Wang]{wu2024adkd}
Siyue Wu, Hongzhan Chen, Xiaojun Quan, Qifan Wang, and Rui Wang.
\newblock Ad-kd: Attribution-driven knowledge distillation for language model
  compression.
\newblock \emph{arXiv preprint arXiv:2305.10010}, 2023.

\bibitem[Xiong et~al.(2025)Xiong, Yao, Xu, Pang, Wang, Sahoo, Li, Jiang, Zhang,
  Xiong, et~al.]{yuan2025minimalist}
Wei Xiong, Jiarui Yao, Yuhui Xu, Bo~Pang, Lei Wang, Doyen Sahoo, Junnan Li, Nan
  Jiang, Tong Zhang, Caiming Xiong, et~al.
\newblock A minimalist approach to llm reasoning: from rejection sampling to
  reinforce.
\newblock \emph{arXiv preprint arXiv:2504.11343}, 2025.

\bibitem[Yang et~al.(2024)Yang, Yang, Zhang, Hui, Zheng, Yu, Li, Liu, Huang,
  Wei, Lin, Yang, Tu, Zhang, Yang, Yang, Zhou, Lin, Dang, Lu, Bao, Yang, Yu,
  Li, Xue, Zhang, Zhu, Men, Lin, Li, Xia, Ren, Ren, Fan, Su, Zhang, Wan, Liu,
  Cui, Zhang, and Qiu]{qwen2025qwen25technicalreport}
An~Yang, Baosong Yang, Beichen Zhang, Binyuan Hui, Bo~Zheng, Bowen Yu,
  Chengyuan Li, Dayiheng Liu, Fei Huang, Haoran Wei, Huan Lin, Jian Yang,
  Jianhong Tu, Jianwei Zhang, Jianxin Yang, Jiaxi Yang, Jingren Zhou, Junyang
  Lin, Kai Dang, Keming Lu, Keqin Bao, Kexin Yang, Le~Yu, Mei Li, Mingfeng Xue,
  Pei Zhang, Qin Zhu, Rui Men, Runji Lin, Tianhao Li, Tingyu Xia, Xingzhang
  Ren, Xuancheng Ren, Yang Fan, Yang Su, Yichang Zhang, Yu~Wan, Yuqiong Liu,
  Zeyu Cui, Zhenru Zhang, and Zihan Qiu.
\newblock Qwen2.5 technical report.
\newblock \emph{arXiv preprint arXiv:2412.15115}, 2024.

\bibitem[Yang et~al.(2018)Yang, Qi, Zhang, Bengio, Cohen, Salakhutdinov, and
  Manning]{yang2018hotpotqa}
Zhilin Yang, Peng Qi, Saizheng Zhang, Yoshua Bengio, William Cohen, Ruslan
  Salakhutdinov, and Christopher~D Manning.
\newblock Hotpotqa: A dataset for diverse, explainable multi-hop question
  answering.
\newblock In \emph{Proceedings of the 2018 conference on empirical methods in
  natural language processing}, pp.\  2369--2380, 2018.

\bibitem[Ye et~al.(2025)Ye, Dong, Chi, Wu, Huang, and Wei]{ye2025black}
Tianzhu Ye, Li~Dong, Zewen Chi, Xun Wu, Shaohan Huang, and Furu Wei.
\newblock Black-box on-policy distillation of large language models.
\newblock \emph{arXiv preprint arXiv:2511.10643}, 2025.

\bibitem[Yu et~al.(2025)Yu, Zhang, Zhu, Yuan, Zuo, Yue, Dai, Fan, Liu, Liu,
  et~al.]{yu2025dapo}
Qiying Yu, Zheng Zhang, Ruofei Zhu, Yufeng Yuan, Xiaochen Zuo, Yu~Yue, Weinan
  Dai, Tiantian Fan, Gaohong Liu, Lingjun Liu, et~al.
\newblock Dapo: An open-source llm reinforcement learning system at scale.
\newblock \emph{arXiv preprint arXiv:2503.14476}, 2025.

\bibitem[Yue et~al.(2025)Yue, Chen, Lu, Zhao, Wang, Yue, Song, and
  Huang]{yue2025does}
Yang Yue, Zhiqi Chen, Rui Lu, Andrew Zhao, Zhaokai Wang, Yang Yue, Shiji Song,
  and Gao Huang.
\newblock Does reinforcement learning really incentivize reasoning capacity in
  {LLMs} beyond the base model?
\newblock In \emph{Advances in Neural Information Processing Systems},
  volume~38, pp.\  57654--57689, 2025.
\newblock URL
  \url{https://papers.neurips.cc/paper_files/paper/2025/hash/537d5aa768c2d534016a4d06f87bc8fb-Abstract-Conference.html}.

\bibitem[Zelikman et~al.(2022)Zelikman, Wu, Mu, and Goodman]{zelikman2022star}
Eric Zelikman, Yuhuai Wu, Jesse Mu, and Noah Goodman.
\newblock Star: Bootstrapping reasoning with reasoning.
\newblock \emph{Advances in Neural Information Processing Systems},
  35:\penalty0 15476--15488, 2022.

\bibitem[Zheng et~al.(2025)Zheng, An, Wang, Wang, and Wu]{zheng2025stepsearch}
Xuhui Zheng, Kang An, Ziliang Wang, Yuhang Wang, and Yichao Wu.
\newblock {S}tep{S}earch: Igniting {LLM}s search ability via step-wise proximal
  policy optimization.
\newblock In \emph{Proceedings of the 2025 Conference on Empirical Methods in
  Natural Language Processing}, pp.\  21805--21830, Suzhou, China, November
  2025. Association for Computational Linguistics.
\newblock \doi{10.18653/v1/2025.emnlp-main.1106}.
\newblock URL \url{https://aclanthology.org/2025.emnlp-main.1106/}.

\bibitem[Ziegler et~al.(2019)Ziegler, Stiennon, Wu, Brown, Radford, Amodei,
  Christiano, and Irving]{ziegler2019fine}
Daniel~M Ziegler, Nisan Stiennon, Jeffrey Wu, Tom~B Brown, Alec Radford, Dario
  Amodei, Paul Christiano, and Geoffrey Irving.
\newblock Fine-tuning language models from human preferences.
\newblock \emph{arXiv preprint arXiv:1909.08593}, 2019.

\end{thebibliography}
\endgroup

\clearpage
\appendix
\section{Appendix}
\label{sec:appendix}

\subsection{Comparison of Distillation Methods}
\label{sec:method_comparison_app}
\noindent\begin{minipage}[t]{0.5\textwidth}
Table~\ref{tab:method_comparison} situates \textsc{OVD} next to common distillation templates discussed in \S\ref{subsec:token_opd_limitations}. Sequence-level SeqKD aligns whole sampled trajectories without per-step logits, typically using rollouts that the teacher has sampled in advance (offline target trajectories); token-level variants supervise every decoding step, either with full-vocabulary logits (the OPD reverse-KL setting of~\citet{agarwal2024onpolicy}) or with a truncated top-$k$ support that trades accuracy for the memory scaling in Figure~\ref{fig:memory_scaling}. \textsc{OVD} instead scores coarse steps with verbal judgments, avoiding dense vocabulary tensors while keeping on-policy rollouts under GRPO.
\end{minipage}%
\hfill
\begin{minipage}[t]{0.47\textwidth}
\centering
\captionsetup{type=table,aboveskip=0.25ex,belowskip=5pt,skip=5pt}
\captionof{table}{Comparison of language-model distillation methods. Logit-based methods use either the full vocabulary or the top-$k$ tokens.}
\label{tab:method_comparison}
\small
\setlength{\tabcolsep}{4pt}
\resizebox{\linewidth}{!}{%
\begin{tabular}{lcccc}
\toprule
\textbf{Method} & \textbf{Granularity} & \textbf{Feedback} & \textbf{On-Policy} & \textbf{Mem.} \\
\midrule
SeqKD & Sequence & Samples & \ding{55} & Low \\
Off-Policy KD (full $V$) & Token & Full-vocab.\ logits & \ding{55} & High \\
Off-Policy KD (top-$k$) & Token & Top-$k$ logits & \ding{55} & Med. \\
On-Policy KD (full $V$) & Token & Full-vocab.\ logits & \ding{51} & High \\
On-Policy KD (top-$k$) & Token & Top-$k$ logits & \ding{51} & Med. \\
\textsc{OVD} (Ours) & Step & Verbal & \ding{51} & Med. \\
\bottomrule
\end{tabular}%
}

\end{minipage}
\par\medskip
\noindent
In the same reading, \textbf{Granularity} marks whether feedback attaches to an entire sequence, to each token, or to step-sized segments $s_t$; \textbf{Feedback} distinguishes target samples, full-$V$ or top-$k$ logits, and discrete verbal scores; \textbf{On-Policy} flags methods that optimize on trajectories drawn from the student (checkmarks), including token-level OPD and \textsc{OVD}, as opposed to off-policy KD where behavior and learning policies need not coincide; and \textbf{Mem.} is a qualitative footprint where full-$V$ matching is highest, top-$k$ and verbal step scoring sit in the middle ground between that cost and low-footprint sequence KD. This layout is qualitatively aligned with the OPD$^\ast$ rows (full-$V$ reverse KL vs.\ top-$k$ logits) and the \textsc{OVD} configurations with $(\tau_{\mathrm{tr}},\tau_{\mathrm{te}})$ in Table~\ref{tab:qa_em_results}; Table~\ref{tab:method_comparison} should nevertheless be read as a coarse qualitative summary of memory efficiency and on-policy training across method families, not as a separate empirical claim.
\WFclear

\subsection{Theoretical Analysis}
\label{sec:theory}

We analyze how verbal rejection sampling constructs the training distribution and derive a conditional approximation bound in terms of acceptance calibration and teacher-replacement error. Section~\ref{sec:appendix_proofs} discusses normalization, the scope of the bound, and empirical score-scale comparisons.

\textbf{Conditions for Exact Matching of the Accepted Component.} In Eq.~\eqref{eq:verbal_density_ratio_calibration}, the proportionality constant must be positive and independent of $y$ for fixed $q$ and $\tau$, with acceptance probabilities bounded by one. The student must assign nonzero probability to every trajectory with positive probability under $p_T(\cdot\mid q)$. Under these conditions, substituting the acceptance rule into Eq.~\eqref{eq:vrs_distribution} and normalizing yields $p_{\mathrm a,\tau}(y\mid q)=p_T(y\mid q)$. This result concerns the accepted student component; the mixed distribution in Eq.~\eqref{eq:vrs_mixed_subtrajectory} also depends on teacher replacements.

\begin{assumption}[Approximate Density-Ratio Condition]
\label{ass:approx_density_ratio}
Fix a problem $q$, trajectory prefix $y_{<t}$, and threshold $\tau$, and suppress $q$ in the following local conditionals. Let $p_T(\cdot\mid y_{<t})$ be a normalized target specified independently of the acceptance rule, with support contained in that of $\pi_S(\cdot\mid y_{<t})$. There exists a constant $c>0$, independent of $s_t$, such that $c\,p_T(s_t\mid y_{<t})/\pi_S(s_t\mid y_{<t})\in[0,1]$ on the student support and
\begin{equation}
\mathbb E_{s_t\sim\pi_S(\cdot\mid y_{<t})}
\left[\left|a_\tau(s_t\mid y_{<t})-c\,\frac{p_T(s_t\mid y_{<t})}{\pi_S(s_t\mid y_{<t})}\right|\right]
\leq\varepsilon,
\label{eq:approx_density_ratio_condition}
\end{equation}
where $\varepsilon\geq0$ measures the mean absolute error in acceptance probabilities and $Z_\tau(y_{<t})>0$.
\end{assumption}

The empirical diagnostic in Figure~\ref{fig:verbal_acceptance_likelihood} of the main text examines how verbal acceptance varies with a teacher-generation density-ratio proxy. The following proposition states the separate assumptions required for a distribution-approximation guarantee.

\begin{proposition}[Approximation of the Accepted and Mixed Distributions]
\label{prop:approx_density_ratio}
Under Assumption~\ref{ass:approx_density_ratio}, the accepted student distribution at prefix $y_{<t}$ satisfies
\begin{equation}
\operatorname{TV}\!\left(p_{\mathrm a,\tau}(\cdot\mid y_{<t}),p_T(\cdot\mid y_{<t})\right)
\leq\min\!\left\{1,\frac{\varepsilon}{Z_\tau(y_{<t})}\right\}.
\label{eq:accepted_approximation_bound}
\end{equation}
If the teacher replacement distribution satisfies $\operatorname{TV}(r_{\mathcal E,\tau}(\cdot\mid y_{<t}),p_T(\cdot\mid y_{<t}))\leq\eta$, then
\begin{equation}
\operatorname{TV}\!\left(p_{\mathrm m,\tau}(\cdot\mid y_{<t}),p_T(\cdot\mid y_{<t})\right)
\leq\min\!\left\{1,\varepsilon+\bigl(1-Z_\tau(y_{<t})\bigr)\eta\right\}.
\label{eq:mixed_approximation_bound}
\end{equation}
Here $\operatorname{TV}(p,r)=\tfrac12\sum_{s_t}|p(s_t)-r(s_t)|$ is total variation distance. When $Z_\tau(y_{<t})=1$, the replacement term vanishes and no condition on teacher replacements is needed.
\end{proposition}

\begin{proof}
Suppress the fixed conditioning on $y_{<t}$ and write $Z=Z_\tau(y_{<t})$ and $u(s_t)=\pi_S(s_t\mid y_{<t})a_\tau(s_t\mid y_{<t})$. Assumption~\ref{ass:approx_density_ratio} gives $\lVert u-cp_T\rVert_1\leq\varepsilon$, hence $|Z-c|\leq\varepsilon$. Since $p_{\mathrm a,\tau}=u/Z$,
\[
\operatorname{TV}(p_{\mathrm a,\tau},p_T)
=\frac{\lVert u-Zp_T\rVert_1}{2Z}
\leq\frac{\lVert u-cp_T\rVert_1+|c-Z|}{2Z}
\leq\frac{\varepsilon}{Z}.
\]
Equation~\eqref{eq:vrs_mixed_subtrajectory} gives $p_{\mathrm m,\tau}=Zp_{\mathrm a,\tau}+(1-Z)r_{\mathcal E,\tau}$. Convexity of total variation therefore yields
\[
\operatorname{TV}(p_{\mathrm m,\tau},p_T)
\leq Z\operatorname{TV}(p_{\mathrm a,\tau},p_T)
+(1-Z)\operatorname{TV}(r_{\mathcal E,\tau},p_T)
\leq\varepsilon+(1-Z)\eta.
\]
Total variation is at most one, giving the stated bounds.
\end{proof}

The condition quantifies proportionality of acceptance probabilities to a density ratio, which ranking accuracy alone does not establish. The bounds apply to each fixed prefix: the accepted distribution has error at most $\varepsilon/Z_\tau(y_{<t})$, while the bound for the mixed distribution also accounts for teacher replacement error. Setting $\varepsilon=0$ recovers exact matching of the accepted component; if the replacement error is also zero, the mixed component matches $p_T(\cdot\mid y_{<t})$.

\begin{proposition}[Approximation of Complete Mixed Trajectories]
\label{prop:complete_mixed_trajectories}
Fix a problem $q$ and a finite horizon $T$ of sub-trajectory transitions. Use a common segment representation and termination rule for the mixed and target processes, padding terminated trajectories with an absorbing symbol. Suppress $q$, and let
\[
P_{\mathrm m}(y)=\prod_{t=1}^{T}p_{\mathrm m,t}(s_t\mid y_{<t}),
\qquad
P_T(y)=\prod_{t=1}^{T}p_{T,t}(s_t\mid y_{<t}).
\]
The target conditionals are specified independently of acceptance and jointly define one target process. Suppose that, for each $t$, there is a bound $\delta_t\in[0,1]$ such that
\[
\operatorname{TV}\!\left(p_{\mathrm m,t}(\cdot\mid h),p_{T,t}(\cdot\mid h)\right)
\leq\delta_t
\]
for every prefix $h$ reachable under either process. Then
\begin{equation}
\operatorname{TV}(P_{\mathrm m},P_T)
\leq 1-\prod_{t=1}^{T}(1-\delta_t)
\leq\min\!\left\{1,\sum_{t=1}^{T}\delta_t\right\}.
\label{eq:complete_mixed_trajectory_bound}
\end{equation}
In particular, if Proposition~\ref{prop:approx_density_ratio} applies at every such nonterminal prefix with local errors $\varepsilon_t(h)$, $\eta_t(h)$ and acceptance rate $Z_t(h)>0$, one may take
\[
\delta_t=\sup_h\min\!\left\{1,\varepsilon_t(h)+(1-Z_t(h))\eta_t(h)\right\},
\]
with zero local error at absorbed prefixes. At a prefix with $Z_t(h)=0$, the local bound instead comes directly from the replacement distribution.
\end{proposition}

\begin{proof}
Couple the two processes sequentially. Whenever their prefixes agree, use a maximal coupling of their next-segment conditionals, whose probability of disagreement is their total variation distance and hence at most $\delta_t$. Once their prefixes differ, continue with any coupling preserving the respective conditionals. If $E_t$ is the event that the prefixes agree through step $t$, then
\[
\Pr(E_t)\geq(1-\delta_t)\Pr(E_{t-1}),
\qquad
\Pr(E_T)\geq\prod_{t=1}^{T}(1-\delta_t).
\]
The coupling inequality gives $\operatorname{TV}(P_{\mathrm m},P_T)\leq\Pr(E_T^c)$. Finally, $1-\prod_t(1-\delta_t)\leq\sum_t\delta_t$ and total variation is at most one. Absorbing transitions agree deterministically after joint termination, so padding introduces no additional error.
\end{proof}

Consequently, for any common final-answer success indicator $f(y)\in\{0,1\}$, the difference $|\mathbb E_{P_{\mathrm m}}f-\mathbb E_{P_T}f|$ is bounded by the right-hand side of Eq.~\eqref{eq:complete_mixed_trajectory_bound}. The bound concerns the trajectory-construction process at a fixed student policy. It applies to complete trajectories when termination occurs by $T$, or to capped trajectories when a generation limit is imposed.

\paragraph{Discussion.}
The target is specified independently of verbal scores. Ranking accuracy alone does not imply the density-ratio condition, and teacher resampling alone does not guarantee a small replacement error. Proposition~\ref{prop:complete_mixed_trajectories} extends local approximation to complete mixed trajectories under uniform prefix-wise bounds and a common sequential representation. It does not establish convergence of the trained student to the target or a reduction in gradient variance. The mixed distribution constructs training data, while the student is updated using the clipped objective defined in the main text.

\subsubsection{Empirical Diagnostic of Verbal Acceptance}
\label{sec:verbal_acceptance_diagnostic}

Figure~\ref{fig:verbal_acceptance_likelihood_full} examines the relationship between verbal acceptance and the teacher-to-student log density ratio on 6,400 base-student answers to 200 Math500 questions, with 32 samples per question. We use the teacher generation policy $\pi_{\mathcal E}$ as an observable proxy for the teacher-preferred target $p_T$. Both models score the same answer text using their respective tokenizers and prompt templates; terminal stop tokens are excluded, and outputs reaching the generation cap are scored as prefixes. After within-question centering, we group answers into ten equal-count bins and estimate the fraction accepted at each threshold, with 95\% question-bootstrap intervals.

Acceptance generally rises from the lowest-ratio bins and then levels off, with a similar broad trend in the length-normalized diagnostic. Raising the threshold from $\tau=3$ or $5$ to $7$ lowers acceptance substantially; the highest length-normalized bin also shows a decline at $\tau=7$. These patterns are consistent with verbal filtering favoring answers with higher relative teacher likelihood, but do not establish the proportionality condition in Eq.~\eqref{eq:verbal_density_ratio_calibration}. In particular, $\pi_{\mathcal E}$ is only a proxy for $p_T$, and a binned association does not demonstrate trajectory-wise calibration or bound the replacement error.

\begin{figure}[!htbp]
\centering
\begin{minipage}{0.49\linewidth}
\centering
\begin{tikzpicture}
\begin{axis}[width=\linewidth,height=4.2cm,scale only axis=false,
ylabel={Teacher acceptance rate},ymin=0,ymax=1,
axis lines=left,ymajorgrids=true,grid style={gray!20},
tick label style={font=\scriptsize},label style={font=\scriptsize},
legend style={font=\scriptsize,draw=none,at={(0.02,0.98)},anchor=north west},
title={Log density ratio},title style={font=\small},xlabel={Centered $\log(\pi_{\mathcal E}/\pi_S)$}]
\addplot+[color=blue!75!black,mark=*,mark size=1.5pt,line width=0.8pt,error bars/.cd,y dir=both,y explicit] table[x=x,y=y,y error plus=plus,y error minus=minus,row sep=\\]{
x y plus minus\\
-1016.389267 0.589063 0.044176 0.045874\\
-316.084502 0.637500 0.042113 0.041269\\
-100.698856 0.785937 0.034699 0.033251\\
18.452301 0.843750 0.032008 0.033410\\
88.514558 0.870313 0.028521 0.032012\\
140.438597 0.878125 0.031815 0.033972\\
185.693899 0.873437 0.031113 0.032803\\
237.850273 0.846875 0.034900 0.036030\\
305.836314 0.857812 0.028698 0.030889\\
456.386682 0.857812 0.034933 0.035077\\
};
\addlegendentry{$\tau=3$}
\addplot+[color=orange!85!black,mark=square*,mark size=1.5pt,line width=0.8pt,error bars/.cd,y dir=both,y explicit] table[x=x,y=y,y error plus=plus,y error minus=minus,row sep=\\]{
x y plus minus\\
-1016.389267 0.579688 0.044156 0.045637\\
-316.084502 0.634375 0.041662 0.042137\\
-100.698856 0.779687 0.035076 0.034404\\
18.452301 0.834375 0.032104 0.035044\\
88.514558 0.867188 0.028045 0.031128\\
140.438597 0.875000 0.032321 0.035574\\
185.693899 0.870313 0.031181 0.032611\\
237.850273 0.846875 0.034900 0.036030\\
305.836314 0.857812 0.028698 0.030889\\
456.386682 0.856250 0.034469 0.034824\\
};
\addlegendentry{$\tau=5$}
\addplot+[color=teal!80!black,mark=triangle*,mark size=1.5pt,line width=0.8pt,error bars/.cd,y dir=both,y explicit] table[x=x,y=y,y error plus=plus,y error minus=minus,row sep=\\]{
x y plus minus\\
-1016.389267 0.135937 0.027140 0.027541\\
-316.084502 0.242188 0.033420 0.032312\\
-100.698856 0.337500 0.044597 0.043102\\
18.452301 0.442188 0.066160 0.066361\\
88.514558 0.506250 0.059045 0.058590\\
140.438597 0.562500 0.060406 0.064073\\
185.693899 0.523438 0.056030 0.055600\\
237.850273 0.548438 0.056472 0.054933\\
305.836314 0.545312 0.051059 0.054025\\
456.386682 0.545312 0.055683 0.055345\\
};
\addlegendentry{$\tau=7$}
\end{axis}
\end{tikzpicture}
\end{minipage}
\begin{minipage}{0.49\linewidth}
\centering
\begin{tikzpicture}
\begin{axis}[width=\linewidth,height=4.2cm,scale only axis=false,
ylabel={Teacher acceptance rate},ymin=0,ymax=1,
axis lines=left,ymajorgrids=true,grid style={gray!20},
tick label style={font=\scriptsize},label style={font=\scriptsize},
legend style={font=\scriptsize,draw=none,at={(0.02,0.98)},anchor=north west},
title={Length-normalized log density ratio},title style={font=\small},xlabel={Centered $\log(\pi_{\mathcal E}/\pi_S)/|y|$}]
\addplot+[color=blue!75!black,mark=*,mark size=1.5pt,line width=0.8pt,error bars/.cd,y dir=both,y explicit] table[x=x,y=y,y error plus=plus,y error minus=minus,row sep=\\]{
x y plus minus\\
-1.397479 0.657813 0.038215 0.038841\\
-0.308017 0.732812 0.037765 0.033126\\
-0.054088 0.809375 0.036143 0.039380\\
0.031010 0.793750 0.040877 0.045435\\
0.089579 0.820313 0.036634 0.037580\\
0.150007 0.831250 0.032470 0.034330\\
0.212723 0.845313 0.034447 0.035409\\
0.286381 0.846875 0.028130 0.027710\\
0.391741 0.848437 0.029297 0.031838\\
0.598143 0.854688 0.036195 0.039371\\
};
\addlegendentry{$\tau=3$}
\addplot+[color=orange!85!black,mark=square*,mark size=1.5pt,line width=0.8pt,error bars/.cd,y dir=both,y explicit] table[x=x,y=y,y error plus=plus,y error minus=minus,row sep=\\]{
x y plus minus\\
-1.397479 0.657813 0.038215 0.038841\\
-0.308017 0.725000 0.037654 0.034365\\
-0.054088 0.807813 0.036716 0.039697\\
0.031010 0.784375 0.041145 0.044986\\
0.089579 0.817187 0.036506 0.037619\\
0.150007 0.828125 0.032236 0.033965\\
0.212723 0.840625 0.034380 0.035222\\
0.286381 0.843750 0.027894 0.027211\\
0.391741 0.845313 0.028817 0.032071\\
0.598143 0.851563 0.036773 0.039383\\
};
\addlegendentry{$\tau=5$}
\addplot+[color=teal!80!black,mark=triangle*,mark size=1.5pt,line width=0.8pt,error bars/.cd,y dir=both,y explicit] table[x=x,y=y,y error plus=plus,y error minus=minus,row sep=\\]{
x y plus minus\\
-1.397479 0.128125 0.030830 0.029899\\
-0.308017 0.164063 0.037531 0.031758\\
-0.054088 0.396875 0.053459 0.054984\\
0.031010 0.479687 0.061463 0.064568\\
0.089579 0.545312 0.054705 0.056205\\
0.150007 0.543750 0.052290 0.053061\\
0.212723 0.557813 0.054296 0.055562\\
0.286381 0.559375 0.052444 0.054051\\
0.391741 0.565625 0.047904 0.049606\\
0.598143 0.448437 0.064154 0.063536\\
};
\addlegendentry{$\tau=7$}
\end{axis}
\end{tikzpicture}
\end{minipage}
\caption{Verbal acceptance $a_\tau(y\mid q)=\Pr[S(y,q)\geq\tau]$ versus teacher-to-student log density ratios for 6,400 base-student answers (200 Math500 questions $\times$ 32 samples). Left: raw; right: length-normalized. Points show acceptance rates in ten equal-count bins after within-question centering; bars indicate 95\% question-bootstrap intervals. Teacher generation policy $\pi_{\mathcal E}$ is a proxy for $p_T$, not its definition.}
\label{fig:verbal_acceptance_likelihood_full}
\end{figure}

\begin{table*}[t]
\centering
\caption{Effect of the teacher's score range on one-problem training with full-response resampling. Scores are accuracy (\%), evaluated on AMD GPUs. Avg.\,10 and Avg.\,23 average the ten displayed benchmarks and all 23 benchmarks, respectively.}
\label{tab:pi1_score_scale}
\scriptsize
\setlength{\tabcolsep}{2.2pt}
\resizebox{\textwidth}{!}{%
\begin{tabular}{c*{12}{c}}
\toprule
\textbf{Steps} & \textbf{SVAMP} & \textbf{ASDiv} & \textbf{MAWPS} & \textbf{TABMWP} & \textbf{Minerva} & \textbf{Olympiad} & \textbf{SAT-Math} & \textbf{MMLU} & \textbf{Gaokao} & \textbf{AIME24} & \textbf{Avg.\,10} & \textbf{Avg.\,23} \\
\midrule
\rowcolor{black!5}\multicolumn{13}{l}{\textit{Score scale 0--5, reject if score $<4$}} \\
300 & 89.90 & 92.28 & 95.98 & 81.80 & 28.68 & 30.52 & 75.00 & 52.02 & 30.77 & 10.00 & 58.69 & 56.51 \\
500 & 90.30 & 91.96 & 96.22 & 81.50 & 28.31 & 28.89 & 71.88 & 51.16 & 32.97 & 16.67 & 58.99 & 56.95 \\
600 & 89.90 & 91.74 & 96.13 & 81.70 & 30.15 & 29.63 & 78.12 & 52.05 & 30.77 & 16.67 & 59.69 & 57.78 \\
800 & 89.50 & 92.01 & 96.66 & 82.80 & 27.57 & 28.59 & 81.25 & 51.86 & 36.26 & 13.33 & 59.98 & 57.55 \\
\midrule
\rowcolor{black!5}\multicolumn{13}{l}{\textit{Score scale 0--9 (baseline), reject if score $<7$}} \\
300 & 90.20 & 92.14 & 96.27 & 80.10 & 22.06 & 30.81 & 87.50 & 54.04 & 35.16 & 13.33 & 60.16 & 56.33 \\
500 & 90.20 & 91.83 & 96.80 & 82.30 & 26.84 & 31.11 & 78.12 & 54.41 & 31.87 & 16.67 & 60.01 & 56.92 \\
600 & 90.40 & 91.83 & 96.76 & 82.70 & 26.47 & 30.37 & 78.12 & 53.98 & 32.97 & 13.33 & 59.69 & 56.96 \\
800 & 90.00 & 92.42 & 96.42 & 83.60 & 25.74 & 31.56 & 75.00 & 53.08 & 29.67 & 10.00 & 58.75 & 57.81 \\
\midrule
\rowcolor{black!5}\multicolumn{13}{l}{\textit{Score scale 0--20, reject if score $<16$}} \\
300 & 90.60 & 91.83 & 96.22 & 80.80 & 26.10 & 31.11 & 81.25 & 53.25 & 32.97 & 6.67 & 59.08 & 54.82 \\
500 & 90.10 & 91.83 & 96.13 & 81.10 & 28.31 & 29.48 & 78.12 & 53.61 & 35.16 & 13.33 & 59.72 & 57.38 \\
600 & 89.90 & 92.05 & 96.27 & 81.30 & 26.10 & 30.07 & 78.12 & 52.98 & 34.07 & 10.00 & 59.09 & 56.62 \\
800 & 90.00 & 91.92 & 96.42 & 81.20 & 27.94 & 30.52 & 75.00 & 51.56 & 35.16 & 16.67 & 59.64 & 56.75 \\
\bottomrule
\end{tabular}%
}
\vspace{2pt}
\parbox{\textwidth}{\scriptsize All rows use actor-only evaluation with temperature 0 and top-$p$ 1.0. The 0--5 and 0--9 settings use single-token logprob scoring; 0--20 uses constrained integer output.}
\end{table*}

\subsection{Web Q\&A Rollout Format}
\label{sec:algorithm_details}

\paragraph{Web Q\&A serialization.}
A simplified rollout is:
\par\noindent
\begin{minipage}[t]{0.42\linewidth}
\vspace{0pt}
The closing markers give explicit segmentation points for query, returned-information, and reasoning spans; these define the coarse sub-trajectories for verbal scoring. The \texttt{<information>} span is inserted by the search environment and is masked during policy optimization.
\end{minipage}\hfill
\begin{minipage}[t]{0.56\linewidth}
\vspace{0pt}
\begin{tcolorbox}[
  width=\linewidth,
  colback=calloutback,
  colframe=calloutframe,
  colbacktitle=callouttitle,
  coltitle=black,
  fonttitle=\bfseries,
  title={Example Web Q\&A rollout},
  titlerule=0pt,
  boxrule=0.6pt,
  arc=1pt,
  boxsep=4pt,
  left=4pt,right=4pt,top=3pt,bottom=3pt,
  before skip=0pt,
  after skip=0pt,
]
\small
\emph{Question:} What city hosted the 2012 Summer Olympics?\\[-1pt]
\texttt{<think>} Need the host. \texttt{</think>}\\
\texttt{<search>} 2012 Olympics host \texttt{</search>}\\
\texttt{<information>} London. \texttt{</information>}\\
\texttt{<think>} The host is London. \texttt{</think>}\\
\texttt{<answer>} London \texttt{</answer>}
\end{tcolorbox}
\end{minipage}\par

\subsection{Additional Analysis and Score Granularity}
\label{sec:appendix_proofs}

\paragraph{Normalization of the Training Distribution.}
Fix a problem and a student prefix, and suppress this conditioning. Let $a_\tau(s)$ be the probability of accepting a candidate sub-trajectory $s$, including any randomness in score decoding, and let $Z_\tau=\sum_s\pi_S(s)a_\tau(s)$. For $Z_\tau>0$, the accepted distribution is $p_{\mathrm a,\tau}(s)=\pi_S(s)a_\tau(s)/Z_\tau$. Accepting the student candidate or replacing it after rejection gives
\begin{equation}
p_{\mathrm m,\tau}(s)
=\pi_S(s)a_\tau(s)+(1-Z_\tau)r_{\mathcal E,\tau}(s)
=Z_\tau p_{\mathrm a,\tau}(s)+(1-Z_\tau)r_{\mathcal E,\tau}(s).
\end{equation}
Here $r_{\mathcal E,\tau}$ is the normalized replacement distribution conditional on rejection at the fixed prefix, averaging over rejected candidates if replacements depend on them. The two components have masses $Z_\tau$ and $1-Z_\tau$, so their sum is normalized. When $Z_\tau=0$, the mixed distribution consists entirely of replacements; when $Z_\tau=1$, it equals the student distribution and no replacement distribution is needed.

\paragraph{Distribution Approximation and Policy Optimization.}
Proposition~\ref{prop:approx_density_ratio} bounds the next sub-trajectory distribution at a fixed prefix. Proposition~\ref{prop:complete_mixed_trajectories} extends this result to complete mixed trajectories when the local errors are controlled at every reachable prefix under a common sequential representation. The accumulated bound can become uninformative for long horizons unless local errors remain small. Approximation of the training distribution does not imply that policy optimization converges to it. In particular, the policy-gradient identity for $\pi_S$ cannot be applied directly to expected rewards under a teacher--student mixture. The clipped training objective in the main text is a surrogate evaluated on collected mixed trajectories; these distribution bounds do not establish unbiased gradients for the student's expected return, lower gradient variance, or a convergence rate.

\begin{figure}[!htb]
\noindent\textbf{Training Cost.}
Figure~\ref{fig:training_efficiency} reports cumulative wall-clock time across training checkpoints, complementing the per-step measurements in Figure~\ref{fig:datasize_training_time}.
\par\medskip
\centering
\definecolor{timeblue}{HTML}{1F77B4}
\definecolor{timeorange}{HTML}{FF7F0E}
\definecolor{timegreen}{HTML}{2CA02C}
\pgfplotsset{
 wall time/.style={width=0.94\linewidth,height=4.5cm,
 axis lines=box,axis line style={draw=black!75,line width=.45pt},
 grid=major,major grid style={dashed,line width=.3pt,draw=gray!30},
 tick label style={font=\tiny\sffamily},
 label style={font=\scriptsize\sffamily\bfseries},
 xlabel={Training steps},ylabel={Training time (hours)},
 xmin=270,xmax=840,xtick={300,500,600,800},
 legend style={font=\tiny\sffamily,at={(0.5,1.03)},anchor=south,
 legend columns=3,draw=gray!55,fill=white,rounded corners=1pt,inner sep=1.5pt}},
 time hes/.style={timeblue,line width=1pt,mark=*,mark size=1.8pt},
 time fr/.style={timeorange,line width=1pt,dashed,mark=square*,mark size=1.7pt},
 time rs/.style={timegreen,line width=1pt,dashdotted,mark=triangle*,mark size=1.8pt}}
\begin{minipage}[t]{0.485\textwidth}
\centering
\begin{tikzpicture}
\begin{axis}[wall time,ymin=10,ymax=52,ytick={10,20,30,40,50}]
\addplot[time hes] coordinates {(300,17.9) (500,29.8) (600,35.7) (800,47.6)};
\addlegendentry{OVD-HES}
\addplot[time fr] coordinates {(300,15.2) (500,25.3) (600,30.3) (800,40.4)};
\addlegendentry{OVD-FR}
\addplot[time rs] coordinates {(300,16.3) (500,27.1) (600,32.6) (800,43.4)};
\addlegendentry{OVD-RS}
\end{axis}
\end{tikzpicture}
\par\small (a) One fixed training problem.
\end{minipage}\hfill
\begin{minipage}[t]{0.485\textwidth}
\centering
\begin{tikzpicture}
\begin{axis}[wall time,ymin=20,ymax=130,ytick={20,40,60,80,100,120}]
\addplot[time hes] coordinates {(300,52.0) (500,77.3) (600,89.2) (800,111.3)};
\addlegendentry{OVD-HES}
\addplot[time fr] coordinates {(300,68.0) (500,92.4) (600,101.8) (800,120.2)};
\addlegendentry{OVD-FR}
\addplot[time rs] coordinates {(300,34.9) (500,55.1) (600,64.8) (800,83.5)};
\addlegendentry{OVD-RS}
\end{axis}
\end{tikzpicture}
\par\small (b) 128 distinct training problems.
\end{minipage}
\caption{Cumulative wall-clock training time under different resampling strategies. At 800 steps, OVD-HES and OVD-RS take 17.9\% and 7.5\% more time than OVD-FR with one fixed problem, but 7.4\% and 30.5\% less time with 128 problems. Panels use different vertical scales.}
\label{fig:training_efficiency}
\end{figure}

\subsubsection{Score Granularity: Mathematical-Reasoning Empirics}
\label{sec:score_granularity_empirical}

\noindent We examine the score vocabulary empirically using the same rejection-sampling interface. The $0$--$9$ scale is the default in the main experiments.

\begin{figure}[!htb]
\begin{minipage}[t]{0.45\textwidth}
\vspace{0pt}
\noindent Table~\ref{tab:pi1_score_scale} compares three score ranges for mathematical reasoning with full-trajectory resampling. Coarse verbal scores can provide an effective selection signal: at 800 steps, the $0$--$5$ scale outperforms the default $0$--$9$ scale on five benchmarks, with particularly large gains on Gaokao and SAT-Math. Figure~\ref{fig:score_scale_selected_gains} tracks average accuracy across training steps on these five benchmarks, selected by their gains over the default at this checkpoint. Their average improvement is 3.65 percentage points. Across all 23 benchmarks, however, the default reaches 57.81\%, compared with 57.55\% for $0$--$5$ and 56.75\% for $0$--$20$. Thus, expanding the score range does not consistently improve performance. The $0$--$20$ setting also changes the scoring format, so this comparison does not isolate score granularity alone.
\end{minipage}\hfill
\begin{minipage}[t]{0.52\textwidth}
\vspace{0pt}
\centering
\begin{tikzpicture}
\begin{axis}[
width=\linewidth,height=4.3cm,
xlabel={Training steps},ylabel={Average accuracy (\%)},
xtick={300,500,600,800},
grid=major,major grid style={dashed,gray!30},
tick label style={font=\scriptsize},label style={font=\scriptsize},
legend style={font=\scriptsize,at={(0.5,1.03)},anchor=south,legend columns=3,draw=none},
]
\addplot[blue!75!black,mark=*,line width=1pt] coordinates {(300,48.086) (500,49.210) (600,50.368) (800,51.014)};
\addlegendentry{0--5}
\addplot[orange!85!black,mark=square*,dashed,line width=1pt] coordinates {(300,50.864) (500,50.060) (600,49.530) (800,47.366)};
\addlegendentry{0--9}
\addplot[teal!80!black,mark=triangle*,dashdotted,line width=1pt] coordinates {(300,48.642) (500,50.210) (600,48.912) (800,50.238)};
\addlegendentry{0--20}
\end{axis}
\end{tikzpicture}
\captionsetup{font=small,skip=2pt}
\caption{Average accuracy on Gaokao, SAT-Math, AIME24, Minerva, and MAWPS across training checkpoints. These five benchmarks were selected from the ten displayed in Table~\ref{tab:pi1_score_scale} by the largest gains of $0$--$5$ over $0$--$9$ at 800 steps; the same set is used for every curve and checkpoint. Rejection thresholds are 4, 7, and 16 for the $0$--$5$, $0$--$9$, and $0$--$20$ scales, respectively. Complete results and scoring formats are reported in Table~\ref{tab:pi1_score_scale}.}
\label{fig:score_scale_selected_gains}
\vspace{-0.6\baselineskip}
\end{minipage}
\end{figure}

\FloatBarrier

\subsection{Additional Mathematical-Reasoning Results}
\label{sec:additional_math_results}

\paragraph{Random-Suffix Resampling Control.}
Table~\ref{tab:fixed_nine_math} reports AMC23 alongside an eight-benchmark subset selected post hoc from 23 tasks, with at most one Gaokao benchmark, to maximize the smaller of the single-problem and 128-problem best-checkpoint HES gains over RLVR across steps 300, 500, 600, and 800. The resulting nine-benchmark set is fixed across all rows. AMC23 scores are restored from the earlier evaluation summary. The nine-benchmark averages are reconstructed as $(8A_8 + A_{\mathrm{AMC23}})/9$, where $A_8$ is the previously reported eight-benchmark average; their last decimal may be affected by rounding.

Table~\ref{tab:fixed_nine_math_rs} reports \textsc{OVD}-RS after training on 128 distinct problems, on the same nine benchmarks and checkpoints as Table~\ref{tab:fixed_nine_math}. At 800 steps, its macro average is 56.34\%, compared with 53.77\% for \textsc{OVD}-HES. The ordering varies across checkpoints, so these results do not establish a consistent macro-average advantage from selecting a high-entropy suffix over a random suffix.

\begin{table*}[!t]
\centering
\caption{Random-suffix resampling on 128 training problems. Benchmarks and averaging follow Table~\ref{tab:fixed_nine_math}.}
\vspace{-2mm}
\label{tab:fixed_nine_math_rs}
\scriptsize
\setlength{\tabcolsep}{2.2pt}
\resizebox{\textwidth}{!}{%
\begin{tabular}{lccccccccccc}
\toprule
\textbf{Method} & \textbf{Steps} & \textbf{AIME24} & \textbf{AMC23} & \textbf{CARP} & \textbf{Middle} & \textbf{College} & \textbf{Gaokao-II} & \textbf{MATH} & \textbf{MAWPS} & \textbf{SAT-Math} & \textbf{Avg.} \\
\midrule
\multicolumn{12}{l}{\textit{Training on 128 Distinct Problems}} \\
\textsc{OVD}-RS & 300 & 16.7 & 47.5 & 52.8 & 44.6 & 35.8 & 42.9 & 68.2 & 93.0 & 81.3 & 53.62 \\
\textsc{OVD}-RS & 500 & 6.7 & 52.5 & 53.8 & 46.5 & 36.6 & 50.0 & 69.4 & 93.6 & 78.1 & 54.13 \\
\textsc{OVD}-RS & 600 & 16.7 & 60.0 & 54.4 & 44.6 & 38.0 & 50.0 & 70.7 & 95.8 & 71.9 & 55.78 \\
\textsc{OVD}-RS & 800 & 16.7 & 57.5 & 53.9 & 45.5 & 39.0 & 42.9 & 70.8 & 96.3 & 84.4 & \textbf{56.34} \\
\bottomrule
\end{tabular}%
}
\end{table*}

\begin{table*}[!t]
\centering
\caption{Sampling results for the three resampling strategies after training on the fixed 128-problem DeepScaleR subset, with $(\tau_{\mathrm{tr}},\tau_{\mathrm{te}})=(7,0)$. M@$k$ is majority-vote accuracy; tied votes count as incorrect. P@$k$ counts a question as correct if any of the $k$ answers is correct. We sample 16 answers per question and use a fixed-seed random subset for $k=8$. Temperature is 0.6 and top-$p$ is 1.0.}
\label{tab:dsr128_passk_full}
\setlength{\tabcolsep}{1.8pt}
\resizebox{\textwidth}{!}{%
\begin{tabular}{c*{11}{cc}}
\toprule
\multicolumn{23}{l}{\textit{(a) Majority-vote accuracy}} \\
\midrule
\multirow{2}{*}{\textbf{Steps}} & \multicolumn{2}{c}{\textbf{SVAMP}} & \multicolumn{2}{c}{\textbf{ASDiv}} & \multicolumn{2}{c}{\textbf{MAWPS}} & \multicolumn{2}{c}{\textbf{TABMWP}} & \multicolumn{2}{c}{\textbf{Minerva}} & \multicolumn{2}{c}{\textbf{Olympiad}} & \multicolumn{2}{c}{\textbf{SAT-Math}} & \multicolumn{2}{c}{\textbf{MMLU}} & \multicolumn{2}{c}{\textbf{Gaokao}} & \multicolumn{2}{c}{\textbf{AIME24}} & \multicolumn{2}{c}{\textbf{Avg.}} \\
\cmidrule(lr){2-3} \cmidrule(lr){4-5} \cmidrule(lr){6-7} \cmidrule(lr){8-9} \cmidrule(lr){10-11} \cmidrule(lr){12-13} \cmidrule(lr){14-15} \cmidrule(lr){16-17} \cmidrule(lr){18-19} \cmidrule(lr){20-21} \cmidrule(lr){22-23}
 & \textbf{M@8} & \textbf{M@16} & \textbf{M@8} & \textbf{M@16} & \textbf{M@8} & \textbf{M@16} & \textbf{M@8} & \textbf{M@16} & \textbf{M@8} & \textbf{M@16} & \textbf{M@8} & \textbf{M@16} & \textbf{M@8} & \textbf{M@16} & \textbf{M@8} & \textbf{M@16} & \textbf{M@8} & \textbf{M@16} & \textbf{M@8} & \textbf{M@16} & \textbf{M@8} & \textbf{M@16} \\
\midrule
\rowcolor{black!5}\multicolumn{23}{l}{\textsc{OVD}-FR} \\
300 & 91.0 & 91.6 & 93.0 & 93.8 & 97.0 & 97.6 & 88.5 & 89.8 & 32.7 & 34.9 & 36.6 & 39.4 & 81.2 & 90.6 & 53.5 & 56.5 & 36.3 & 40.7 & 20.0 & 26.7 & 63.0 & 66.1 \\
500 & 92.2 & 92.1 & 93.4 & 94.0 & 97.1 & 97.2 & 89.2 & 90.5 & 34.6 & 35.3 & 34.4 & 37.2 & 87.5 & 90.6 & 55.8 & 58.2 & 35.2 & 46.2 & 20.0 & 20.0 & 63.9 & 66.1 \\
600 & 92.0 & 93.0 & 93.2 & 93.7 & 97.0 & 97.4 & 88.6 & 90.2 & 35.3 & 38.2 & 34.4 & 36.6 & 84.4 & 84.4 & 54.7 & 57.7 & 39.6 & 49.5 & 20.0 & 23.3 & 63.9 & 66.4 \\
800 & 91.7 & 92.0 & 93.1 & 93.2 & 97.0 & 97.1 & 87.9 & 89.9 & 36.0 & 37.5 & 33.0 & 35.4 & 84.4 & 81.2 & 54.8 & 57.3 & 42.9 & 44.0 & 23.3 & 23.3 & 64.4 & 65.1 \\
\midrule
\rowcolor{black!5}\multicolumn{23}{l}{\textsc{OVD}-HES} \\
300 & 91.3 & 92.2 & 93.7 & 94.0 & 97.2 & 97.4 & 88.8 & 90.9 & 33.1 & 37.5 & 36.4 & 38.8 & 78.1 & 87.5 & 54.9 & 57.3 & 40.7 & 39.6 & 26.7 & 30.0 & 64.1 & 66.5 \\
500 & 91.5 & 91.9 & 93.0 & 93.3 & 97.2 & 97.5 & 87.1 & 89.6 & 31.2 & 34.6 & 36.0 & 37.9 & 81.2 & 87.5 & 53.2 & 56.4 & 41.8 & 46.2 & 20.0 & 23.3 & 63.2 & 65.8 \\
600 & 92.2 & 92.2 & 93.3 & 94.0 & 97.3 & 97.5 & 90.5 & 91.3 & 36.0 & 36.4 & 36.0 & 37.9 & 87.5 & 93.8 & 53.3 & 56.8 & 45.1 & 41.8 & 23.3 & 26.7 & 65.5 & 66.8 \\
800 & 91.8 & 92.4 & 93.0 & 93.6 & 97.2 & 97.6 & 88.7 & 90.1 & 30.9 & 34.6 & 34.1 & 37.3 & 84.4 & 84.4 & 53.7 & 56.7 & 35.2 & 45.1 & 23.3 & 23.3 & 63.2 & 65.5 \\
\midrule
\rowcolor{black!5}\multicolumn{23}{l}{\textsc{OVD}-RS} \\
300 & 91.4 & 92.0 & 93.0 & 93.8 & 97.4 & 97.7 & 88.4 & 89.5 & 31.2 & 36.8 & 36.3 & 38.2 & 90.6 & 93.8 & 54.3 & 57.1 & 36.3 & 40.7 & 20.0 & 23.3 & 63.9 & 66.3 \\
500 & 92.2 & 92.9 & 93.6 & 94.0 & 97.1 & 97.6 & 88.8 & 90.6 & 32.7 & 34.2 & 34.8 & 36.0 & 84.4 & 90.6 & 54.8 & 56.8 & 38.5 & 41.8 & 23.3 & 26.7 & 64.0 & 66.1 \\
600 & 91.7 & 92.3 & 93.8 & 94.3 & 97.2 & 97.7 & 89.6 & 90.4 & 29.0 & 33.1 & 36.6 & 37.5 & 84.4 & 81.2 & 54.4 & 57.5 & 40.7 & 40.7 & 26.7 & 23.3 & 64.4 & 64.8 \\
800 & 92.4 & 92.7 & 93.4 & 93.9 & 97.6 & 97.5 & 87.4 & 89.4 & 28.3 & 32.0 & 35.4 & 36.4 & 84.4 & 90.6 & 54.4 & 56.8 & 36.3 & 40.7 & 20.0 & 23.3 & 63.0 & 65.3 \\
\bottomrule
\end{tabular}%
}
\vspace{0.15em}

\resizebox{\textwidth}{!}{%
\begin{tabular}{c*{11}{cc}}
\multicolumn{23}{l}{\textit{(b) Pass@$k$}} \\
\midrule
\multirow{2}{*}{\textbf{Steps}} & \multicolumn{2}{c}{\textbf{SVAMP}} & \multicolumn{2}{c}{\textbf{ASDiv}} & \multicolumn{2}{c}{\textbf{MAWPS}} & \multicolumn{2}{c}{\textbf{TABMWP}} & \multicolumn{2}{c}{\textbf{Minerva}} & \multicolumn{2}{c}{\textbf{Olympiad}} & \multicolumn{2}{c}{\textbf{SAT-Math}} & \multicolumn{2}{c}{\textbf{MMLU}} & \multicolumn{2}{c}{\textbf{Gaokao}} & \multicolumn{2}{c}{\textbf{AIME24}} & \multicolumn{2}{c}{\textbf{Avg.}} \\
\cmidrule(lr){2-3} \cmidrule(lr){4-5} \cmidrule(lr){6-7} \cmidrule(lr){8-9} \cmidrule(lr){10-11} \cmidrule(lr){12-13} \cmidrule(lr){14-15} \cmidrule(lr){16-17} \cmidrule(lr){18-19} \cmidrule(lr){20-21} \cmidrule(lr){22-23}
 & \textbf{P@8} & \textbf{P@16} & \textbf{P@8} & \textbf{P@16} & \textbf{P@8} & \textbf{P@16} & \textbf{P@8} & \textbf{P@16} & \textbf{P@8} & \textbf{P@16} & \textbf{P@8} & \textbf{P@16} & \textbf{P@8} & \textbf{P@16} & \textbf{P@8} & \textbf{P@16} & \textbf{P@8} & \textbf{P@16} & \textbf{P@8} & \textbf{P@16} & \textbf{P@8} & \textbf{P@16} \\
\midrule
\rowcolor{black!5}\multicolumn{23}{l}{\textsc{OVD}-FR} \\
300 & 96.5 & 97.6 & 96.9 & 97.5 & 98.6 & 98.8 & 94.9 & 95.9 & 49.3 & 55.5 & 50.9 & 56.4 & 96.6 & 96.9 & 89.7 & 95.6 & 70.7 & 76.9 & 33.1 & 40.0 & 77.7 & 81.1 \\
500 & 96.1 & 96.9 & 96.8 & 97.4 & 98.6 & 98.7 & 95.4 & 96.6 & 49.3 & 55.5 & 50.8 & 56.6 & 98.2 & 100.0 & 89.0 & 94.8 & 70.4 & 74.7 & 32.3 & 36.7 & 77.7 & 80.8 \\
600 & 96.3 & 97.2 & 96.7 & 97.2 & 98.5 & 98.7 & 95.8 & 96.8 & 53.1 & 60.3 & 50.0 & 56.0 & 95.4 & 96.9 & 88.7 & 94.9 & 70.4 & 74.7 & 35.7 & 43.3 & 78.1 & 81.6 \\
800 & 95.9 & 96.7 & 96.5 & 97.0 & 98.5 & 98.7 & 95.0 & 96.1 & 49.9 & 54.4 & 49.2 & 54.5 & 95.2 & 100.0 & 87.9 & 94.1 & 70.1 & 75.8 & 27.9 & 30.0 & 76.6 & 79.7 \\
\midrule
\rowcolor{black!5}\multicolumn{23}{l}{\textsc{OVD}-HES} \\
300 & 96.1 & 97.2 & 96.9 & 97.2 & 98.7 & 98.9 & 95.5 & 96.8 & 53.1 & 59.2 & 52.9 & 59.6 & 96.0 & 96.9 & 89.3 & 95.2 & 71.0 & 76.9 & 32.7 & 36.7 & 78.2 & 81.5 \\
500 & 96.2 & 97.3 & 96.9 & 97.6 & 98.7 & 98.9 & 95.1 & 96.5 & 47.0 & 52.6 & 50.9 & 56.4 & 98.8 & 100.0 & 89.2 & 95.3 & 69.8 & 75.8 & 32.2 & 36.7 & 77.5 & 80.7 \\
600 & 96.0 & 96.9 & 96.9 & 97.4 & 98.7 & 98.9 & 95.7 & 96.6 & 52.1 & 56.6 & 51.2 & 56.9 & 96.8 & 96.9 & 89.1 & 95.1 & 66.1 & 71.4 & 34.1 & 40.0 & 77.7 & 80.7 \\
800 & 96.1 & 97.1 & 96.8 & 97.3 & 98.7 & 98.9 & 95.4 & 96.4 & 44.9 & 50.7 & 51.1 & 56.6 & 95.2 & 96.9 & 88.7 & 94.7 & 67.0 & 72.5 & 37.4 & 43.3 & 77.1 & 80.4 \\
\midrule
\rowcolor{black!5}\multicolumn{23}{l}{\textsc{OVD}-RS} \\
300 & 95.9 & 96.7 & 97.0 & 97.6 & 98.6 & 98.8 & 95.3 & 96.4 & 49.8 & 54.8 & 52.0 & 57.5 & 96.8 & 96.9 & 89.1 & 94.9 & 70.4 & 75.8 & 33.3 & 40.0 & 77.8 & 80.9 \\
500 & 96.1 & 97.2 & 97.0 & 97.5 & 98.6 & 98.8 & 95.6 & 96.9 & 47.5 & 53.7 & 50.4 & 56.1 & 98.3 & 100.0 & 89.6 & 95.3 & 68.4 & 75.8 & 31.4 & 36.7 & 77.3 & 80.8 \\
600 & 96.7 & 97.8 & 97.2 & 97.7 & 98.8 & 98.9 & 95.2 & 96.3 & 43.8 & 49.3 & 49.9 & 54.7 & 98.3 & 100.0 & 89.0 & 94.9 & 68.3 & 73.6 & 31.6 & 33.3 & 76.9 & 79.7 \\
800 & 96.1 & 96.9 & 97.0 & 97.4 & 98.7 & 98.7 & 95.6 & 96.5 & 49.0 & 55.9 & 49.8 & 54.8 & 95.2 & 96.9 & 89.0 & 94.9 & 68.8 & 75.8 & 31.6 & 33.3 & 77.1 & 80.1 \\
\bottomrule
\end{tabular}%
}
\end{table*}

\paragraph{Majority Voting and Solution Coverage.}
Table~\ref{tab:dsr128_passk_full} reports results on all ten benchmarks after training on 128 distinct problems. M@$k$ measures the accuracy of the majority-selected answer, whereas P@$k$ measures whether the sampled set contains at least one correct answer. At 800 steps, \textsc{OVD}-HES reaches 80.4\% average P@16, compared with 79.7\% for \textsc{OVD}-FR and 80.1\% for \textsc{OVD}-RS; their M@16 averages are 65.5\%, 65.1\%, and 65.3\%, respectively. The P@16 advantage of \textsc{OVD}-HES is therefore smaller across all ten benchmarks than on the five-benchmark subset shown in Figure~\ref{fig:hes_sampling_collapse}. The ordering also varies across checkpoints: at 600 steps, \textsc{OVD}-FR has the highest average P@16 (81.6\%). These results distinguish solution coverage from majority-vote accuracy and show that the benefit of preserving student prefixes depends on the benchmark and training checkpoint.

Table~\ref{tab:scaling_rl_two_example} reports the complete two-example scaling results omitted from the main-text comparison.

\begin{table*}[t]
\centering
\caption{Math accuracy after training on two problems. All methods use Qwen2.5-Math-1.5B and are evaluated without a teacher. OPD$^\ast$ denotes token-level distillation at 800 steps.}
\label{tab:scaling_rl_two_example}
\scriptsize
\setlength{\tabcolsep}{2.2pt}
\renewcommand{\arraystretch}{1.2}
\resizebox{\textwidth}{!}{%
\begin{tabular}{lcccccccccccc}
\toprule
\textbf{Method} & \textbf{Steps} & \textbf{SVAMP} & \textbf{ASDiv} & \textbf{MAWPS} & \textbf{TABMWP} & \textbf{Minerva} & \textbf{Olympiad} & \textbf{SAT-Math} & \textbf{MMLU} & \textbf{Gaokao} & \textbf{AIME24} & \textbf{Avg.} \\
\midrule
\rowcolor{black!5}\multicolumn{13}{l}{\textit{Scaling RLVR (2 random examples)}} \\
Two-example & 300 & 73.7 & 76.2 & 80.4 & 70.8 & 16.5 & 30.2 & 81.2 & 51.1 & 35.7 & 6.7 & 52.2 \\
Two-example & 500 & 77.1 & 80.0 & 86.0 & 75.17 & 16.5 & 27.7 & 81.2 & 51.3 & 42.9 & 13.3 & 55.1 \\
Two-example & 600 & 75.9 & 78.1 & 84.6 & 75.6 & 16.5 & 31.1 & 84.4 & 49.6 & 35.7 & 13.3 & 54.5 \\
Two-example & 800 & 81.1 & 80.0 & 87.6 & 73.7 & 15.8 & 29.6 & 81.2 & 49.0 & 35.7 & 13.3 & 54.7 \\
OPD$^\ast$ (reverse KL) & 800 & 80.5 & 79.3 & 87.0 & 73.2 & 15.6 & 29.3 & 80.8 & 48.9 & 35.4 & 13.0 & 53.8 \\
OPD$^\ast$ (top-$k$ logits) & 800 & 81.5 & 80.3 & 88.0 & 74.0 & 16.0 & 29.8 & 81.5 & 49.2 & 36.0 & 13.6 & 55.1 \\
\midrule
\rowcolor{black!5}\multicolumn{13}{l}{\textit{\textsc{OVD} (2 random examples, $\tau_{\mathrm{tr}}{=}7$, $\tau_{\mathrm{te}}{=}0$)}} \\
\textsc{OVD} & 300 & 78.6 & 81.8 & 87.1 & 72.3 & 15.1 & 29.5 & 81.2 & 50.2 & 50.0 & 6.7 & 55.2 \\
\textsc{OVD} & 500 & 79.8 & 81.1 & 87.7 & 74.9 & 17.3 & 29.3 & 75.0 & 50.3 & 35.7 & 10.0 & 54.1 \\
\textsc{OVD} & 600 & 76.3 & 77.7 & 82.6 & 74.0 & 17.3 & 30.4 & 87.5 & 50.6 & 28.6 & 10.0 & 53.5 \\
\textsc{OVD} & 800 & \textbf{83.1} & \textbf{83.3} & \textbf{90.6} & \textbf{77.6} & \textbf{18.0} & \textbf{31.6} & \textbf{84.4} & 49.4 & 35.7 & 13.3 & \textbf{56.7} \\
\bottomrule
\end{tabular}%
}
\end{table*}

\FloatBarrier

\subsection{Illustrative Mathematical-Reasoning Problems}

The box accompanying Figure~\ref{fig:valid_search} shows two example problems illustrating the mathematical-reasoning training setting.

\subsection{Reward Convergence Across Training Steps}

\begin{figure}[t!]
    \centering
    \begin{minipage}[t]{0.48\linewidth}
        \centering
        \includegraphics[width=\linewidth]{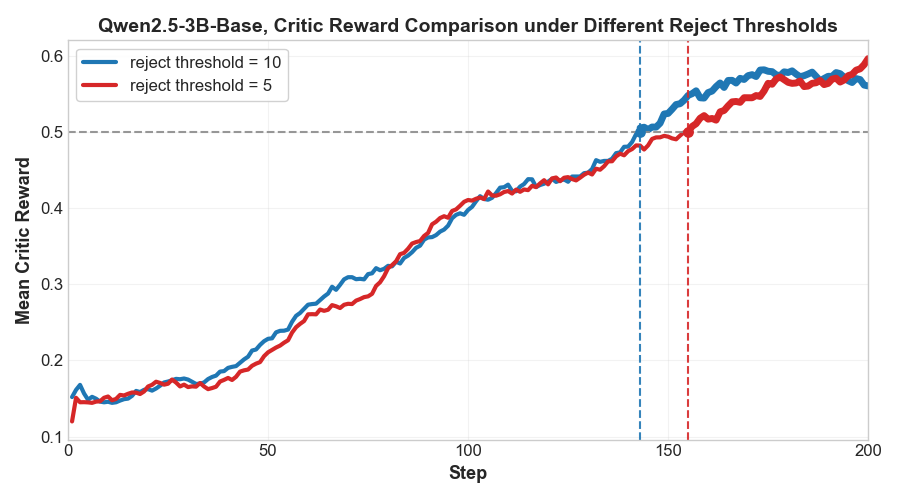}
        \caption{Reward convergence on Qwen-2.5-7B across training steps under different \emph{training-time} thresholds $\tau_{\mathrm{tr}}$. Dashed horizontal line: reward $0.5$; vertical markers: first $0.5$ crossing for each $\tau_{\mathrm{tr}}$ curve.}
        \label{fig:reward_convergence_qwen}
    \end{minipage}
    \hfill
    \begin{minipage}[t]{0.48\linewidth}
        \centering
        \includegraphics[width=\linewidth]{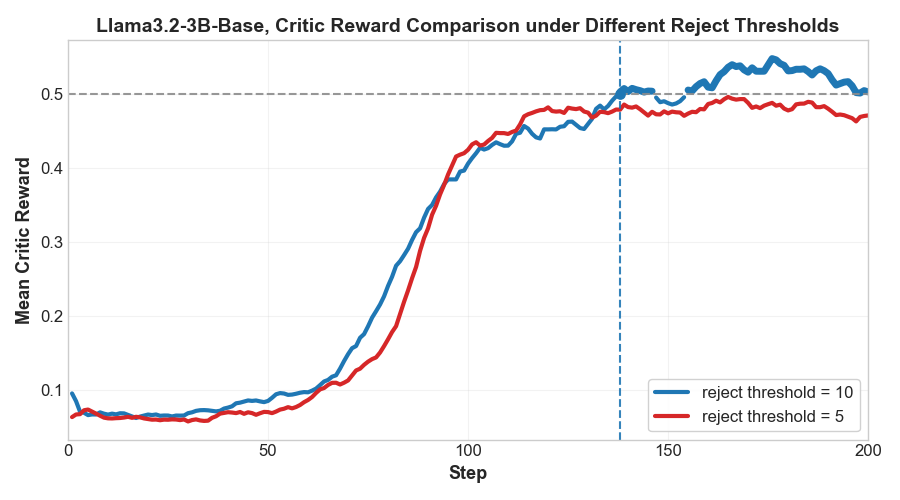}
        \caption{Reward convergence on LLaMA-3.2-3B across training steps under different \emph{training-time} thresholds $\tau_{\mathrm{tr}}$. Dashed horizontal line: reward $0.5$; vertical marker: $\tau_{\mathrm{tr}}{=}10$ first crossing $0.5$.}
        \label{fig:reward_convergence_llama}
    \end{minipage}
\end{figure}

We analyze the reward convergence behavior of \textsc{OVD} across different training steps while varying the \emph{training-time} verbal rejection threshold $\tau_{\mathrm{tr}}$ (the test-time gate $\tau_{\mathrm{te}}$ used when reporting EM is fixed and not swept in these curves). Figure~\ref{fig:reward_convergence_qwen} and Figure~\ref{fig:reward_convergence_llama} illustrate the reward trajectories for Qwen-2.5-7B and LLaMA-3.2-3B models respectively, demonstrating that appropriate training threshold settings lead to faster convergence and higher final rewards. The results show that $\tau_{\mathrm{tr}}$ significantly impacts learning dynamics, with moderate values (5--7) typically achieving the best balance between exploration and exploitation.

\subsection{Valid Search Across Training Steps}
\label{sec:valid_search}

We report the \emph{valid search} metric during Web Q\&A training, which measures the average number of search interactions per episode that are considered valid by the simulated environment (higher values indicate more consistent and well-formed tool-use behavior).

\begin{figure}[!htb]
\begin{minipage}[t]{0.49\textwidth}
\vspace{0pt}
\begin{tcolorbox}[
	width=\linewidth,
	colframe=calloutframe,
	colback=calloutback,
	boxrule=0.8pt,
	arc=0pt,
	boxsep=6pt,
	colbacktitle=callouttitle,
	coltitle=black,
	fonttitle=\bfseries,
	title={Illustrative Mathematical-Reasoning Problems},
	titlerule=0pt,
    before skip=0pt,
    fontupper=\small,
]
\textbf{Problem 1 :} The lock opens only if a specific three-digit number is entered. An attempt consists of randomly selecting three digits from a given set of five. The code was guessed correctly only on the last of all attempts. How many attempts preceded the successful one? Let's think step by step and output the final answer within \boxed{}.\\[0.5em]
\textbf{Answer:} 124\\[1em]
\textbf{Problem 2 :} For the ellipse $25x^2 - 100x + 4y^2 + 8y + 16 = 0,$ find the distance between the foci. Let's think step by step and output the final answer within \boxed{}.\\[0.5em]
\textbf{Answer:} $\frac{2\sqrt{462}}{5}$
\end{tcolorbox}
\end{minipage}\hfill
\begin{minipage}[t]{0.48\textwidth}
\vspace{0pt}
    \centering
    \includegraphics[width=\linewidth]{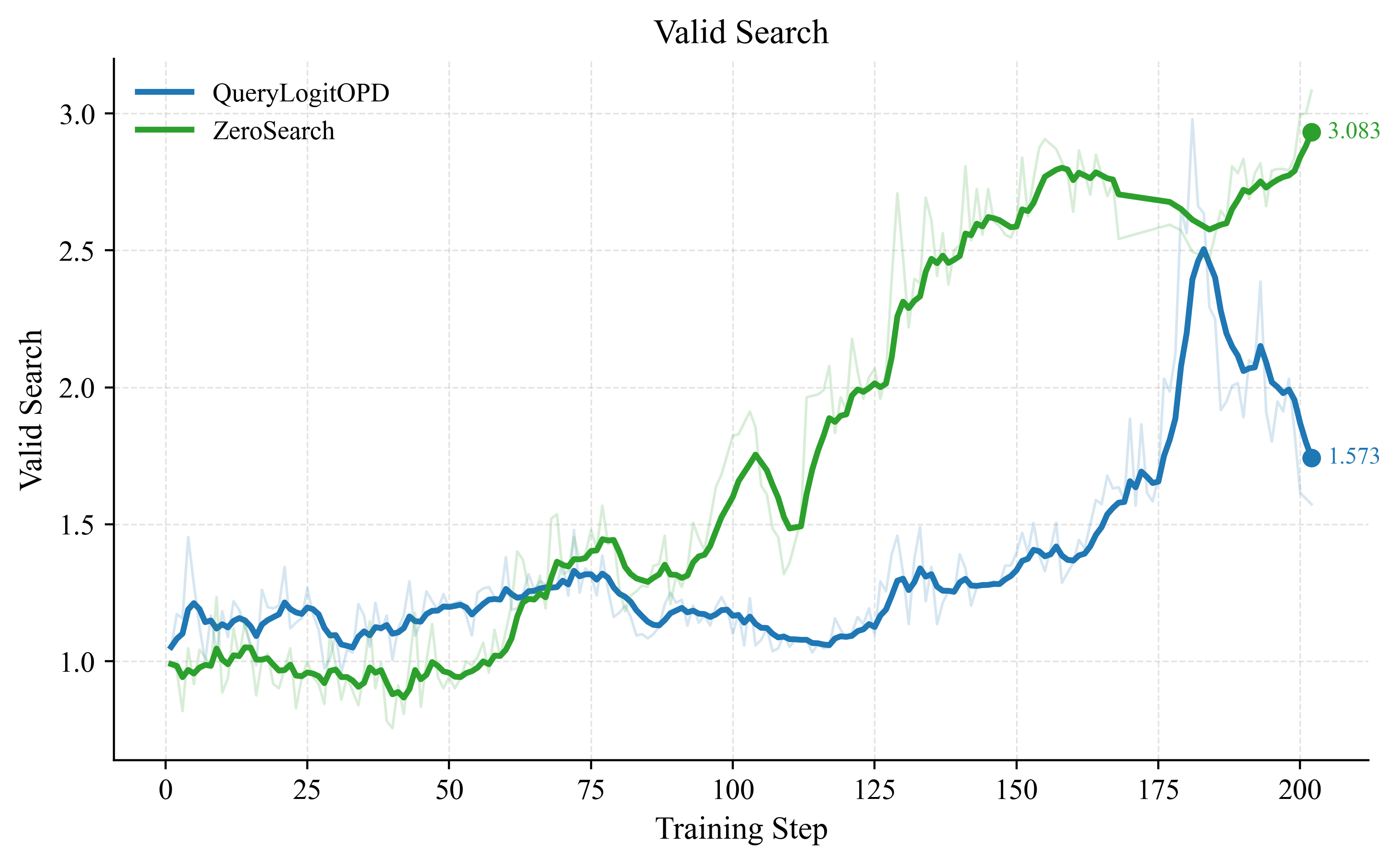}
    \captionsetup{font=small,skip=2pt}
    \caption{\textbf{Valid search count across training steps.} \textsc{OVD} (green curve) maintains higher valid-search rates throughout training compared to QueryLogitOPD (blue curve). The gap widens as training progresses, demonstrating that verbal step-level feedback helps the student policy learn more consistent search behaviors without requiring dense token-level logit matching at every query generation step.}
    \label{fig:valid_search}
    \end{minipage}
\end{figure}

Figure~\ref{fig:valid_search} compares \textsc{OVD} and a token-level OPD baseline (QueryLogitOPD, using top-$k$ query logits for on-policy distillation) over the same training horizon.

\FloatBarrier
\subsection{Verbal Scoring Prompt}
\label{sec:verbal_prompt}

We present the detailed prompt template used for verbal scoring in \textsc{OVD}. The prompt instructs the teacher model to evaluate the quality of each reasoning step on a discrete 0--9 scale, providing structured feedback without requiring access to token-level probability distributions. This design enables step-level quality assessment while maintaining compatibility with black-box teacher models. The scoring rubric divides the scale into four meaningful ranges: incorrect/irrelevant (0--2), partially correct (3--5), correct and well-reasoned (6--8), and optimal (9), allowing the teacher to express nuanced judgments about reasoning quality. By outputting only a single score token, this approach dramatically reduces memory overhead compared to full-vocabulary logit storage, as analyzed in Section~\ref{subsec:token_opd_limitations}.

\begin{tcolorbox}[
	width=\linewidth,
	colframe=calloutframe,
	colback=calloutback,
	boxrule=0.8pt,
	arc=0pt,
	boxsep=5pt,
	leftupper=6pt,rightupper=6pt,
	leftlower=6pt,rightlower=6pt,
	top=2pt,bottom=2pt,
	colbacktitle=callouttitle,
	coltitle=black,
	fonttitle=\bfseries,
	title={Verbal Scoring Prompt},
	titlerule=0pt,
	before skip=4pt,
	after skip=4pt,
]
\small
\textbf{Problem:} $q$\\[0.25em]
\textbf{Previous Steps:} $s_1, s_2, \ldots, s_{t-1}$\\[0.25em]
\textbf{Current Step:} $s_t$\\[0.4em]
Rate the quality of the current reasoning step on a scale of 0--9, where:\\
\begin{itemize}\setlength{\itemsep}{2pt}\setlength{\parsep}{0pt}\setlength{\topsep}{4pt}
\item 0--2: Incorrect or irrelevant
\item 3--5: Partially correct but incomplete
\item 6--8: Correct and well-reasoned
\item 9: Optimal step toward the solution
\end{itemize}
\textbf{Output only a single number from 0 to 9.}
\end{tcolorbox}
\FloatBarrier

\subsection{Hyperparameters and Implementation Details}
\label{sec:hyperparameters}

\subsubsection{Implementation Details}
We provide detailed hyperparameter configurations and implementation details to ensure the reproducibility of our experiments.

\paragraph{Computational Infrastructure.}
All experiments are conducted on a high-performance computing cluster equipped with 8 AMD Instinct MI210 accelerators, each featuring 64\,GB of HBM2e memory. The GPUs are based on the CDNA~2 microarchitecture (gfx90a). Distributed training is performed using 4 GPUs per training instance. GPU acceleration is supported via the ROCm platform.

\paragraph{Teacher Models.}
\label{sec:simulator}
\textsc{OVD} uses a task-specific teacher $\pi_\mathcal{E}$ to score student sub-trajectories and resample low-scoring ones through the verbal interface in Section~\ref{sec:verbal_scoring}. Teacher-side verbal scoring during training is served on NVIDIA A100 GPUs (40\,GB) for low-latency inference alongside student rollouts. For \emph{Web Q\&A}, simulated search and scoring use \texttt{sunhaonlp/Simulation\_LLM\_google\_7B\_V2}; for \emph{mathematical reasoning}, the critic is \path{deepseek-ai/DeepSeek-R1-Distill-Qwen-7B}~\citep{deepseek2025r1}, applied at each reasoning step in the same serving regime.

\paragraph{Baselines}
To evaluate the effectiveness of \textsc{OVD}, we compare our method with the following baselines. (i) \emph{Vanilla Prompting Methods}: This category includes Direct Answer, Chain-of-Thought (CoT)~\citep{wei2022chain}, and standard Retrieval-Augmented Generation (RAG)~\citep{lewis2020retrieval}. Direct Answer generates responses without explicit reasoning, CoT encourages step-by-step reasoning, and RAG retrieves relevant documents before answering. (ii) \emph{Advanced Search Methods}: We consider RA-Agent and Search-o1~\citep{li2025search}, which iteratively search for relevant information and perform multi-step reasoning with retrieved context. (iii) \emph{RL-Based Reasoning Methods}: This category includes R1~\citep{deepseek2025r1}, Search-R1~\citep{jin2025searchr1}, and ZeroSearch~\citep{sun2025zerosearch}. In R1, the student policy is trained to perform in-depth reasoning based solely on its internal knowledge. Search-R1 enables the student policy to interact with a real search engine during inference. ZeroSearch learns search-like reasoning without actual retrieval through simulated search environments. (iv) \emph{Additional search baselines (Qwen-2.5-3B-Base)}: We further include SmartSearch~\citep{wen2026smartsearch} and StepSearch~\citep{zheng2025stepsearch} in Table~\ref{tab:qa_em_results} under the same evaluation protocol.

\subsubsection{Hyperparameters}

\paragraph{OPD$^\ast$ top-$k$ logit baseline.}
Token-level OPD$^\ast$ rows with \emph{top-$k$ logits} (Table~\ref{tab:qa_em_results} and Table~\ref{tab:scaling_rl_two_example}) use the truncated support described in paragraph~\nameref{para:memory_bottleneck}: at each decoding step we retain only the teacher's $k$ highest-probability tokens, restrict both the student and teacher distributions to that support, and renormalize them before computing the reverse-KL objective. We fix $k{=}8192$ for all reported top-$k$ runs, which matches the top-$k$ memory curves in Figure~\ref{fig:memory_scaling} under $V{\approx}152\mathrm{K}$ (Qwen2.5-7B-class vocabulary).

We employ different hyperparameters for trajectory scoring depending on the task type.

\textbf{Mathematical Reasoning.} For math tasks, we use \path{deepseek-ai/DeepSeek-R1-Distill-Qwen-7B}~\citep{deepseek2025r1} as the teacher model. The teacher evaluates reasoning trajectories by computing verbal scores on a 0--9 scale based on the prompt template described in Appendix~\ref{sec:verbal_prompt}. By default, we greedily decode the highest-probability score token; score-vocabulary sampling variants use a temperature of 0.7. The student policies are initialized from pre-trained base models without instruction tuning, with Qwen2.5-Math-1.5B used for mathematical tasks.

We train the Generalized Reward Preference Optimization (GRPO) framework with carefully selected hyperparameters to ensure stable and effective optimization under long-horizon rollouts. We adopt Qwen2.5-Math-1.5B as the base student policy for mathematical reasoning tasks.

The training batch size is set to 128, with dynamic batch sizing enabled to accommodate variable-length responses. The maximum prompt length is 1{,}024 tokens, and the maximum response length is capped at 3{,}072 tokens, allowing sufficient capacity for long-form reasoning trajectories.

We use a constant learning rate of $1 \times 10^{-6}$ for policy optimization. For each prompt, rollout is performed by sampling 8 candidate responses with a temperature of 0.6, and rollouts are generated using a vLLM-based inference engine with a tensor parallel size of 2. To stabilize training and prevent policy drift, we apply KL-divergence regularization at both the actor and algorithm levels, using a low-variance KL formulation with a coefficient of 0.001, together with an additional KL control term using the same coefficient.

\textbf{Web Q\&A.} For web search tasks, we compare prompt-based and SFT-based teacher agents built from the same pretrained backbone. The prompt-based teacher uses the base model directly and relies on an instruction prompt to simulate search-engine outputs. The SFT-based teacher starts from the same base model and is further fine-tuned on simulated search data following ZeroSearch~\citep{sun2025zerosearch}. The teacher backbone is Qwen2.5-7B-Base or Qwen2.5-14B-Base. Training-time student rollouts and teacher scoring use these teacher simulators only; all reported Web Q\&A test EM numbers use the Serper API as the sole \emph{live} search tool at inference, aligned with Search-R1 and ZeroSearch evaluation. During training, each teacher provides step-level verbal scores evaluating both query quality and reasoning coherence. We set $\tau_{\mathrm{tr}}=7$ during teacher trajectory generation. For the 14B teacher rows in Appendix~\ref{sec:ovd_teacher_scale_full} (Table~\ref{tab:ovd_teacher_scale}), we use the 14B-based teacher agent. For Web Q\&A tasks, the student policies are initialized from the pre-trained base models Qwen2.5-3B-Base and LLaMA-3.2-3B-Base without instruction tuning.

\paragraph{Prompt-Based Search Simulation Example.} The illustrative example below shows how an instruction prompt turns a search query into a simulated result. It is not the verbatim experimental prompt or a recorded model output.
\begin{tcolorbox}[
  width=\linewidth,
  colback=calloutback,
  colframe=calloutframe,
  colbacktitle=callouttitle,
  coltitle=black,
  fonttitle=\bfseries,
  title={Illustrative prompt and simulated search output},
  titlerule=0pt,
  boxrule=0.6pt,
  arc=1pt,
  boxsep=4pt,
  left=4pt,right=4pt,top=3pt,bottom=3pt,
  before skip=3pt,
  after skip=3pt,
]
\small
\emph{Instruction prompt:}\\
Simulate a search engine. Given a search query, return a concise result with a title and a factual snippet relevant to the query. Generate the result without accessing a live search engine.\\[0.5\baselineskip]
\emph{Search query:}\\
2012 Summer Olympics host city\\[0.5\baselineskip]
\emph{Simulated search output:}\\
Title: London 2012 Summer Olympics\\
Snippet: London, United Kingdom, hosted the 2012 Summer Olympics from 27 July to 12 August 2012.
\end{tcolorbox}

We train the Generalized Reward Preference Optimization (GRPO) framework with carefully selected hyperparameters to ensure stable and effective optimization. We adopt Llama-3.2-3B (\texttt{meta-llama/Llama-3.2-3B}) and Qwen-2.5-3B (\texttt{Qwen/Qwen-2.5-3B}) as the base student policies. Training is conducted for a total of 203 optimization steps.

The training batch size is set to 64, while the validation batch size is 512. The maximum prompt length is 4{,}096 tokens, and the maximum response length is capped at 500 tokens, allowing sufficient capacity for both mathematical reasoning and web-based question answering tasks.

We use a learning rate of $1 \times 10^{-6}$ with a linear warmup schedule covering 95\% of the total training steps. Policy optimization is performed within the GRPO framework, using a mini-batch size of 32 and a micro-batch size of 16. To stabilize training and prevent policy drift, we apply KL-divergence regularization with a coefficient of 0.001, using a low-variance KL formulation.

\section{Web Q\&A Exact Match on LLaMA-3.2-3B-Base}
\label{sec:webqa_llama_app}

\noindent
\begin{minipage}[t]{0.36\textwidth}
\small
This section reports the companion Web Q\&A benchmark sheet for the LLaMA-3.2-3B-Base student. We use the same eight datasets, exact-match metric, column abbreviations, and $(\tau_{\mathrm{tr}},\tau_{\mathrm{te}})$ notation as Table~\ref{tab:qa_em_results} in the main paper (Triv.{=}TriviaQA, Hot.{=}HotpotQA; 2W, Mus., Bam., GAI as there). The table lists search baselines (Search-o1, Search-R1, ZeroSearch) and \textsc{OVD} variants under different $(\tau_{\mathrm{tr}},\tau_{\mathrm{te}})$ settings; OVD$^\dagger$ denotes score-vocabulary sampling at test time, consistent with the main table. \textsc{OVD} markedly outperforms these search baselines on average exact match, mirroring the large margins observed for the Qwen-2.5-3B-Base student in Table~\ref{tab:qa_em_results}.
\end{minipage}%
\hfill
\begin{minipage}[t]{0.61\textwidth}
\centering
\captionsetup{type=table,aboveskip=0.25ex,belowskip=5pt,skip=5pt}
\captionof{table}{Web Q\&A exact match (\%) with LLaMA-3.2-3B-Base as the student. Benchmark abbreviations and threshold settings follow Table~\ref{tab:qa_em_results}.}
\label{tab:qa_em_llama}
\footnotesize
\setlength{\tabcolsep}{2pt}
\resizebox{\linewidth}{!}{%
\begin{tabular}{lccccccccc}
\toprule
\multicolumn{10}{c}{\cellcolor[HTML]{E5E5FC}\textbf{\textit{LLaMA-3.2-3B-Base}}} \\
\midrule
\textbf{Method} & \textbf{NQ} & \textbf{Triv.} & \textbf{Pop} & \textbf{Hot.} & \textbf{2W} & \textbf{Mus.} & \textbf{Bam.} & \textbf{GAI} & \textbf{Avg.} \\
\midrule
Search-o1 & 24.20 & 48.40 & 8.80 & 19.40 & 17.40 & 6.00 & 32.00 & 1.20 & 19.68 \\
Search-R1 & 41.20 & 60.00 & 44.00 & 29.60 & 31.60 & 13.60 & 19.44 & 4.69 & 30.52 \\
ZeroSearch & 23.80 & 41.00 & 18.00 & 22.00 & 26.80 & 4.80 & 11.10 & 3.03 & 18.69 \\
\cdashline{1-10}[5pt/5pt]
\noalign{\vskip 2pt}
OVD ($\tau_{\mathrm{tr}}{=}10$, $\tau_{\mathrm{te}}{=}0$) & 39.40 & 57.40 & 48.20 & 31.40 & 31.00 & 14.00 & 15.20 & 7.27 & 30.58 \\
OVD ($\tau_{\mathrm{tr}}{=}10$, $\tau_{\mathrm{te}}{=}5$) & 43.40 & 62.20 & 54.80 & 31.00 & 29.20 & 18.40 & 25.60 & 7.88 & 34.06 \\
OVD ($\tau_{\mathrm{tr}}{=}10$, $\tau_{\mathrm{te}}{=}10$) & 41.60 & 57.80 & 33.00 & 34.00 & 21.00 & \textbf{34.40} & 21.00 & 7.27 & 31.01 \\
OVD ($\tau_{\mathrm{tr}}{=}5$, $\tau_{\mathrm{te}}{=}0$) & 44.80 & \textbf{64.80} & 54.60 & 33.60 & 33.60 & 18.80 & 22.40 & \textbf{9.09} & 35.19 \\
OVD ($\tau_{\mathrm{tr}}{=}5$, $\tau_{\mathrm{te}}{=}5$) & \textbf{46.80} & 64.20 & 54.80 & \textbf{37.00} & \textbf{36.20} & 22.20 & 27.20 & 8.48 & 37.01 \\
OVD ($\tau_{\mathrm{tr}}{=}5$, $\tau_{\mathrm{te}}{=}10$) & 45.60 & 64.00 & \textbf{57.20} & 35.20 & 35.60 & 18.00 & \textbf{32.80} & 7.88 & \textbf{37.04} \\
OVD$^\dagger$ ($\tau_{\mathrm{tr}}{=}5$, $\tau_{\mathrm{te}}{=}5$) & 45.80 & 64.40 & 54.80 & 34.20 & \textbf{36.20} & 22.20 & 28.00 & 7.27 & 36.51 \\
OVD$^\dagger$ ($\tau_{\mathrm{tr}}{=}10$, $\tau_{\mathrm{te}}{=}5$) & 42.00 & 57.80 & 54.40 & 31.00 & 34.40 & 17.20 & 28.00 & 8.48 & 34.16 \\
\bottomrule
\end{tabular}%
}
\end{minipage}

\section{Test-Time Threshold Ablation}
\label{sec:webqa_threshold_ablation}

Table~\ref{tab:webqa_threshold_ablation} reports the complete Web Q\&A results for $\tau_{\mathrm{te}}\in\{0,5,10\}$ with $\tau_{\mathrm{tr}}=5$ fixed.

\begin{table*}[t]
\centering
\caption{Effect of the test-time rejection threshold on Web Q\&A exact match (\%). The student is Qwen-2.5-3B-Base, the teacher is Qwen2.5-7B, and $\tau_{\mathrm{tr}}=5$. Best results are in bold.}
\label{tab:webqa_threshold_ablation}
\small
\setlength{\tabcolsep}{3pt}
\renewcommand{\arraystretch}{1.15}
\begin{tabular}{@{}cccccccccc@{}}
\toprule
& \multicolumn{3}{c}{Single-Hop QA} & \multicolumn{5}{c}{Multi-Hop QA} & \\
\cmidrule(lr){2-4}\cmidrule(lr){5-9}
$\tau_{\mathrm{te}}$ & NQ & TriviaQA & PopQA & HotpotQA & 2Wiki & Musique & Bamboogle & GAIA & Avg. \\
\midrule
0 & \textbf{49.20} & 67.20 & 63.20 & \textbf{39.20} & 35.20 & 21.80 & 15.20 & 9.09 & 37.51 \\
5 & 47.00 & \textbf{67.40} & \textbf{66.40} & 38.00 & 37.60 & 23.20 & 24.80 & 7.88 & 39.04 \\
10 & 46.00 & 66.80 & 64.80 & 38.80 & \textbf{43.80} & \textbf{25.00} & \textbf{40.00} & \textbf{10.30} & \textbf{41.94} \\
\bottomrule
\end{tabular}
\end{table*}

Table~\ref{tab:webqa_score_sampling_ablation} compares greedy score decoding with score-vocabulary sampling and includes the sampling result at the higher training threshold $\tau_{\mathrm{tr}}=10$.

\begin{table*}[t]
\centering
\caption{Greedy versus sampled teacher scores on Web Q\&A, with Qwen-2.5-3B-Base as student and Qwen2.5-7B as teacher. Greedy decoding takes the most likely score; sampling uses temperature 0.7 over the ten score tokens. We report exact match (\%) at $\tau_{\mathrm{te}}=5$. Best results are in bold.}
\label{tab:webqa_score_sampling_ablation}
\small
\setlength{\tabcolsep}{2.8pt}
\resizebox{\textwidth}{!}{%
\begin{tabular}{lcccccccccc}
\toprule
\textbf{Score Decoding} & $\boldsymbol{\tau_{\mathrm{tr}}}$ & \textbf{NQ} & \textbf{TriviaQA} & \textbf{PopQA} & \textbf{HotpotQA} & \textbf{2Wiki} & \textbf{Musique} & \textbf{Bamboogle} & \textbf{GAIA} & \textbf{Avg.} \\
\midrule
Greedy & 5 & 47.00 & 67.40 & 66.40 & 38.00 & 37.60 & 23.20 & 24.80 & 7.88 & 39.04 \\
Sampling & 5 & \textbf{48.80} & \textbf{68.00} & \textbf{66.60} & \textbf{39.40} & \textbf{41.40} & \textbf{24.20} & \textbf{28.80} & 11.50 & \textbf{41.09} \\
Sampling & 10 & 25.60 & 39.40 & 43.20 & 22.20 & 15.80 & 10.80 & 12.80 & \textbf{12.72} & 22.82 \\
\bottomrule
\end{tabular}%
}
\end{table*}

\FloatBarrier

\section{SFT Teacher Scale}
\label{sec:ovd_teacher_scale_full}

\noindent Table~\ref{tab:ovd_teacher_scale} reports per-dataset EM and F1 for the 7B vs.\ 14B SFT teacher ablation. Bold EM entries match the strongest 7B-teacher configuration from Table~\ref{tab:qa_em_results}; F1 includes estimated values where noted in the caption.

\noindent We hold the student at Qwen-2.5-3B-Base and compare SFT search teachers built on Qwen2.5-7B-Base versus Qwen2.5-14B-Base, with $\tau_{\mathrm{tr}}=10$ and $\tau_{\mathrm{te}}\in\{0,5,10\}$. Macro-average EM is higher with the 7B teacher than the 14B teacher at each $\tau_{\mathrm{te}}$ under this schedule (40.97\% vs.\ 38.41\% at $\tau_{\mathrm{te}}=0$; 42.79\% vs.\ 39.59\% at~5; 43.96\% vs.\ 41.09\% at~10). The pattern is not explained by a single outlier dataset: the 7B teacher leads on GAIA and on several multi-hop benchmarks at comparable thresholds, even though a larger backbone might be expected to yield stricter or more informative verbal judgments. We therefore treat teacher scale as a knob that must be validated rather than assumed monotonic when rejection schedules are held fixed.

\begin{table*}[t]
\centering
\caption{Web Q\&A results with 7B and 14B SFT teachers and a Qwen-2.5-3B-Base student. Training uses $\tau_{\mathrm{tr}}=10$; test thresholds are 0, 5, and 10. Scores are EM and F1 (\%). Unavailable F1 scores are estimated from server logs; GAIA F1 is extrapolated from a fitted EM-to-F1 trend. Bold EM scores mark the best 7B-teacher setting for each benchmark and Avg.}
\label{tab:ovd_teacher_scale}
\tiny
\setlength{\tabcolsep}{1.5pt}
\resizebox{\textwidth}{!}{%
\begin{tabular}{ll*{18}{c}}
\toprule
\textbf{Teacher} & $\boldsymbol{\tau_{\mathrm{te}}}$ &
\multicolumn{2}{c}{\textbf{NQ}} & \multicolumn{2}{c}{\textbf{TriviaQA}} &
\multicolumn{2}{c}{\textbf{PopQA}} & \multicolumn{2}{c}{\textbf{HotpotQA}} &
\multicolumn{2}{c}{\textbf{2Wiki}} & \multicolumn{2}{c}{\textbf{Musique}} &
\multicolumn{2}{c}{\textbf{Bamboogle}} & \multicolumn{2}{c}{\textbf{GAIA}} &
\multicolumn{2}{c}{\textbf{Avg.}} \\
& & EM & F1 & EM & F1 & EM & F1 & EM & F1 & EM & F1 & EM & F1 & EM & F1 & EM & F1 & EM & F1 \\
\midrule
\multicolumn{20}{l}{\textit{SFT teacher: Qwen2.5-7B-Base}} \\
7B & 0 & \textbf{51.00} & 53.89 & 68.80 & 73.57 & 69.00 & 56.58 & 39.40 & 46.69 & 42.00 & 46.42 & 21.00 & 27.58 & 23.20 & 35.59 & \textbf{13.33} & 25.56 & 40.97 & 45.73 \\
7B & 5 & 50.00 & 52.89 & 67.80 & 72.57 & \textbf{69.80} & 57.38 & 39.00 & 46.29 & 43.70 & 48.12 & \textbf{27.90} & 34.48 & 32.00 & 44.39 & 12.12 & 24.69 & 42.79 & 47.60 \\
7B & 10 & 50.00 & 48.88 & \textbf{69.20} & 70.81 & 65.20 & 45.33 & \textbf{42.00} & 48.04 & \textbf{45.00} & 46.07 & 26.40 & 33.50 & \textbf{42.40} & 48.42 & 11.51 & 24.26 & \textbf{43.96} & 45.66 \\
\midrule
\multicolumn{20}{l}{\textit{SFT teacher: Qwen2.5-14B-Base}} \\
14B & 0 & 48.60 & 51.49 & 66.40 & 71.17 & 64.80 & 52.38 & 38.40 & 45.69 & 38.60 & 43.02 & 22.80 & 29.38 & 16.80 & 29.19 & 10.90 & 23.82 & 38.41 & 43.27 \\
14B & 5 & 47.40 & 50.29 & 65.80 & 70.57 & 66.40 & 53.98 & 37.00 & 44.29 & 43.60 & 48.02 & 23.20 & 29.78 & 24.80 & 37.19 & 8.50 & 22.10 & 39.59 & 44.53 \\
14B & 10 & 48.00 & 50.89 & 66.60 & 71.37 & 65.40 & 52.98 & 39.00 & 46.29 & 41.80 & 46.22 & 24.00 & 30.58 & 33.60 & 45.99 & 10.30 & 23.39 & 41.09 & 45.96 \\
\bottomrule
\end{tabular}%
}
\end{table*}

\section{Related Work}
\label{sec:related_work}
This section reviews reinforcement learning methods for improving language models, covering preference optimization, inducing reasoning abilities, and agent training.
\subsection{Post-training Methods for Language Models}

\textbf{Proximal Policy Optimization.}
Reinforcement learning from human feedback (RLHF)~\citep{christiano2017deep} has become a standard approach for aligning LLMs with human preferences. \citet{stiennon2020learning} first demonstrated the effectiveness of training reward models from human feedback for open-ended language generation tasks, which was later scaled to instruction-following settings by \citet{ouyang2022training}. The RLHF paradigm typically involves training a reward model on human preference data and then fine-tuning the LLM using policy gradient methods, among which Proximal Policy Optimization (PPO)~\citep{schulman2017proximal} has been most widely adopted due to its stability and sample efficiency. \citet{ziegler2019fine} first applied PPO to fine-tune language models from human preferences, establishing the foundation for modern RLHF pipelines. More recently, \citet{ahmadian2024back} revisited vanilla policy gradient methods, showing that they can achieve competitive performance while reducing implementation complexity. Despite its effectiveness, PPO-based training requires maintaining multiple models and is computationally expensive, often requiring thousands of GPU hours.

\textbf{Direct Preference Optimization.}
Direct Preference Optimization (DPO)~\citep{rafailov2024direct} reformulates the RLHF objective to directly optimize the policy without explicit reward modeling, significantly simplifying the training pipeline. Subsequent work has proposed various improvements: IPO~\citep{azar2024ipo} provides a general theoretical framework addressing overfitting issues in DPO, ORPO~\citep{hong2024orpo} eliminates the need for a reference model through odds ratio-based optimization, and SimPO~\citep{meng2024simpo} further simplifies the objective with a reference-free reward formulation.

\textbf{Reasoning via Reinforcement Learning.}
Recent work demonstrates that reinforcement learning (RL) and environment interaction provide effective supervision for enhancing reasoning and agentic capabilities of LLMs. Existing methods can be broadly categorized by their supervision source:
(1) \emph{Self-Training RL} bootstraps reasoning from model-generated rationales, including iterative refinement in STaR~\citep{zelikman2022star}, generate-filter-finetune iterations for math and coding in ReST-EM~\citep{singh2024humanlevel}, and ReST-style self-improvement of ReAct agents~\citep{gulcehre2023rest}; (2) \emph{Policy Optimization} directly optimizes reasoning behaviors with RL objectives, where GRPO substantially improves mathematical reasoning~\citep{shao2024deepseekmath}, pure RL induces emergent step-by-step reasoning transferable via distillation~\citep{deepseek2025r1}, and recent variants enhance training stability and implicit process supervision~\citep{yu2025dapo,cui2025process}; (3) \emph{Environment-Grounded Agents} learn through interaction with external or simulated environments, enabling search-augmented reasoning~\citep{jin2025searchr1}, simulated retrieval behaviors~\citep{sun2025zerosearch}, and improved generalization via environment scaling~\citep{fang2025envscaling}. Effectively leveraging environment-derived supervision remains an open problem.

\subsection{Knowledge Distillation for Large Language Models}
Knowledge distillation~\citep{hinton2015distilling} is a core technique for compressing large models into smaller ones, and LLM distillation methods can be broadly categorized by training paradigm and supervision signal.

\textbf{Token-Level Distillation.}
Token-level methods provide dense supervision at each decoding step, offering richer gradient signals than sequence-level approaches. These methods can be categorized by their divergence objectives: (1) \emph{Forward KL} minimizes $\text{KL}(\pi_{\mathcal E} \| \pi_S)$, encouraging the student to cover the teacher's probability mass~\citep{hinton2015distilling,gu2024minillm}; (2) \emph{Reverse KL} minimizes $\text{KL}(\pi_S \| \pi_{\mathcal E})$, promoting mode-seeking behavior that prevents the student from assigning probability mass to unlikely regions~\citep{gu2024minillm}; (3) \emph{Adaptive schemes} dynamically adjust the distillation objective based on token importance, using fine-grained divergence control~\citep{song2024todi} or attribution-based token selection~\citep{wu2024adkd}.

\textbf{Sequence-Level Distillation.}
Sequence-level methods use complete outputs or reasoning traces rather than matching every next-token distribution, providing global and structured learning signals. These approaches can be grouped by the form of sequence supervision: (1) \emph{Output Imitation} directly trains the student on teacher-generated sequences~\citep{kim2016sequence}, with extensions based on imitation learning~\citep{lin2020autoregressive}, symbolic supervision~\citep{west2022symbolic}, and unified f-divergence formulations~\citep{wen2023fdistill}; (2) \emph{Trajectory Traces} leverage intermediate reasoning steps as supervision. Inspired by Chain-of-Thought prompting~\citep{wei2022chain}, teachers generate step-by-step rationales that are distilled into smaller students~\citep{ho2023large,magister2023teaching,fu2023specializing}, enabling effective transfer of multi-step reasoning ability; (3) \emph{Policy-Based Reasoning Distillation} applies preference optimization to generated chains, as in CoTD-PO~\citep{niu2025cotd}. Related work in general reinforcement learning, such as Proximal Policy Distillation~\citep{wang2024proximal}, combines student-driven distillation with PPO but does not study language-model reasoning sequences. Notably, distilling structured rationales can even enable smaller models to outperform larger ones under limited data~\citep{hsieh2023distilling}.

\textbf{On-Policy vs Off-Policy Distillation.} Off-policy distillation pre-generates training data from teacher models, but suffers from distribution mismatch. Generalized Knowledge Distillation (GKD)~\citep{agarwal2024onpolicy} addresses this by training on student-generated samples and allows a mixture of on-policy and fixed-data sampling. More recently, \citet{ye2025black} proposed black-box on-policy distillation, which extends on-policy learning to settings where the teacher model's internal probabilities are inaccessible. However, their approach still requires a discriminative model to match the distribution. To address this limitation, our work incorporates verbal scoring from teacher agents, eliminating the need for token-level probability distillation while providing dense step-level supervision for reasoning tasks.


\clearpage
\section{Mathematical Evaluation Datasets}
\label{sec:math_dataset_descriptions}

We describe the datasets used in Table~\ref{tab:fixed_nine_math} and Figure~\ref{fig:hes_sampling_collapse}, with evaluated question counts. Evaluation files come from the \href{https://github.com/QwenLM/Qwen2.5-Math/tree/main/evaluation/data}{Qwen2.5-Math evaluation release}.

\paragraph{AIME24 (30 problems).}
AIME24 contains problems from the 2024 American Invitational Mathematics Examination, a competition for students who qualify through the AMC. The questions require mathematical problem solving and have integer-valued answers. See the \href{https://maa.org/maa-invitational-competitions/}{organizer's overview}.
Evaluation file: \href{https://github.com/QwenLM/Qwen2.5-Math/tree/main/evaluation/data/aime24/test.jsonl}{\texttt{aime24}}.

\paragraph{AMC23 (40 problems).}
AMC23 comprises a 40-question subset drawn from the 2023 AMC 12A and AMC 12B competitions. We use the problems and answers provided in the evaluation release.
Evaluation file: \href{https://github.com/QwenLM/Qwen2.5-Math/tree/main/evaluation/data/amc23/test.jsonl}{\texttt{amc23}}.

\paragraph{ASDiv (2,215 problems).}
ASDiv~\citep{miao2020diverse} contains English math word problems spanning diverse linguistic patterns and elementary-school problem types, with answers and solution formulas.
Evaluation file: \href{https://github.com/QwenLM/Qwen2.5-Math/tree/main/evaluation/data/asdiv/test.jsonl}{\texttt{asdiv}}.

\paragraph{CARP (976 problems).}
CARP focuses on computation-intensive middle-school algebra, with natural-language solutions and annotated expression-flow graphs in its original release. We evaluate the English test split. See the \href{https://zenodo.org/records/8005507}{authors' dataset release}.
Evaluation file: \href{https://github.com/QwenLM/Qwen2.5-Math/tree/main/evaluation/data/carp_en/test.jsonl}{\texttt{carp\_en}}.

\paragraph{Middle (101 problems).}
Middle is the Chinese-language CN Middle School 24 set, drawn from the 2024 high-school entrance examinations. It includes school-level numerical, algebraic, and geometric questions, including multiple-choice items. The \href{https://qwenlm.github.io/blog/qwen2-math/}{Qwen2-Math evaluation overview} identifies this benchmark separately from CMATH.
Evaluation file: \href{https://github.com/QwenLM/Qwen2.5-Math/tree/main/evaluation/data/cn_middle_school/test.jsonl}{\texttt{cn\_middle\_school}}.

\paragraph{College (2,818 problems).}
CollegeMath contains textbook exercises across seven areas: algebra, precalculus, calculus, vector calculus, probability, linear algebra, and differential equations. We use the 2,818-question test file, which records the source textbook and mathematical topic for each item.
Evaluation file: \href{https://github.com/QwenLM/Qwen2.5-Math/tree/main/evaluation/data/college_math/test.jsonl}{\texttt{college\_math}}.

\paragraph{Gaokao-II (14 problems).}
Gaokao-II contains Chinese mathematics questions from the 2024 Gaokao Paper II. The released evaluation subset comprises eight single-choice, three multiple-choice, and three fill-in-the-blank questions; it is not the complete examination paper.
Evaluation file: \href{https://github.com/QwenLM/Qwen2.5-Math/tree/main/evaluation/data/gaokao2024_II/test.jsonl}{\texttt{gaokao2024\_II}}.

\paragraph{MATH (5,000 problems).}
MATH contains competition mathematics problems with worked solutions, covering a range of topics and difficulty levels. We evaluate its 5,000-problem test split rather than the smaller MATH500 subset. See the \href{https://arxiv.org/abs/2103.03874}{original dataset paper}.
Evaluation file: \href{https://github.com/QwenLM/Qwen2.5-Math/tree/main/evaluation/data/math/test.jsonl}{\texttt{math}}.

\paragraph{MAWPS (2,065 problems).}
MAWPS brings together arithmetic and algebra word problems from several sources. Solving these problems requires translating the quantities and relationships described in text into mathematical operations. We use the 2,065-question collection in the evaluation release. See the \href{https://aclanthology.org/N16-1136/}{original dataset paper}.
Evaluation file: \href{https://github.com/QwenLM/Qwen2.5-Math/tree/main/evaluation/data/mawps/test.jsonl}{\texttt{mawps}}.

\paragraph{Olympiad (675 problems).}
Olympiad denotes the mathematics evaluation subset of OlympiadBench~\citep{he2024olympiadbench}, not the full bilingual, multimodal benchmark. It contains competition-level problems with worked solutions.
Evaluation file: \href{https://github.com/QwenLM/Qwen2.5-Math/tree/main/evaluation/data/olympiadbench/test.jsonl}{\texttt{olympiadbench}}.

\paragraph{SAT-Math (32 problems).}
SAT-Math contains SAT-style mathematics questions with answer choices, including equations, functions, and geometry. We evaluate the 32 questions provided in the released file, not a complete SAT examination. The \href{https://satsuite.collegeboard.org/higher-ed-professionals/sat-validity/content-domains}{College Board overview} describes the broader SAT mathematics content.
Evaluation file: \href{https://github.com/QwenLM/Qwen2.5-Math/tree/main/evaluation/data/sat_math/test.jsonl}{\texttt{sat\_math}}.

\paragraph{TABMWP (1,000 problems).}
TABMWP~\citep{lu2022dynamic} pairs mathematical word problems with tabular data. We use the evaluation subset with text-form tables, answers, and worked solutions.
Evaluation file: \href{https://github.com/QwenLM/Qwen2.5-Math/tree/main/evaluation/data/tabmwp/test.jsonl}{\texttt{tabmwp}}.

\end{document}